\documentclass[11pt,a4paper]{article}
\usepackage{times,latexsym}
\usepackage{url}
\usepackage[T1]{fontenc}

\usepackage[dvipsnames]{xcolor}
\usepackage[acceptedWithA]{tacl2021v1}
\usepackage{xspace,mfirstuc,tabulary}

\newif\iftaclinstructions
\taclinstructionsfalse % AUTHORS: do NOT set this to true
\iftaclinstructions
\renewcommand{\confidential}{}
\renewcommand{\anonsubtext}{(No author info supplied here, for consistency with
TACL-submission anonymization requirements)}
\newcommand{\instr}
\fi

\iftaclpubformat % this "if" is set by the choice of options

\else

\fi

\usepackage{inconsolata}

\usepackage{graphicx}
\usepackage{xurl}
\usepackage{csquotes}
\usepackage{twemojis}
\usepackage{tabularray}
\usepackage{tikz}
\usepackage{amsmath}
\usepackage{pgfplots}
\pgfplotsset{compat=1.18}
\usepackage{subfigure}
\usepackage{caption}
\usepackage{soul}

\usepackage[capitalize,noabbrev]{cleveref}

\newcommand{\rarr}{\scalebox{1.25}{\twemoji{right arrow}}}
\newcommand{\recom}{\scalebox{1.25}{\twemoji{warning}}}
\newcommand{\futdir}{\scalebox{1.25}{\twemoji{light bulb}}}

\def\NPAPERS{110}
\def\NPAPERSUNBOUNDED{20}
\def\NPAPERSUNBOUNDEDSIMULAUTO{14}
\def\NPAPERSUNBOUNDEDSEGMFREE{6}

\definecolor{lightblue}{RGB}{66,133,244}

\title{How \enquote{Real} is Your Real-Time Simultaneous\\Speech-to-Text Translation System?}

\author{Sara Papi$^\dagger$ \and Peter Pol{\' a}k$^\diamond$  \and Dominik Macháček$^\diamond$  \and Ondřej Bojar$^\diamond$ 
  \\ \\
$^\dagger$Fondazione Bruno Kessler, Italy\\
$^\diamond$Charles University, Faculty of Mathematics and Physics, \\ 
Institute of Formal and Applied Linguistics, Czech Republic \\
$^\dagger$\texttt{spapi@fbk.eu}, $^\diamond$\textit{surname}\texttt{@ufal.mff.cuni.cz} 
}

\date{}

\begin{document}
\maketitle
\begin{abstract}
Simultaneous speech-to-text translation (SimulST) translates source-language speech into target-language text concurrently with the speaker's speech, ensuring low latency for better user comprehension. Despite its intended application to unbounded speech, most research has focused on human pre-segmented speech, simplifying the task and overlooking significant challenges. This narrow focus, coupled with widespread terminological inconsistencies, is limiting the applicability of research outcomes to real-world applications, ultimately hindering progress in the field. Our extensive literature review of \NPAPERS{} papers not only reveals these critical issues in current research but also serves as the foundation for our key contributions. We 1) define the steps and core components of a SimulST system, proposing a standardized terminology and taxonomy; 2) conduct a thorough analysis of community trends, and 3) offer concrete recommendations and future directions to bridge the gaps in existing literature, from evaluation frameworks to system architectures, for advancing the field towards more realistic and effective SimulST solutions.
\end{abstract}

\section{Introduction}
The term of \enquote{simultaneous} was first coined in the field of language interpretation, which is the practice of conveying a speaker's message orally in another language to listeners who would not otherwise understand it.\footnote{Source: \url{https://knowledge-centre-interpretation.education.ec.europa.eu/}.} 
Unlike consecutive interpreting \citep{paulik10_interspeech,doi:10.1080/0907676X.2018.1498531}, where interpretation occurs after the speaker has finished talking, simultaneous interpreting\footnote{It is worth noting that, while this paper draws on the concept of simultaneous human interpreting, which is generally speech-to-speech, our focus here is on speech-to-text translation, with speech-to-speech translation falling outside the scope of this study.} happens concurrently with the speech.\footnote{Source: \url{https://www.atanet.org/client-assistance/consecutive-vs-simultaneous-interpreting-whats-the-difference/}.}

Applying this concept to computer science, specifically in automatic
translation, \textbf{simultaneous speech-to-text translation} (SimulST) is defined as
the process that \enquote{\textit{translates source-language speech into
target-language text concurrently}} \citep{ren-etal-2020-simulspeech}, meaning
that the translation process occurs in parallel with the incremental
acquisition of the input speech. Within this context, the \textbf{real-time}
aspect, i.e.\ the \enquote{\textit{immediate processing and response to inputs,
often within milliseconds to seconds}} \citep{laplante1992real} and, in general with low latency, is crucial for ensuring the synchronicity between input and output, enhancing user comprehension of the translated content \citep{bangalore-etal-2012-real}.

In \citet{fugen2007simultaneous}, the SimulST task has been formalized for the first time and described as the process that takes as input an \enquote{audio stream}, a continuous and unsegmented flow of speech information, and produces the automatic textual translation. Despite this broad definition, the field has since predominantly focused on a much narrower task: translating speech that has been pre-segmented into short utterances of few seconds by humans before translation \citep[among others]{kolss-etal-2008-simultaneous,cho-etal-2015-punctuation,ma-etal-2020-simulmt,zhang2024streamspeech}, following sentence boundaries.
While this approach simplifies the translation process by sidestepping challenges related to audio segmentation \citep{polak-2023-long} and 
selecting audio-textual context to retain from the past \citep{papi2024streamatt}, it offers an incomplete and overly simplistic view of the broader challenges inherent in translating continuous audio streams.

This narrow focus has been reinforced over the years, and recent surveys have continued to emphasize this view, assuming human-segmented audio as the standard setting for task \citep{liu2024recent}, as well as reinforcing a glaring terminological inconsistency affecting the SimulST literature. Terms such as \enquote{streaming}, \enquote{online}, and \enquote{real-time} are often used interchangeably with \enquote{simultaneous,} and many terms are used without explicit definitions, leading to significant ambiguity and confusion in understanding and comparing research work, their results, and subsequent findings, ultimately hindering the progress in the field. 

In this paper, we aim to address this \textit{terminological chaos} and provide a clearer understanding of SimulST and all its challenges, with a particular focus on processing continuous audio streams and the difficulties therein. 
After a brief overview of the speech translation landscape (\S\ref{sec:background}), our contributions are structured as follows:
\begin{itemize}
    \item We define the steps required to build a SimulST system, from audio acquisition to translation presentation, and propose a unified terminology to standardize the task. We also introduce a taxonomy based on the dichotomies identified in our analysis of fundamental system components (\S\ref{sec:simulst});
    \item We present a comprehensive and systematic survey of \NPAPERS{} relevant papers in the field of SimulST, showing significant terminological inconsistencies in the literature, highlighting the prevalent focus of the research on human-segmented speech, and identifying trends within the community  (\S\ref{sec:analysis-and-discussion});
    \item Based on our findings, we advocate for the adoption of coherent terminology in the field and call for a shift in research efforts towards more holistic systems capable of effectively processing and translating continuous audio streams. We also provide general recommendations for the research community and suggest promising directions for future investigations spanning from evaluation frameworks to architectural novelties (\S\ref{sec:recommendations}).
\end{itemize}

\section{Background}
\label{sec:background}

\subsection{Offline Speech Translation}
\label{subsec:offline}

Offline speech translation (ST) is the task of translating speech from the source language into text in the target language. Differently from simultaneous ST, which processes input incrementally, offline ST deals with complete and typically well-formed speech segments, representing one or more sentences. 
This task was the first addressed by the community \citep{Waibel2004SpeechTP}, and its model architectures have evolved significantly over time. Initially, offline ST was tackled using cascade architectures \citep{cascade,Waibel1991JANUSAS}, consisting of an automatic speech recognition model (ASR) that transcribes the speech content, followed by a machine translation (MT) model that translates the transcript into the target language.
Lately, direct architectures -- first developed as statistical approaches \citep{940906,1661501} and, later, as neural-based models \citep{Berard2016ListenAT,Weiss2017SequencetoSequenceMC} -- emerged with the promise of overcoming cascade architectures' inherent limitations \citep{sperber-paulik-2020-speech}, such as error propagation\footnote{Errors in the ASR are directly transferred to the MT model, which cannot recover from them, making it more difficult for the user to understand the original content.} by bypassing intermediate ASR outputs. Although direct architectures initially faced a performance gap compared to cascade models \citep{niehues-etal-2018-iwslt,niehues-etal-2019-iwslt}, their effectiveness has been steadily improving \citep{bentivogli-etal-2021-cascade}, with an increasing number of works adopting this paradigm, as highlighted in the survey by \citet{latif2023sparks}.

\subsection{Audio Segmentation}
\label{subsec:audio-segmentation}
Most contemporary neural systems for speech processing, both cascade and direct models, are primarily designed to handle short utterances due to inherent memory and modeling limitations \cite{dai-etal-2019-transformer,chiu2019comparison}. To address this, the common approach has been to segment speech into smaller chunks before feeding it into the model. 
Ever since the early SimulST systems (e.g., \citealp{woszczyna1998modular, fugen2006open,fugen2007simultaneous}), audio segmentation has been a natural part of the pipeline in practical settings.
In cascaded systems, a typical method for segmentation involves introducing punctuation into the ASR-generated text\footnote{Typically, ASR outputs are lowercase words without any punctuation.} \citep{lu2010better, rangarajan-sridhar-etal-2013-segmentation,cho-etal-2015-punctuation,cho17_interspeech, iranzo-sanchez-etal-2020-direct} and segmenting based on the punctuation obtained for the subsequent steps of the SimulST process.
Direct models, which lack an intermediate transcript, rely on segmentation based solely on speech information.
Early approaches used voice activity detection (VAD; \citealp{sohn1999statistical}), supplemented by some heuristics to improve performance \cite{potapczyk-przybysz-2020-srpols,inaguma-etal-2021-espnet, gaido2021beyond}.
Alternatively, fixed-length segmentation, which divides speech into equally sized segments (usually between 10 and 30 seconds), has been found to often outperform VAD-based methods \citep{sinclair14_interspeech,gaido2021beyond}. However, both approaches neglect syntactic and semantic cues in speech, leading to suboptimal results for ST \cite{sinclair14_interspeech,tsiamas22_interspeech,polak2023long}.
To bridge this gap, recent data-driven approaches have been proposed to model sentence-level segmentation \cite{tsiamas22_interspeech,fukuda22b_interspeech}. 
Although these methods were initially developed for the offline regime, \citet{gaido2021beyond} introduced an algorithm that allows them to be applied to SimulST. Despite this advancement, the effectiveness of these methods in simultaneous settings remains limited \cite{polak2023long}.

\subsection{Long-Form Speech}
\label{subsec:long-form}

Long-form speech refers to long audio segments, such as entire lectures, podcasts, or interviews, where the speech is continuous and unsegmented. 
In the related field of ASR, handling such inputs typically involves segmenting the audio into smaller segments, commonly using VAD tools to detect pauses or speech boundaries \citep{1162800,ferrer2003prosody,novitasari2022improving}. More recent work has introduced approaches where segmentation decisions are embedded directly within the ASR model itself \cite{yoshimura2020end,huang2022e2e}. Additionally, some methods employ fixed segmentation with heuristics to stitch segments together, ensuring the continuity of the recognized speech \cite{chiu2019comparison,radford2023robust} or explore architectures capable of performing ASR without segmentation, processing the speech in its entirety \cite{narayanan2019recognizing,chiu2019comparison,lu2021input,zhang2023google}.

In cascaded ST, the challenge extends to MT systems, which have to handle the long texts generated by ASR models. 
While segmenting long text is usually guided by punctuation and supported by using past sentences as context \cite{tiedemann-scherrer-2017-neural, agrawal-etal-2018-contextual, Kim_Tran_Ney_2019, donato-etal-2021-diverse,fernandes-etal-2021-measuring}, it becomes challenging when ASR output lacks punctuation. This issue is typically addressed by inserting punctuation \cite{lu2010better, rangarajan-sridhar-etal-2013-segmentation, cho17_interspeech}. Recent methods have aimed to completely bypass segmentation, allowing translation models to process continuous text streams and improving translation coherence \cite{schneider-waibel-2020-towards, iranzo2023segmentation}.
In direct ST, research on long-form speech has primarily focused on addressing segmentation challenges. Some studies have integrated previous context to improve translation coherence and quality by mitigating audio segmentation errors \cite{gaido20_interspeech, zhang-etal-2021-beyond, ahmad-etal-2024-findings}. Recent advances in SimulST suggest the potential to completely eliminate external segmentation, significantly reducing latency and improving translation quality \cite{polak2023long, papi2024streamatt}.

\section{What is Simultaneous Speech-to-Text Translation?}
\label{sec:simulst}
In this section, we present the first contribution of our work. We begin with the definition of steps characterizing the SimulST process (\S\ref{subsec:process}), and then provide a unified terminology and taxonomy of the current models developed in the field (\S\ref{subsec:terminology-components}).

\subsection{Process Decomposition}
\label{subsec:process}

We describe the SimulST as a 6-step process, deriving it from a high-level conceptualization of the task from which system implementations may depart in many ways. We start with audio acquisition and conclude with the translation presentation to the user. 
Throughout the paper, we assume the processing of clean non-overlapping speech in one language, delivered by a single speaker. We leave aspects such as robustness to background noise \citep{9747755,hwang2024textless}, speaker diarization \citep{PARK2022101317}, overlapping speech \citep{9746873}, code-switching \citep{weller-etal-2022-end,huber2022code}, and any other issues connected to sound to future work on the topic.

The entire process is illustrated in \cref{fig:steps} and described as follows:

\begin{enumerate}
    \item \label{step:audio-acquisition} \textbf{Audio Acquisition}: 
    The speaker speaks to a microphone that is constantly recording, i.e., collecting the flow of information including unvoiced parts such as pauses or hesitations.
    
    \rarr{} \textbf{Output:} unbounded speech (audio stream) $\mathbf{S}$.
    
    \item \label{step:audio-segmentation} \textit{(Optional)} \textbf{Audio Segmentation}: 
    The audio stream $\mathbf{S}$ is segmented into smaller audio segments, usually of a few seconds, based on the utterances contained in the audio using an audio segmenter model. 
    
    \rarr{} \textbf{Output:} bounded speech (audio segments) $\mathbf{S}_\text{seg}=[S_1,...,S_{U}]$ where $U$ is the number of utterances detected by the audio segmenter.
    
    \item \label{step:buffer-update}\textbf{Speech Buffer Update}: The incoming speech $\mathbf{S}$ (or current segment $S_u \in \mathbf{S}_\text{seg}$) is split into fixed-sized audio chunks  (e.g., $500$ms each) for incremental feeding to the ST model and the available input is updated. The resulting speech chunks $\mathbf{C}=[C_1,...,C_{|\frac{len(\mathbf{S})}{D}|}]$, where $D$ is the fixed-sized duration, are added to the Speech Buffer $\mathbf{B_S}$, which stores the accumulated speech. The model can process the whole buffer or only part of it at each step.
    
    \rarr{} \textbf{Output:} the Speech Buffer $\mathbf{B_S}$ at step $t$ is updated with the \underline{new} content: 
    \begin{equation*}
        \mathbf{B_{S}^t}\leftarrow\mathbf{B_{S}^{t-1}\oplus\mathbf{C}}
    \end{equation*}
    
    \item \label{step:hypothesis-generation} \textbf{Hypothesis Generation}: The current Speech Buffer $\mathbf{B_S^t}$ is fed into an ST model $\mathcal{M}$ (either cascade or direct) together with the Text Buffer $\mathbf{B_T^{t-1}}$, storing the emitted output previously emitted at step $t-1$. The ST model returns the translation hypothesis $\mathbf{H}$:
    \begin{equation*}
        \mathbf{H} \leftarrow \mathcal{M}(\mathbf{B_{S}^t},\mathbf{B_{T}^{t-1}})
    \end{equation*}
    The final output $\mathbf{E}$ is obtained by applying a \emph{decision policy} \citep{grissom-ii-etal-2014-dont}, which is the strategy determining whether to emit the generated hypothesis or part of it or to wait for more input.
    
    \rarr{} \textbf{Output:} the \underline{new} translated text selected by the policy $\mathbf{E}=policy(\mathbf{H})$, which is also appended to the Text Buffer $\mathbf{B_T}$ at step $t$:
    \begin{equation*}
        \mathbf{B_{T}^t}\leftarrow\mathbf{B_{T}^{t-1}}\oplus\mathbf{E}
    \end{equation*}

    \begin{figure}[!t]
    \centering
    \includegraphics[width=0.48\textwidth]{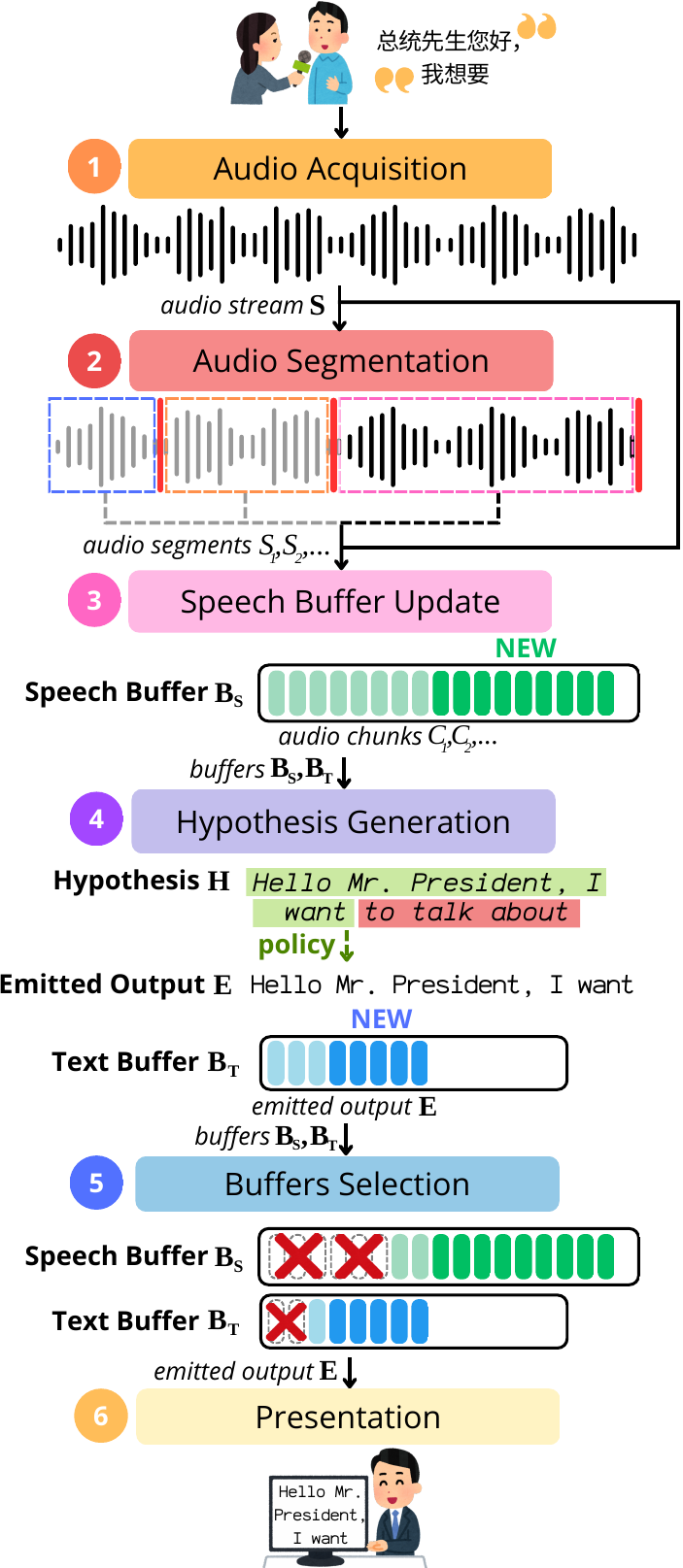}
    \caption{Representation of the steps (1 to 6) of the SimulST process.}
    \label{fig:steps}
    \end{figure}
    
    \item \label{step:buffer-trimming} \textit{(Optional)} \textbf{Speech and Text Buffers Trimming}:
    The content of the Speech and Text Buffers ($\mathbf{B_S}$ and $\mathbf{B_T}$) is trimmed based on the audio-textual information to be retained from the past. This step makes the size of the buffers manageable by ST models, which cannot deal with an infinitely growing context. The content is determined by a \textit{trim} function, which keeps the useful history in the memory for the Hypothesis Generation step (Step 4) at the next step $t+1$: 
    \begin{gather*}
        \mathbf{B_{S}^{t+1}, B_{T}^{t+1}} \leftarrow trim(\mathbf{B_{S}^t}, \mathbf{B_{T}^t})
    \end{gather*}
    The trim function should ensure semantic alignment of the speech and text buffer contents, as significant misalignment between the two may lead to inaccurate translations by the ST model. 
    If the Audio Segmentation step (Step 2) is applied, both Speech and Text Buffers are typically reset (i.e., completely trimmed $\mathbf{B_{\{S,T\}}^{t+1}}\leftarrow\emptyset$) between each audio segment $S_u$ contained in $\mathbf{S}$.
    
    \rarr{} \textbf{Output:} the \underline{old} content contained in the buffers is either reset, trimmed, or left unaltered, providing the Speech and Text Buffers for the next step $\mathbf{B_{S}^{t+1}}$ and $\mathbf{B_{T}^{t+1}}$.
    
    \item \label{step:output-presentation} \textbf{Output Presentation}: The translation is either incrementally presented (e.g., word by word, or using meaningful units), or revised (e.g., such as in re-translation). 
    
    \rarr{} \textbf{Output:} the emitted translation $\mathbf{E}$ is displayed to the user.
\end{enumerate}

The SimulST process aims to balance the \textit{quality} and \textit{latency} of spoken content translation, a balance often referred to as the \textit{quality-latency trade-off}. Latency measures the time from when an information is spoken to when the corresponding output is delivered. 
The quality-latency trade-off is mainly determined by the \textit{decision policy} or, more simply, \textit{policy} in the Hypothesis Generation (Step 4), which decides whether and what part of the hypothesis generated by the model has to be emitted. The decisions made by the policy determine the final output quality and latency, as waiting for more input generally results in higher quality due to increased context but also increases latency. Conversely, emitting output with less context reduces latency but may compromise translation quality. 

The Audio Segmentation (Step 2), in which the audio stream is segmented into short utterances, is commonly employed in the SimulST process (see \S\ref{subsec:terminology-components}). This segmentation addresses the current limitations of neural models in processing very long inputs,\footnote{Suffice it to say that audio input is at least one order of magnitude longer than textual input.} mainly due to memory constraints \citep{10.1145/3530811}.
Utterance boundaries are typically detected using silence-based tools (e.g., VAD, \S\ref{subsec:audio-segmentation}), but since silence often misaligns with semantic boundaries, newer neural models (e.g., SHAS; \citealp{tsiamas22_interspeech}) use semantic content for better accuracy, enhancing translation quality.
This step is optional for approaches that handle unbounded speech \citep{polak-2023-long,papi2024streamatt}, where Speech and Text Buffer Trimming (Step 5) becomes crucial to balance past information with the context length manageable by the ST system.

\begin{table*}[!ht]
\footnotesize
    \centering
    \begin{tblr}{
      colspec={|X|X[5]|}, row{1} = {c}, hlines,
      width=0.98\textwidth,
    }
    \textbf{Term} & \textbf{Definition} \\
    \hline
    \textit{simultaneous} & concurrently receiving input and generating output \\ 
    \textit{real-time} & processing and response to inputs with low latency \\  
    \textit{policy} & the rules regulating when to emit output versus when to wait for more input \\
    \textit{incremental} & sequential over time rather than all at once \\
    \textit{re-translation} & the process of generating hypothesis and revising (either entirely or partially) the previously emitted translation \\
    \textit{unbounded} & a long stream without any explicit information about the overall length 
    \\  
    \textit{bounded} & inputs with a limited length \\
    \textit{segmentation} & the process that splits unbounded inputs into bounded inputs \\ 
    \textit{segmentation-free} & an approach that works on unbounded inputs and does not require segmentation \\  
    \textit{pre-segmentation} & the segmentation is applied to the input before starting the translation process \\
    \textit{audio stream} & a continuous and unsegmented flow of speech data \\
    \textit{audio segment} & a portion of speech of a few seconds resulting from the audio segmentation process \\
    \textit{audio chunk} & a short piece of audio information, usually of fixed length (e.g., $500ms$), used for incremental feeding into ST models \\ 
    \textit{computationally unaware latency} & a metric that measures the time between when information is spoken to when the corresponding output is delivered, assuming zero model computation time \\
    \textit{computationally aware latency} & a metric that measures the time from when information is spoken to when the corresponding output is delivered, also accounting for the model's actual computation time \\
    \end{tblr}
    \caption{Proposed terminology for the SimulST task.}
    \label{tab:terminology}
\end{table*}

\subsection{Terminology and Models' Components}
\label{subsec:terminology-components}

Considering the process described in \S\ref{subsec:process}, we define the terminology related to the SimulST task in Table \ref{tab:terminology}. This terminology offers a precise and unified framework for understanding and analyzing SimulST models and will be consistently adopted throughout this paper.

\begin{figure}
    \centering
    \includegraphics[width=0.48\textwidth]{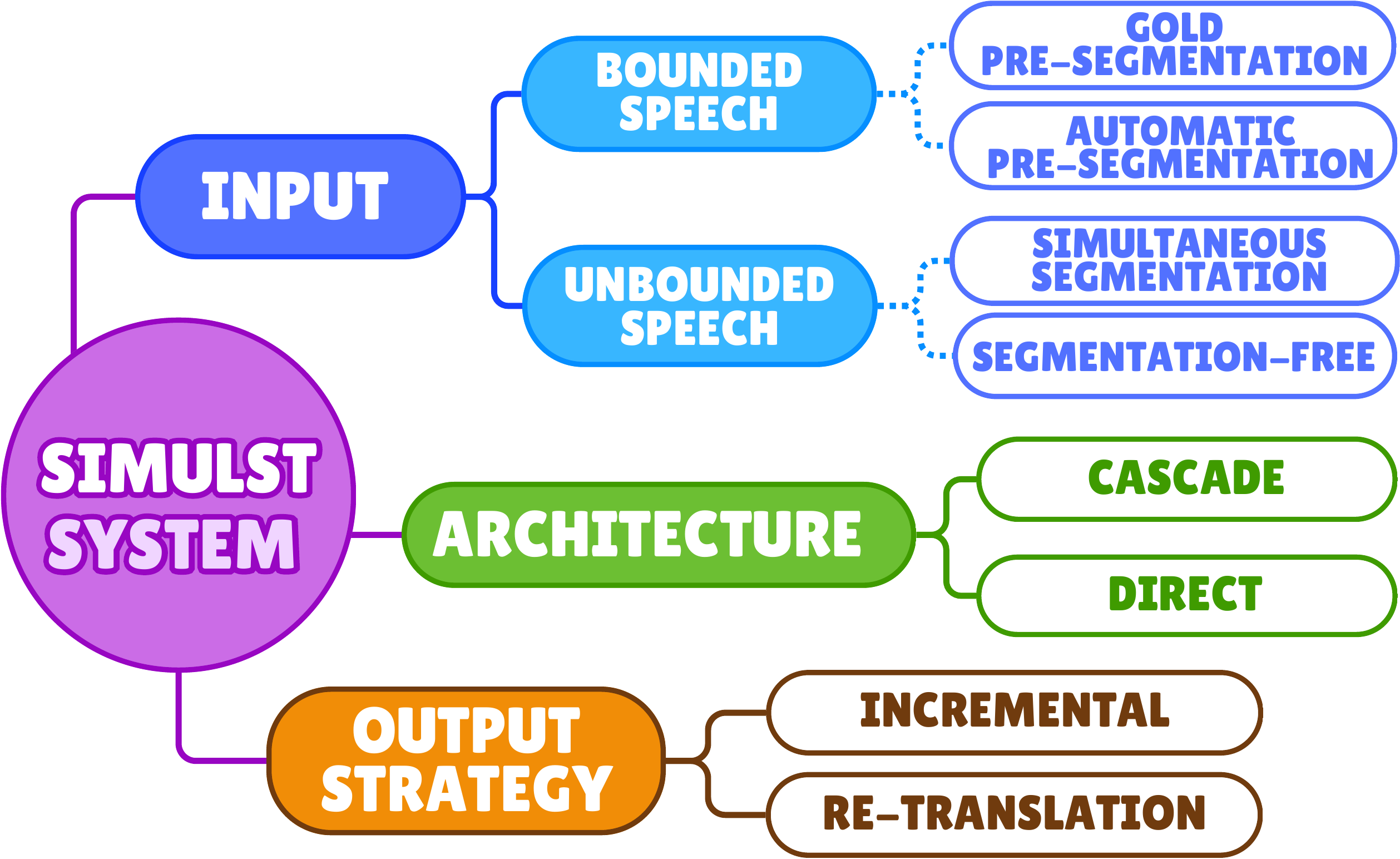}
    \caption{Taxonomy of the SimulST solutions.}
    \label{fig:taxonomy}
\end{figure}

Building on this terminology and considering the common distinctions in the context of speech translation (\S\ref{sec:background}), we classify \NPAPERS{} papers proposing SimulST solutions based on their fundamental components, namely: \textit{input} (either bounded or unbounded speech), \textit{architecture} (either direct or cascade), and \textit{output strategy} (either incremental or re-translation). 
The papers are collected through Semantic Scholar\footnote{\url{https://www.semanticscholar.org/}} using relevant keywords, whose details and specific categorization are presented in Appendix \ref{app:papers-list}.
The resulting taxonomy is visualized in \cref{fig:taxonomy}.

\paragraph{Bounded vs.\ Unbounded Input Speech.}
The input of a SimulST system can be either \textit{bounded} or \textit{unbounded} speech, depending on whether the audio has been pre-segmented into sentences in advance (i.e., offline) or not. Bounded speech refers to short audio segments, usually of a few seconds, representing one or more sentences,\footnote{Sentence-level segmentation should not be confused with word-level segmentation, which is commonly used in SimulST policies \citep{ma-etal-2020-simulmt,dong-etal-2022-learning,zhang-feng-2023-end} to determine which words to emit.} while unbounded speech refers to long audio segments or streams with an unknown duration (\S\ref{subsec:long-form}).
When the input is unbounded and the system processes audio streams directly without any segmentation step (without Step~\ref{step:audio-segmentation} in \cref{subsec:process}), we categorize it as a \textit{segmentation-free} system \citep{iranzo2023segmentation}.
In this case, selecting the speech and text history to retain from the past -- stored in the Speech and Text Buffers (Step~\ref{step:buffer-trimming} in \S\ref{subsec:process}) -- is crucial since audio streams do not have a clear beginning and end, leading to a growing audio-textual context without an explicit resetting mechanism \citep{polak-etal-2023-towards,papi2024streamatt}. 
When the input is unbounded but the system integrates an audio segmentation mechanism that operates jointly with the model in real-time (Step~\ref{step:audio-segmentation} in \S\ref{subsec:process}), we use the term \textit{simultaneous segmentation} \citep{fugen2007simultaneous}. In this case, the history to retain from the past is reset between each automatically detected audio segment.
When the input is bounded, the system is not responsible for audio segmentation or managing the growing context of processing incremental audio streams. Instead, it only handles the hypothesis generation (Step~\ref{step:hypothesis-generation}, \S\ref{subsec:process}), starting from either \textit{automatically pre-segmented audio} (e.g., using VAD tools) or \textit{gold pre-segmented speech} (i.e., audio manually split or post-edited by humans).

\paragraph{Direct vs.\ Cascade Architecture.}

Direct or end-to-end ST architectures are systems that \enquote{\textit{translate speech without using explicitly generated intermediate ASR output}} \citep{sperber-paulik-2020-speech}. This definition extends to the simultaneous translation scenario, distinguishing direct approaches from cascade architectures that employ separate ASR and MT systems, where the best hypothesis of the former serves as input to the latter. \citet{bahar2019comparative} surveyed various direct architectures, many of which leverage multi-task training \citep{44928} -- e.g., incorporating Connectionist Temporal Classification (CTC) loss computed on transcripts \citep{10.1145/1143844.1143891} alongside standard cross-entropy loss -- and pre-training techniques \citep{bansal18_interspeech,bansal-etal-2019-pre} -- e.g., initially training on the ASR task before the ST task -- to enhance model performance. In the context of simultaneous translation, the most prevalent direct architectures include single-encoder single-decoder models (e.g., \citealp{ma-etal-2020-simulmt}), double-encoder models (e.g., \citealp{chen-etal-2021-direct}), and double-decoder models (e.g., \citealp{ren-etal-2020-simulspeech,zeng-etal-2021-realtrans}).

\paragraph{Incremental vs.\ Re-translation.}
SimulST systems produce partial translations to provide a real-time experience to the end user. 
Based on their output strategies, these systems are categorized into \emph{incremental} and \emph{re-translation}.
Re-translation \cite{niehues16_interspeech,NiehuesPhamHa2018_1000087584} allows the system to revise its previous outputs, even after they have been shown to the user. 
Each time, the SimulST system generates the best translation based on the current incremental speech input and decides whether to change the previous partial translation, either entirely or partially \citep{10389709}. The advantage of this approach is that the final translation can achieve a comparable translation quality to an offline system \cite{arivazhagan-etal-2020-translation}. However, frequent changes in the translation can be challenging to process for users, as they need to identify and re-read the updated parts of the translation \citep{9054585}, causing many saccades (i.e., quick movements of eyes). Consequently, evaluating the stability of the emitted output and the flickering phenomena (i.e., how frequently the visualized output changes and how far back the user has to scan to see updates), referred to as \textit{stability-latency trade-off} \citep{Arkhangorodsky_2023}, has become an integral part of re-translation system assessment \citep{zheng-etal-2020-opportunistic}.
Differently, incremental systems \cite{cho2016can,dalvi-etal-2018-incremental} update the translation shown to the user only by appending new tokens. While a wrong output cannot be corrected in subsequent steps, this approach ensures complete stability of the output, minimizing user cognitive effort and eye movements due to the absence of revisions in the visualized output \citep{doi:10.1177/0301006616657097}. Moreover, incremental systems are also well-suited for speech output, where the produced sound can only be extended and never revised.

\paragraph{Computationally aware vs.\ unaware latency.}
The output of a SimulST system is typically evaluated in terms of both quality and latency, as already mentioned in \S\ref{subsec:process}. Latency metrics can be computed in two ways based on how timestamps are assigned to each emitted word or character: either by assuming the \textit{ideal} time, i.e., with zero computational overhead, referred to as \textit{computationally unaware latency}, or by considering the actual \textit{elapsed} time of producing the output, known as \textit{computationally aware latency} \citep{ma-etal-2020-simuleval}. Unlike the computationally unaware latency, which captures aspects such as the timing of decisions made by the SimulST policy and differences in word order between languages, the computationally aware latency includes both the computationally unaware latency and the actual computational time required for the entire process. This measure provides a more realistic assessment of the latency of the SimulST system \citep{ma-etal-2020-simulmt}, but it is strongly influenced by external factors such as the hardware and process optimization being applied (e.g., a more efficient codebase).

\section{Is it \enquote{Real} Simultaneous Translation?}
\label{sec:analysis-and-discussion}

In the following, we analyze and discuss the results obtained by categorizing the papers using the taxonomy depicted in Figure \ref{fig:taxonomy} and whose differences are discussed in \S\ref{subsec:terminology-components}. 

\begin{figure}
    \centering
    \includegraphics[width=\linewidth]{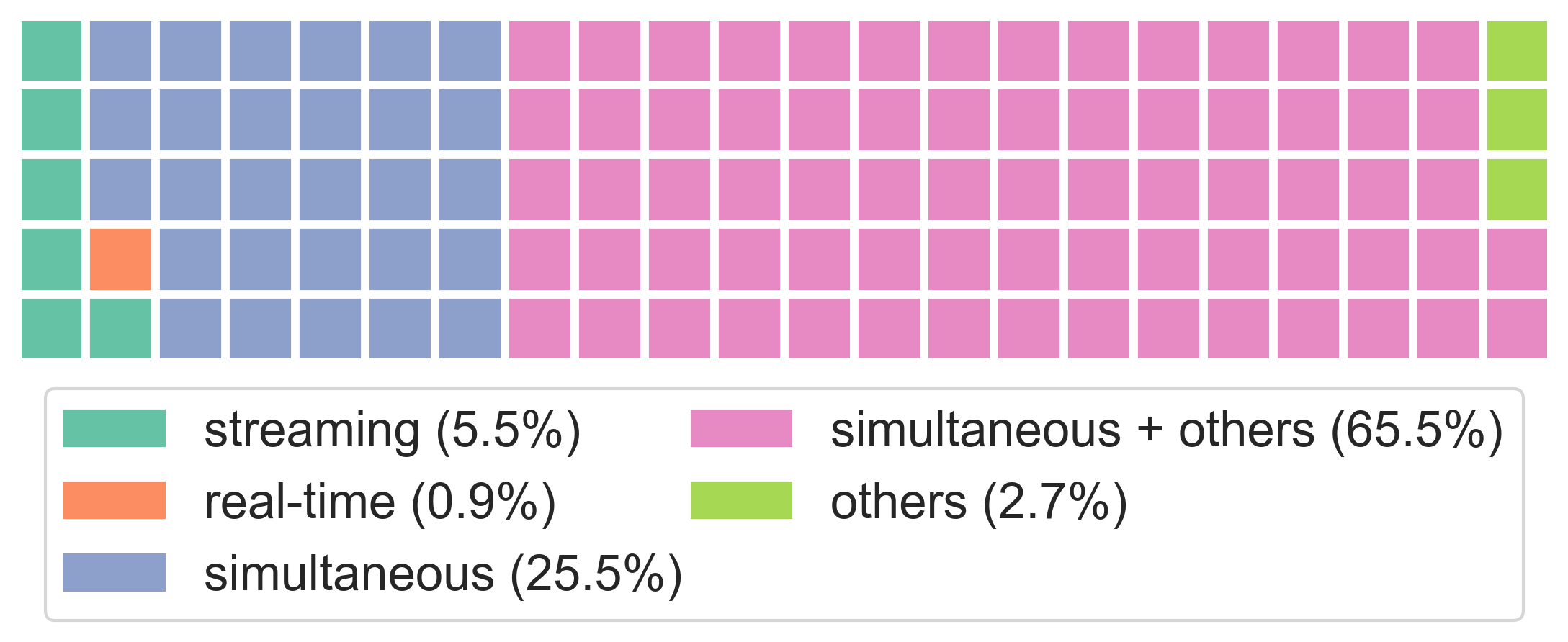}
    \captionsetup{skip=0pt}
    \caption{Waffle plot of the term \enquote{simultaneous} and commonly used synonyms (\enquote{streaming}, \enquote{real-time}, and \enquote{online}) among the \NPAPERS{} categorized papers.}
    \label{fig:waffle-plot}
\end{figure}

\paragraph{The Terminological Chaos.} 
Although \enquote{simultaneous} is the most widely adopted term by the research community to refer to the concurrent speech-to-text translation task, mentioned in 100 out of \NPAPERS{} papers, it is not the only term used in the literature. 
Other commonly used synonyms include \enquote{streaming}, \enquote{online}, and \enquote{real-time}. While \enquote{streaming} is tied to ASR research, where it indicates a model capable of processing incremental speech inputs with the lowest latency possible \citep{9053896,9054476}, \enquote{online} serves to describe the SimulST task as a counterpart to offline speech translation  \citep{ansari-etal-2020-findings,anastasopoulos-etal-2021-findings,anastasopoulos-etal-2022-findings,agrawal-etal-2023-findings}. Instead, \enquote{real-time} is frequently misused to indicate a process that guarantees low latency, which is a goal rather than an accurate description of the concurrent translation task itself. 
We visualize this terminological chaos in \cref{fig:waffle-plot}, which shows that over 65\% of the papers mix and match these terms.
Specifically, 39 papers use at least one of \enquote{streaming}, \enquote{online}, or \enquote{real-time} terms (mostly opting for the former two) interchangeably with \enquote{simultaneous} within the same document, 30 papers employ two of the synonyms (preferring \enquote{streaming} and \enquote{online} over other combinations), and 3 papers even use all four terms. Moreover, some papers exclusively use \enquote{real-time} (1 paper) or \enquote{streaming} (6 papers) to denote the simultaneous translation task, further adding to the confusion. This inconsistent terminology creates significant ambiguity, making it challenging to understand the tasks being addressed, especially when terms are used without explicit definitions. The lack of uniformity calls for a clear, consistent, and standardized task definition in the research landscape, which we addressed in \S\ref{subsec:terminology-components}.

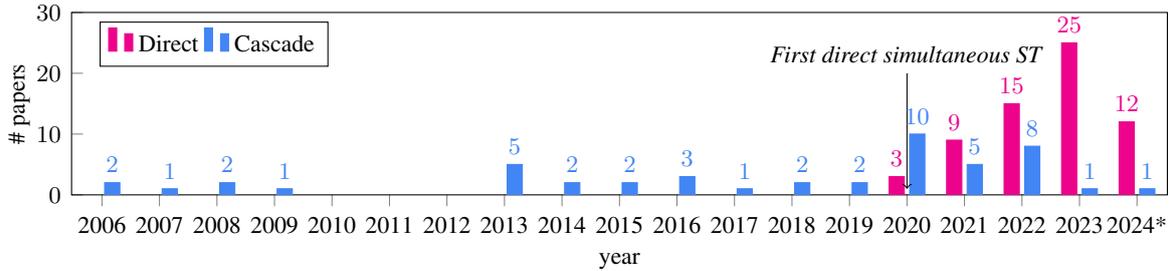
\begin{figure*}[!ht]
    \centering
    \pgfplotstableread[row sep=\\,col sep=&]{
    year & direct & cascade \\
    2006 & 0 & 2 \\
    2007 & 0 & 1 \\
    2008 & 0 & 2 \\
    2009 & 0 & 1 \\
    2010 & 0 & 0 \\
    2011 & 0 & 0 \\
    2012 & 0 & 0 \\
    2013 & 0 & 5 \\
    2014 & 0 & 2 \\
    2015 & 0 & 2 \\
    2016 & 0 & 3 \\
    2017 & 0 & 1 \\
    2018 & 0 & 2 \\
    2019 & 0 & 2 \\
    2020 & 3 & 10 \\
    2021 & 9 & 5 \\
    2022 & 15 & 8 \\
    2023 & 25 & 1 \\
    2024* & 12 & 1 \\
    }\papersnumber
    \begin{tikzpicture}[font=\footnotesize]
    \begin{axis}[
        ybar, width=\textwidth, height=4cm,
        legend style={at={(0.13,0.95)},
                anchor=north,legend columns=-1, font=\footnotesize},
        ylabel={\# papers}, ylabel shift={-5pt},
        xlabel={year},
        y filter/.expression={y==0 ? nan : y},
        symbolic x coords={2006,2007,2008,2009,2010,2011,2012,2013,2014,2015,2016,2017,2018,2019,2020,2021,2022,2023,2024*},
        xtick={2006,2007,2008,2009,2010,2011,2012,2013,2014,2015,2016,2017,2018,2019,2020,2021,2022,2023,2024*},
        nodes near coords,
        nodes near coords align={vertical},
        compat=newest,
        ymin=0,ymax=30,
        xmin=2006, xmax={2024*},
        enlarge x limits={abs=2*\pgfplotbarwidth},
        bar width=0.2cm,
        xtick pos=left,
        ytick pos=left,
        ]
    \addplot[magenta,fill=magenta] table[x=year,y=direct]{\papersnumber};
    \addplot[lightblue,fill=lightblue] table[x=year,y=cascade]{\papersnumber};
    \draw [->] (axis cs:2020,20) node [above] {\textit{First direct simultaneous ST}} -- (axis cs:2020,1);
    \legend{Direct, Cascade}
    \end{axis}
    \end{tikzpicture}
    \captionsetup{skip=0pt}
    \caption{Number of papers in our survey employing direct or cascade simultaneous ST architectures throughout the years. 2024* means that the data are incomplete since the year is not finished yet.}
    \label{fig:direct-trend}
\end{figure*}

\paragraph{Humans will not segment our audio.}
Despite the inherent complexity of SimulST, only a few works address the task from the beginning by handling unbounded speech inputs (\S\ref{subsec:process}). Specifically, only \NPAPERSUNBOUNDED{} papers out of \NPAPERS{} either tackle the concurrent audio segmentation problem for the simultaneous scenario (\NPAPERSUNBOUNDEDSIMULAUTO{} papers) or directly deal with audio streams using a segmentation-free approach (\NPAPERSUNBOUNDEDSEGMFREE{} papers). In stark contrast, most papers (up to 81.8\%) rely on pre-segmented audio as input to their simultaneous models, with nearly all of them (97.7\%) using gold segmentation. This approach oversimplifies the real-world scenario where simultaneous translation is performed, as it is impractical to expect human intervention to segment incoming audio before it is fed to the system.
Although simplifying assumptions are common in research, an astonishing 91.8\% of the papers do not explicitly acknowledge that they assume gold pre-segmented speech for their work.
This oversight means that the majority of research bypasses the challenges associated with simultaneous audio segmentation or with the infinitely growing input, as discussed in \S\ref{subsec:terminology-components}, and silently focuses on the optimal hypothesis generation (Step~\ref{step:hypothesis-generation}, \S\ref{subsec:process}). Moreover, when examining the bounded speech scenario further, we found only 2 papers \citep{kolss-etal-2008-simultaneous,shimizu-etal-2013-constructing} that explore the impact of substituting gold segmentation with automatic segmentation. Consequently, our analysis highlights how divisive the issue of processing unbounded speech is within SimulST research:
a small fraction of research efforts comprehensively analyze and propose solutions for the entire process, while the majority largely ignores these aspects, operating under unrealistic assumptions that are also rarely explicitly mentioned.

\paragraph{A Clear Trend: Direct Models and Incremental Output.} 
Direct models have quickly gained dominance in the SimulST task due to their potential to decrease latency compared to cascade architectures \citep{anastasopoulos-etal-2022-findings}. Among the \NPAPERS{} categorized papers, 64 versus 49 opted for a direct architecture to address the task. This is even more pronounced in the bounded speech scenario, where 67.8\% of the papers leverage a direct approach while being a relatively unaddressed topic in the unbounded speech scenario, with only 3 out of 20 papers using a direct model in their backbone. This trend is also clear in Figure \ref{fig:direct-trend}, which shows that, since their introduction, an increasing number of work employed direct architectures, almost triplicating from 2021 to 2023, while the number of cascade architectures is steadily decreasing after 2020.
The preference for direct models is complemented by a clear prevalence of the incremental output strategy, with 93 out of \NPAPERS{} papers adopting it. Interestingly, in the subset of papers adopting the re-translation strategy, cascade architectures emerge as the preferred choice, with 9 out of 13 papers opting for them. This preference for cascade models in re-translation scenarios contrasts with the general trend in SimulST research, where direct models coupled with incremental output strategies are favored.

\section{Recommendations and Future Directions}
\label{sec:recommendations}

In this section, we outline best practices derived from the analysis in \S\ref{sec:analysis-and-discussion} and the recent advances in the field (\recom), and we highlight key areas where future research is needed to develop more robust, accurate, and efficient SimulST systems capable of meeting real-world demands (\futdir).

\paragraph{\recom{} Use (at least) Automatic Pre-Segmentation.} As discussed in \S\ref{sec:analysis-and-discussion}, the SimulST community has predominantly relied on using gold segmentation for training and evaluating their systems. 
% \repl{Representing}{Because this represents} 
Since this represents unrealistic conditions for real-world SimulST 
%model 
applications, we encourage future research in the bounded speech scenario to use automatic segmentation instead as input for their models. Offline automatic audio segmentation can be achieved using VAD or neural-based tools such as SHAS (\S\ref{subsec:audio-segmentation}). Although all audio files are segmented before starting the simultaneous process, they provide a more realistic input, closer to real-world scenarios where audio segmentation (if any) is performed automatically and on the fly. This shift will better prepare models for practical deployment, ensuring that they can handle the challenges of processing speech that is not always segmented into well-formed sentences.

\paragraph{\recom{} Be Clear about the Type of Speech Input.} While it may sound like a trivial recommendation, it turns out that a vast majority of papers currently neglect the input conditions specification on which the proposed systems work (as highlighted in \S\ref{sec:analysis-and-discussion}). Most SimulST research assumes gold segmentation as the default input for their models, implying that the input is bounded and offline pre-segmented (in advance), a condition that has to be explicitly stated in the experimental settings but almost never is. Some papers only detail the size of the speech chunks that are fed incrementally to the model, which, however, alone does not define the type of speech input but only describes how the information is transferred to the model. Explicitly stating the input type (e.g., gold pre-segmented bounded speech) will provide a more accurate understanding of what are the challenges faced by these systems in practice and has to be included in the model description or, at least, in the experimental settings.

\paragraph{\recom{} Always Report Computationally Unaware Latency (and Optionally Aware).} Latency is one of the key criteria used to evaluate SimulST systems (\S\ref{subsec:process}), and all papers report at least one latency metric. However, there is some variation in how these metrics are presented: some papers report only theoretical (or computationally unaware) latency, others report only computationally aware latency, and a few provide both. Furthermore, in papers using computationally aware metrics, the values are sometimes taken from prior works without recalculating them, even though these metrics are irreproducible without the same hardware setup (\S\ref{subsec:terminology-components}). Given these challenges, we suggest that all papers report computationally unaware metrics, which are always comparable across different hardware setups since they rely solely on theoretical measures. When feasible, computationally aware latency should also be reported, as it provides insight into the real-time usability of the proposed SimulST system, especially when complex or large architectures are involved. In such cases, it is essential to use the same environment (e.g., the GPU and CPU used for running the models and, possibly, the same codebase), for collecting time measurements of the different models being compared to ensure consistency in the resulting metrics.

\paragraph{\futdir{} Create an Evaluation Framework for Unbounded Speech.} 
The most widely adopted evaluation framework for SimulST is SimulEval \citep{ma-etal-2020-simuleval}, with 61 out of \NPAPERS{} papers using the tool, which integrates popular metrics for assessing model performance in terms of both quality (e.g., BLEU; \citealp{papineni-etal-2002-bleu}), and latency (e.g., AL; \citealt{ma-etal-2019-stacl}, DAL; \citealt{cherry2019thinking}, LAAL; \citealt{polak-etal-2022-cuni,papi-etal-2022-generation}, and ATD; \citealt{kano23_interspeech}). However, SimulEval and the aforementioned latency and quality metrics are not designed to compute scores for audio streams and primarily rely on gold pre-segmented inputs. As a result, researchers addressing unbounded speech scenarios have proposed theoretical extensions to these metrics (e.g., StreamLAAL; \citealp{papi2024streamatt}) but have resorted to bounded speech scenarios anyway for comparisons \citep{polak-etal-2023-towards,papi2024streamatt}. This involves calculating sentence-level scores on automatically aligned audio segments adopting tools such as mWERSegmenter \citep{matusov-etal-2005-evaluating}, which is commonly used in ST to handle different audio segmentations between reference and output \citep{anastasopoulos-etal-2021-findings,anastasopoulos-etal-2022-findings,agrawal-etal-2023-findings}. However, mWERSegmenter is prone to alignment errors, which complicates the reliability of the evaluation. 
These reliability issues also impact SLTev \citep{sltev:eacl:2021}, another tool for SimulST model assessment. Despite including useful additions such as stability metrics for re-translation and neural-based quality metrics (e.g., COMET; \citealp{rei-etal-2020-comet,rei-etal-2022-comet}), SLTev still relies on automatic re-alignment.
Another promising starting point is the more recent framework proposed by \citet{huber-etal-2023-end}, which, however, is not as user-friendly as SimulEval, again relies on mWERSegmenter for the alignment, and is currently scarcely adopted.\footnote{At the time of writing, this tool is not even available at the link provided in the paper.}
Given the limitations of the current frameworks and metrics, there emerges a clear need for easy-to-use evaluation methodologies and tools also tailored to the more realistic use case of unbounded speech. Such tools should integrate document-level metrics (e.g., as in SLTev) instead of only sentence-level scores, enabling comparisons between systems that handle audio streams without relying on artificial segmentation settings. This advancement would represent an important step towards shifting the community focus on the unbounded speech scenario, more accurately reflecting the real-world conditions in which SimulST systems operate.

\paragraph{\futdir{} Bear in Mind the Context when Translating.}
Real-world applications of SimulST require systems to operate continuously, processing unbounded speech for extended periods. 
In such scenarios, the context received so far is a valuable source of information that can be employed to improve the accuracy of the provided translations. 
Despite its significance, research explicitly addressing this aspect in SimulST remains limited. 
Existing studies explored the use of memory banks to store relevant information \citep{wu20i_interspeech}, but these solutions are either not suitable for the unbounded speech scenario \citep{raffel-chen-2023-implicit} or claim to support unbounded speech without providing empirical evidence \citep{9414897}.
Beyond SimulST, a limited number of studies focused on explicitly providing context to the ST model for enhancing translation accuracy. Previous approaches include jointly performing document- and sentence-level translation \citep{zhang-etal-2021-beyond} or integrating context through mechanisms like cross-attention \citep{gaido20_interspeech}. 
The selection and memorization of the most relevant information during the translation process is an aspect of particular interest for future research, especially in relation to the emerging paradigm of integrating speech foundation models and large language models for addressing a wide variety of tasks \citep{latif2023sparks}, including speech translation \citep{gaido2024speech}, where elements such as prompts and in-context learning \citep{NEURIPS2020_1457c0d6} become of fundamental importance.

\paragraph{\futdir{} Pay Attention to Output Visualization.} 
An important factor impacting user experience is how the output is delivered. For textual content such as translations, this primarily concerns how they are visualized on the screen \citep{respeaking}. Little work has been devoted to this aspect and existing studies have framed the generated texts as subtitles \citep{machavcek2020presenting,Irvin2021StudentIR,javorsky-etal-2022-continuous} and proposed subtitle-oriented metrics \citep{papi2021visualization}, such as reading speed \citep{doi:10.1080/15213269.2010.502873}, to measure user effort. The aforementioned work also discussed various strategies for delivering the output based on subtitle granularity (i.e., word, lines, and subtitle blocks). However, few studies \citep{javorsky-etal-2022-continuous} have examined the impact of SimulST visualization strategies on user comprehension of the generated content or the cognitive effort introduced by translation revisions (\S\ref{subsec:terminology-components}). For instance, the flickering effect inherent to re-translation approaches \citep{9054585} can cause poor user experience due to re-reading phenomena \citep{doi:10.1080/0907676X.2012.722651} and excessive eye fixations \citep{eyefixation}.  Therefore, an important future direction for the field is to quantify the effect of output visualization on user comprehension, for instance, by involving human evaluation.
Moreover, segmenting the translations for visualization purposes can potentially lead to an overall increased latency of the SimulST systems due to the added processing module. Current subtitle segmentation models, which insert line breaks to satisfy syntactic and semantic constraints for improved readability, were mainly developed for offline ST and are not optimized for low latency or to deal with limited context \citep{matusov-etal-2019-customizing,karakanta-etal-2020-42}. An alternative approach proposed by \citet{papi-etal-2022-dodging} integrates segmentation directly into the sequence-to-sequence model, potentially reducing latency by bypassing additional modules, and represents an interesting direction for further research.

\paragraph{\futdir{} Quantify Quality-Latency Differences in User Experience.} 
The main goal of SimulST research is to maximize translation quality while minimizing latency, aiming for the best quality-latency trade-off. However, few studies have examined the extent to which variations in quality and latency -- whether minor or significant -- actually impact user experience \citep{Irvin2021StudentIR,wang2024exploring}, as well as how automatic translations compare to human interpretations \citep{bizzoni-etal-2020-human,fantinuoli-prandi-2021-towards}. 
Assessing and scoring different SimulST systems with humans in the loop remains a challenging area of ongoing research \citep{sakamoto-etal-2013-evaluation}, as existing methods often suffer from low agreement between participants \citep{wang2024exploring}. \citet{javorsky-etal-2022-continuous} proposed and analyzed the effects of continuous ratings (where human evaluators watch videos or listen to audio with translations created by the model being evaluated and continuously express satisfaction by pressing buttons) against traditional questionnaires, but only for re-translation systems. Later, the continuous rating was shown to correlate with standard quality metrics \citep{machacek-etal-2023-mt}, but its generalizability across different domains and systems remains uncertain.
Future studies should focus not only on ranking different systems but also on providing holistic human judgments for SimulST outputs, placing the user at the center of the evaluation. Quantifying the minimum changes in the quality-latency trade-off that humans can perceive is of the utmost importance to ensure that improvements measured with automatic metrics also have a meaningful impact on final performance.\footnote{Refer to \citet{kocmi-etal-2024-navigating} for a study of meaningful score differences for MT metrics.}

\section{Conclusions}
In this paper, we examined the state of simultaneous speech translation research under several aspects, identifying significant gaps in the existing literature. 
Our analysis of \NPAPERS{} papers revealed a predominant focus in SimulST on human-segmented speech, which oversimplifies the task and neglects the complexities of real-world applications. We also uncovered substantial terminological inconsistencies, revealing real terminological chaos. To address these issues, we formalized the SimulST task as a 6-step process and introduced a unified terminology to standardize research outcomes.
We identified the core components of SimulST systems (input, architecture, and output strategy), discussed current research trends, and provided key recommendations, including transitioning from human to automatic segmentation and adopting consistent terminology.
We also emphasized the need for improvement in current evaluation frameworks, highlighting the importance of creating an easy-to-use tool that can handle unbounded speech, incorporating contextual information during translation, and investigating more user-centric assessments to ensure that improvements measured by automatic metrics align with those in the user experience.

\section*{Acknowledgments}
This paper has received funding from the European Union's Horizon research and innovation programme under grant agreement No 101135798, project Meetween (My Personal AI Mediator for Virtual MEETings BetWEEN People), 
from the Ministry of Education, Youth and Sports of the Czech Republic Project Nr.\ LM2023062 LINDAT/CLARIAH-CZ and Project OP JAK Mezisektorová spolupráce Nr.\ CZ.02.01.01/00/23\_020/0008518 named ``Jazykověda, umělá inteligence a jazykové a řečové technologie: od výzkumu k aplikacím.''
The authors also acknowledge the support of National Recovery Plan funded project MPO 60273/24/21300/21000 CEDMO 2.0 NPO.

\bibliography{biblio}

\begin{thebibliography}{213}
\expandafter\ifx\csname natexlab\endcsname\relax\def\natexlab#1{#1}\fi

\bibitem[{Agarwal et~al.(2023)Agarwal, Agrawal, Anastasopoulos, Bentivogli, Bojar, Borg, Carpuat, Cattoni, Cettolo, Chen, Chen, Choukri, Chronopoulou, Currey, Declerck, Dong, Duh, Est{\`e}ve, Federico, Gahbiche, Haddow, Hsu, Mon~Htut, Inaguma, Javorsk{\'y}, Judge, Kano, Ko, Kumar, Li, Ma, Mathur, Matusov, McNamee, P.~McCrae, Murray, Nadejde, Nakamura, Negri, Nguyen, Niehues, Niu, Kr.~Ojha, E.~Ortega, Pal, Pino, van~der Plas, Pol{\'a}k, Rippeth, Salesky, Shi, Sperber, St{\"u}ker, Sudoh, Tang, Thompson, Tran, Turchi, Waibel, Wang, Watanabe, and Zevallos}]{agrawal-etal-2023-findings}
Milind Agarwal, Sweta Agrawal, Antonios Anastasopoulos, Luisa Bentivogli, Ond{\v{r}}ej Bojar, Claudia Borg, Marine Carpuat, Roldano Cattoni, Mauro Cettolo, Mingda Chen, William Chen, Khalid Choukri, Alexandra Chronopoulou, Anna Currey, Thierry Declerck, Qianqian Dong, Kevin Duh, Yannick Est{\`e}ve, Marcello Federico, Souhir Gahbiche, Barry Haddow, Benjamin Hsu, Phu Mon~Htut, Hirofumi Inaguma, D{\'a}vid Javorsk{\'y}, John Judge, Yasumasa Kano, Tom Ko, Rishu Kumar, Pengwei Li, Xutai Ma, Prashant Mathur, Evgeny Matusov, Paul McNamee, John P.~McCrae, Kenton Murray, Maria Nadejde, Satoshi Nakamura, Matteo Negri, Ha~Nguyen, Jan Niehues, Xing Niu, Atul Kr.~Ojha, John E.~Ortega, Proyag Pal, Juan Pino, Lonneke van~der Plas, Peter Pol{\'a}k, Elijah Rippeth, Elizabeth Salesky, Jiatong Shi, Matthias Sperber, Sebastian St{\"u}ker, Katsuhito Sudoh, Yun Tang, Brian Thompson, Kevin Tran, Marco Turchi, Alex Waibel, Mingxuan Wang, Shinji Watanabe, and Rodolfo Zevallos. 2023.
\newblock \href {https://doi.org/10.18653/v1/2023.iwslt-1.1} {{FINDINGS} {OF} {THE} {IWSLT} 2023 {EVALUATION} {CAMPAIGN}}.
\newblock In \emph{Proceedings of the 20th International Conference on Spoken Language Translation (IWSLT 2023)}, pages 1--61, Toronto, Canada (in-person and online). Association for Computational Linguistics.

\bibitem[{Agrawal et~al.(2018)Agrawal, Turchi, and Negri}]{agrawal-etal-2018-contextual}
Ruchit Agrawal, Marco Turchi, and Matteo Negri. 2018.
\newblock \href {https://aclanthology.org/2018.eamt-main.1} {Contextual handling in neural machine translation: Look behind, ahead and on both sides}.
\newblock In \emph{Proceedings of the 21st Annual Conference of the European Association for Machine Translation}, pages 31--40, Alicante, Spain.

\bibitem[{Ahmad et~al.(2024)Ahmad, Anastasopoulos, Bojar, Borg, Carpuat, Cattoni, Cettolo, Chen, Dong, Federico, Haddow, Javorsk{\'y}, Krubi{\'n}ski, Kim~Lam, Ma, Mathur, Matusov, Maurya, McCrae, Murray, Nakamura, Negri, Niehues, Niu, Ojha, Ortega, Papi, Pol{\'a}k, Posp{\'\i}{\v{s}}il, Pecina, Salesky, Sethiya, Sarkar, Shi, Sikasote, Sperber, St{\"u}ker, Sudoh, Thompson, Waibel, Watanabe, Wilken, Zem{\'a}nek, and Zevallos}]{ahmad-etal-2024-findings}
Ibrahim~Said Ahmad, Antonios Anastasopoulos, Ond{\v{r}}ej Bojar, Claudia Borg, Marine Carpuat, Roldano Cattoni, Mauro Cettolo, William Chen, Qianqian Dong, Marcello Federico, Barry Haddow, D{\'a}vid Javorsk{\'y}, Mateusz Krubi{\'n}ski, Tsz Kim~Lam, Xutai Ma, Prashant Mathur, Evgeny Matusov, Chandresh Maurya, John McCrae, Kenton Murray, Satoshi Nakamura, Matteo Negri, Jan Niehues, Xing Niu, Atul~Kr. Ojha, John Ortega, Sara Papi, Peter Pol{\'a}k, Adam Posp{\'\i}{\v{s}}il, Pavel Pecina, Elizabeth Salesky, Nivedita Sethiya, Balaram Sarkar, Jiatong Shi, Claytone Sikasote, Matthias Sperber, Sebastian St{\"u}ker, Katsuhito Sudoh, Brian Thompson, Alex Waibel, Shinji Watanabe, Patrick Wilken, Petr Zem{\'a}nek, and Rodolfo Zevallos. 2024.
\newblock \href {https://aclanthology.org/2024.iwslt-1.1} {{FINDINGS} {OF} {THE} {IWSLT} 2024 {EVALUATION} {CAMPAIGN}}.
\newblock In \emph{Proceedings of the 21st International Conference on Spoken Language Translation (IWSLT 2024)}, pages 1--11, Bangkok, Thailand (in-person and online). Association for Computational Linguistics.

\bibitem[{Alastruey et~al.(2023)Alastruey, Sperber, Gollan, Telaar, Ng, and Agarwal}]{alastruey-etal-2023-towards}
Belen Alastruey, Matthias Sperber, Christian Gollan, Dominic Telaar, Tim Ng, and Aashish Agarwal. 2023.
\newblock \href {https://aclanthology.org/2023.calcs-1.2} {Towards real-world streaming speech translation for code-switched speech}.
\newblock In \emph{Proceedings of the 6th Workshop on Computational Approaches to Linguistic Code-Switching}, pages 14--22, Singapore. Association for Computational Linguistics.

\bibitem[{Amrhein and Haddow(2022)}]{amrhein-haddow-2022-dont}
Chantal Amrhein and Barry Haddow. 2022.
\newblock \href {https://aclanthology.org/2022.wmt-1.13} {Don{'}t discard fixed-window audio segmentation in speech-to-text translation}.
\newblock In \emph{Proceedings of the Seventh Conference on Machine Translation (WMT)}, pages 203--219, Abu Dhabi, United Arab Emirates (Hybrid). Association for Computational Linguistics.

\bibitem[{Anastasopoulos et~al.(2022)Anastasopoulos, Barrault, Bentivogli, Zanon~Boito, Bojar, Cattoni, Currey, Dinu, Duh, Elbayad, Emmanuel, Est{\`e}ve, Federico, Federmann, Gahbiche, Gong, Grundkiewicz, Haddow, Hsu, Javorsk{\'y}, Kloudov{\'a}, Lakew, Ma, Mathur, McNamee, Murray, N{\v{a}}dejde, Nakamura, Negri, Niehues, Niu, Ortega, Pino, Salesky, Shi, Sperber, St{\"u}ker, Sudoh, Turchi, Virkar, Waibel, Wang, and Watanabe}]{anastasopoulos-etal-2022-findings}
Antonios Anastasopoulos, Lo{\"\i}c Barrault, Luisa Bentivogli, Marcely Zanon~Boito, Ond{\v{r}}ej Bojar, Roldano Cattoni, Anna Currey, Georgiana Dinu, Kevin Duh, Maha Elbayad, Clara Emmanuel, Yannick Est{\`e}ve, Marcello Federico, Christian Federmann, Souhir Gahbiche, Hongyu Gong, Roman Grundkiewicz, Barry Haddow, Benjamin Hsu, D{\'a}vid Javorsk{\'y}, V{\u{e}}ra Kloudov{\'a}, Surafel Lakew, Xutai Ma, Prashant Mathur, Paul McNamee, Kenton Murray, Maria N{\v{a}}dejde, Satoshi Nakamura, Matteo Negri, Jan Niehues, Xing Niu, John Ortega, Juan Pino, Elizabeth Salesky, Jiatong Shi, Matthias Sperber, Sebastian St{\"u}ker, Katsuhito Sudoh, Marco Turchi, Yogesh Virkar, Alexander Waibel, Changhan Wang, and Shinji Watanabe. 2022.
\newblock \href {https://doi.org/10.18653/v1/2022.iwslt-1.10} {Findings of the {IWSLT} 2022 evaluation campaign}.
\newblock In \emph{Proceedings of the 19th International Conference on Spoken Language Translation (IWSLT 2022)}, pages 98--157, Dublin, Ireland (in-person and online). Association for Computational Linguistics.

\bibitem[{Anastasopoulos et~al.(2021)Anastasopoulos, Bojar, Bremerman, Cattoni, Elbayad, Federico, Ma, Nakamura, Negri, Niehues, Pino, Salesky, St{\"u}ker, Sudoh, Turchi, Waibel, Wang, and Wiesner}]{anastasopoulos-etal-2021-findings}
Antonios Anastasopoulos, Ond{\v{r}}ej Bojar, Jacob Bremerman, Roldano Cattoni, Maha Elbayad, Marcello Federico, Xutai Ma, Satoshi Nakamura, Matteo Negri, Jan Niehues, Juan Pino, Elizabeth Salesky, Sebastian St{\"u}ker, Katsuhito Sudoh, Marco Turchi, Alexander Waibel, Changhan Wang, and Matthew Wiesner. 2021.
\newblock \href {https://doi.org/10.18653/v1/2021.iwslt-1.1} {{FINDINGS} {OF} {THE} {IWSLT} 2021 {EVALUATION} {CAMPAIGN}}.
\newblock In \emph{Proceedings of the 18th International Conference on Spoken Language Translation (IWSLT 2021)}, pages 1--29, Bangkok, Thailand (online). Association for Computational Linguistics.

\bibitem[{Ansari et~al.(2020)Ansari, Axelrod, Bach, Bojar, Cattoni, Dalvi, Durrani, Federico, Federmann, Gu, Huang, Knight, Ma, Nagesh, Negri, Niehues, Pino, Salesky, Shi, St{\"u}ker, Turchi, Waibel, and Wang}]{ansari-etal-2020-findings}
Ebrahim Ansari, Amittai Axelrod, Nguyen Bach, Ond{\v{r}}ej Bojar, Roldano Cattoni, Fahim Dalvi, Nadir Durrani, Marcello Federico, Christian Federmann, Jiatao Gu, Fei Huang, Kevin Knight, Xutai Ma, Ajay Nagesh, Matteo Negri, Jan Niehues, Juan Pino, Elizabeth Salesky, Xing Shi, Sebastian St{\"u}ker, Marco Turchi, Alexander Waibel, and Changhan Wang. 2020.
\newblock \href {https://doi.org/10.18653/v1/2020.iwslt-1.1} {{FINDINGS} {OF} {THE} {IWSLT} 2020 {EVALUATION} {CAMPAIGN}}.
\newblock In \emph{Proceedings of the 17th International Conference on Spoken Language Translation}, pages 1--34, Online. Association for Computational Linguistics.

\bibitem[{Ansari et~al.(2021)Ansari, Bojar, Haddow, and Mahmoudi}]{sltev:eacl:2021}
Ebrahim Ansari, Ond{\v{r}}ej Bojar, Barry Haddow, and Mohammad Mahmoudi. 2021.
\newblock \href {https://doi.org/10.18653/v1/2021.eacl-demos.9} {{SLTEV}: Comprehensive evaluation of spoken language translation}.
\newblock In \emph{Proceedings of the 16th Conference of the European Chapter of the Association for Computational Linguistics: System Demonstrations}, pages 71--79, Online. Association for Computational Linguistics.

\bibitem[{Arivazhagan et~al.(2020{\natexlab{a}})Arivazhagan, Cherry, Macherey, and Foster}]{arivazhagan-etal-2020-translation}
Naveen Arivazhagan, Colin Cherry, Wolfgang Macherey, and George Foster. 2020{\natexlab{a}}.
\newblock \href {https://doi.org/10.18653/v1/2020.iwslt-1.27} {Re-translation versus streaming for simultaneous translation}.
\newblock In \emph{Proceedings of the 17th International Conference on Spoken Language Translation}, pages 220--227, Online. Association for Computational Linguistics.

\bibitem[{Arivazhagan et~al.(2020{\natexlab{b}})Arivazhagan, Cherry, Te, Macherey, Baljekar, and Foster}]{9054585}
Naveen Arivazhagan, Colin Cherry, I~Te, Wolfgang Macherey, Pallavi Baljekar, and George Foster. 2020{\natexlab{b}}.
\newblock \href {https://doi.org/10.1109/ICASSP40776.2020.9054585} {Re-translation strategies for long form, simultaneous, spoken language translation}.
\newblock In \emph{ICASSP 2020 - 2020 IEEE International Conference on Acoustics, Speech and Signal Processing (ICASSP)}, pages 7919--7923.

\bibitem[{Arkhangorodsky et~al.(2023)}]{Arkhangorodsky_2023}
Arkady Arkhangorodsky et~al. 2023.
\newblock Method and system for evaluating and improving live translation captioning systems.
\newblock US Patent US20230089902A1.

\bibitem[{Atal and Rabiner(1976)}]{1162800}
Bishnu Atal and Lawrence Rabiner. 1976.
\newblock \href {https://doi.org/10.1109/TASSP.1976.1162800} {A pattern recognition approach to voiced-unvoiced-silence classification with applications to speech recognition}.
\newblock \emph{IEEE Transactions on Acoustics, Speech, and Signal Processing}, 24(3):201--212.

\bibitem[{Bahar et~al.(2019)Bahar, Bieschke, and Ney}]{bahar2019comparative}
Parnia Bahar, Tobias Bieschke, and Hermann Ney. 2019.
\newblock A comparative study on end-to-end speech to text translation.
\newblock In \emph{2019 IEEE Automatic Speech Recognition and Understanding Workshop (ASRU)}, pages 792--799. IEEE.

\bibitem[{Bahar et~al.(2020)Bahar, Wilken, Alkhouli, Guta, Golik, Matusov, and Herold}]{bahar-etal-2020-start}
Parnia Bahar, Patrick Wilken, Tamer Alkhouli, Andreas Guta, Pavel Golik, Evgeny Matusov, and Christian Herold. 2020.
\newblock \href {https://doi.org/10.18653/v1/2020.iwslt-1.3} {Start-before-end and end-to-end: Neural speech translation by {A}pp{T}ek and {RWTH} {A}achen {U}niversity}.
\newblock In \emph{Proceedings of the 17th International Conference on Spoken Language Translation}, pages 44--54, Online. Association for Computational Linguistics.

\bibitem[{Bahar et~al.(2021)Bahar, Wilken, Di~Gangi, and Matusov}]{bahar-etal-2021-without}
Parnia Bahar, Patrick Wilken, Mattia~A. Di~Gangi, and Evgeny Matusov. 2021.
\newblock \href {https://doi.org/10.18653/v1/2021.iwslt-1.5} {Without further ado: Direct and simultaneous speech translation by {A}pp{T}ek in 2021}.
\newblock In \emph{Proceedings of the 18th International Conference on Spoken Language Translation (IWSLT 2021)}, pages 52--63, Bangkok, Thailand (online). Association for Computational Linguistics.

\bibitem[{Bangalore et~al.(2012)Bangalore, Rangarajan~Sridhar, Kolan, Golipour, and Jimenez}]{bangalore-etal-2012-real}
Srinivas Bangalore, Vivek~Kumar Rangarajan~Sridhar, Prakash Kolan, Ladan Golipour, and Aura Jimenez. 2012.
\newblock \href {https://aclanthology.org/N12-1048} {Real-time incremental speech-to-speech translation of dialogs}.
\newblock In \emph{Proceedings of the 2012 Conference of the North {A}merican Chapter of the Association for Computational Linguistics: Human Language Technologies}, pages 437--445, Montr{\'e}al, Canada. Association for Computational Linguistics.

\bibitem[{Bansal et~al.(2018)Bansal, Kamper, Livescu, Lopez, and Goldwater}]{bansal18_interspeech}
Sameer Bansal, Herman Kamper, Karen Livescu, Adam Lopez, and Sharon Goldwater. 2018.
\newblock \href {https://doi.org/10.21437/Interspeech.2018-1326} {{Low-Resource Speech-to-Text Translation}}.
\newblock In \emph{Proc. Interspeech 2018}, pages 1298--1302.

\bibitem[{Bansal et~al.(2019)Bansal, Kamper, Livescu, Lopez, and Goldwater}]{bansal-etal-2019-pre}
Sameer Bansal, Herman Kamper, Karen Livescu, Adam Lopez, and Sharon Goldwater. 2019.
\newblock \href {https://doi.org/10.18653/v1/N19-1006} {Pre-training on high-resource speech recognition improves low-resource speech-to-text translation}.
\newblock In \emph{Proceedings of the 2019 Conference of the North {A}merican Chapter of the Association for Computational Linguistics: Human Language Technologies, Volume 1 (Long and Short Papers)}, pages 58--68, Minneapolis, Minnesota. Association for Computational Linguistics.

\bibitem[{Barrault et~al.(2023)Barrault, Chung, Meglioli, Dale, Dong, Duppenthaler, Duquenne, Ellis, Elsahar, Haaheim et~al.}]{barrault2023seamless}
Lo{\"\i}c Barrault, Yu-An Chung, Mariano~Coria Meglioli, David Dale, Ning Dong, Mark Duppenthaler, Paul-Ambroise Duquenne, Brian Ellis, Hady Elsahar, Justin Haaheim, et~al. 2023.
\newblock Seamless: Multilingual expressive and streaming speech translation.
\newblock \emph{arXiv preprint arXiv:2312.05187}.

\bibitem[{Bentivogli et~al.(2021)Bentivogli, Cettolo, Gaido, Karakanta, Martinelli, Negri, and Turchi}]{bentivogli-etal-2021-cascade}
Luisa Bentivogli, Mauro Cettolo, Marco Gaido, Alina Karakanta, Alberto Martinelli, Matteo Negri, and Marco Turchi. 2021.
\newblock \href {https://doi.org/10.18653/v1/2021.acl-long.224} {Cascade versus direct speech translation: Do the differences still make a difference?}
\newblock In \emph{Proceedings of the 59th Annual Meeting of the Association for Computational Linguistics and the 11th International Joint Conference on Natural Language Processing (Volume 1: Long Papers)}, pages 2873--2887, Online. Association for Computational Linguistics.

\bibitem[{B{\'e}rard et~al.(2016)B{\'e}rard, Pietquin, Besacier, and Servan}]{Berard2016ListenAT}
Alexandre B{\'e}rard, Olivier Pietquin, Laurent Besacier, and Christophe Servan. 2016.
\newblock Listen and translate: A proof of concept for end-to-end speech-to-text translation.
\newblock In \emph{NIPS Workshop on end-to-end learning for speech and audio processing}.

\bibitem[{Bizzoni et~al.(2020)Bizzoni, Juzek, Espa{\~n}a-Bonet, Dutta~Chowdhury, van Genabith, and Teich}]{bizzoni-etal-2020-human}
Yuri Bizzoni, Tom~S Juzek, Cristina Espa{\~n}a-Bonet, Koel Dutta~Chowdhury, Josef van Genabith, and Elke Teich. 2020.
\newblock \href {https://doi.org/10.18653/v1/2020.iwslt-1.34} {How human is machine translationese? comparing human and machine translations of text and speech}.
\newblock In \emph{Proceedings of the 17th International Conference on Spoken Language Translation}, pages 280--290, Online. Association for Computational Linguistics.

\bibitem[{Bojar et~al.(2021)Bojar, Srde{\v{c}}n{\'y}, Kumar, Smr{\v{z}}, Schneider, Haddow, Williams, and Canton}]{bojar-etal-2021-operating}
Ond{\v{r}}ej Bojar, Vojt{\v{e}}ch Srde{\v{c}}n{\'y}, Rishu Kumar, Otakar Smr{\v{z}}, Felix Schneider, Barry Haddow, Phil Williams, and Chiara Canton. 2021.
\newblock \href {https://aclanthology.org/2021.mtsummit-asltrw.3} {Operating a complex {SLT} system with speakers and human interpreters}.
\newblock In \emph{Proceedings of the 1st Workshop on Automatic Spoken Language Translation in Real-World Settings (ASLTRW)}, pages 23--34, Virtual. Association for Machine Translation in the Americas.

\bibitem[{Brown et~al.(2020)Brown, Mann, Ryder, Subbiah, Kaplan, Dhariwal, Neelakantan, Shyam, Sastry, Askell, Agarwal, Herbert-Voss, Krueger, Henighan, Child, Ramesh, Ziegler, Wu, Winter, Hesse, Chen, Sigler, Litwin, Gray, Chess, Clark, Berner, McCandlish, Radford, Sutskever, and Amodei}]{NEURIPS2020_1457c0d6}
Tom Brown, Benjamin Mann, Nick Ryder, Melanie Subbiah, Jared~D Kaplan, Prafulla Dhariwal, Arvind Neelakantan, Pranav Shyam, Girish Sastry, Amanda Askell, Sandhini Agarwal, Ariel Herbert-Voss, Gretchen Krueger, Tom Henighan, Rewon Child, Aditya Ramesh, Daniel Ziegler, Jeffrey Wu, Clemens Winter, Chris Hesse, Mark Chen, Eric Sigler, Mateusz Litwin, Scott Gray, Benjamin Chess, Jack Clark, Christopher Berner, Sam McCandlish, Alec Radford, Ilya Sutskever, and Dario Amodei. 2020.
\newblock \href {https://proceedings.neurips.cc/paper_files/paper/2020/file/1457c0d6bfcb4967418bfb8ac142f64a-Paper.pdf} {Language models are few-shot learners}.
\newblock In \emph{Advances in Neural Information Processing Systems}, volume~33, pages 1877--1901. Curran Associates, Inc.

\bibitem[{Casacuberta et~al.(2001)Casacuberta, Llorens, Martinez, Molau, Nevado, Ney, Pastor, Pico, Sanchis, Vidal, and Vilar}]{940906}
Francisco Casacuberta, David Llorens, Carlos Martinez, Sirko Molau, Francisco Nevado, Hermann Ney, Moisés Pastor, David Pico, Alberto Sanchis, Enrique Vidal, and Juan~M. Vilar. 2001.
\newblock \href {https://doi.org/10.1109/ICASSP.2001.940906} {Speech-to-speech translation based on finite-state transducers}.
\newblock In \emph{2001 IEEE International Conference on Acoustics, Speech, and Signal Processing. Proceedings (Cat. No.01CH37221)}, volume~1, pages 613--616 vol.1.

\bibitem[{Chang and yi~Lee(2022)}]{chang22f_interspeech}
Chih-Chiang Chang and Hung yi~Lee. 2022.
\newblock \href {https://doi.org/10.21437/Interspeech.2022-10627} {{Exploring Continuous Integrate-and-Fire for Adaptive Simultaneous Speech Translation}}.
\newblock In \emph{Proc. Interspeech 2022}, pages 5175--5179.

\bibitem[{Chen et~al.(2022)Chen, Hou, Hu, Shirol, and Chng}]{9747755}
Chen Chen, Nana Hou, Yuchen Hu, Shashank Shirol, and Eng~Siong Chng. 2022.
\newblock \href {https://doi.org/10.1109/ICASSP43922.2022.9747755} {Noise-robust speech recognition with 10 minutes unparalleled in-domain data}.
\newblock In \emph{ICASSP 2022 - 2022 IEEE International Conference on Acoustics, Speech and Signal Processing (ICASSP)}, pages 4298--4302.

\bibitem[{Chen et~al.(2021)Chen, Ma, Zheng, and Huang}]{chen-etal-2021-direct}
Junkun Chen, Mingbo Ma, Renjie Zheng, and Liang Huang. 2021.
\newblock \href {https://doi.org/10.18653/v1/2021.findings-acl.406} {Direct simultaneous speech-to-text translation assisted by synchronized streaming {ASR}}.
\newblock In \emph{Findings of the Association for Computational Linguistics: ACL-IJCNLP 2021}, pages 4618--4624, Online. Association for Computational Linguistics.

\bibitem[{Chen et~al.(2023)Chen, Xue, Wang, Pan, and Li}]{10389709}
Junkun Chen, Jian Xue, Peidong Wang, Jing Pan, and Jinyu Li. 2023.
\newblock \href {https://doi.org/10.1109/ASRU57964.2023.10389709} {Improving stability in simultaneous speech translation: A revision-controllable decoding approach}.
\newblock In \emph{2023 IEEE Automatic Speech Recognition and Understanding Workshop (ASRU)}, pages 1--7.

\bibitem[{Chen et~al.(2024)Chen, Fan, Luo, Zhang, Zhao, Liu, and Huang}]{Chen_Fan_Luo_Zhang_Zhao_Liu_Huang_2024}
Xinjie Chen, Kai Fan, Wei Luo, Linlin Zhang, Libo Zhao, Xinggao Liu, and Zhongqiang Huang. 2024.
\newblock \href {https://doi.org/10.1609/aaai.v38i16.29733} {Divergence-guided simultaneous speech translation}.
\newblock \emph{Proceedings of the AAAI Conference on Artificial Intelligence}, 38(16):17799--17807.

\bibitem[{Cherry and Foster(2019)}]{cherry2019thinking}
Colin Cherry and George Foster. 2019.
\newblock Thinking slow about latency evaluation for simultaneous machine translation.
\newblock \emph{arXiv preprint arXiv:1906.00048}.

\bibitem[{Chiu et~al.(2019)Chiu, Han, Zhang, Pang, Kishchenko, Nguyen, Narayanan, Liao, Zhang, Kannan, Prabhavalkar, Chen, Sainath, and Wu}]{chiu2019comparison}
Chung-Cheng Chiu, Wei Han, Yu~Zhang, Ruoming Pang, Sergey Kishchenko, Patrick Nguyen, Arun Narayanan, Hank Liao, Shuyuan Zhang, Anjuli Kannan, Rohit Prabhavalkar, Zhifeng Chen, Tara Sainath, and Yonghui Wu. 2019.
\newblock \href {https://doi.org/10.1109/ASRU46091.2019.9003854} {A comparison of end-to-end models for long-form speech recognition}.
\newblock In \emph{2019 IEEE Automatic Speech Recognition and Understanding Workshop (ASRU)}, pages 889--896.

\bibitem[{Cho et~al.(2013)Cho, Fügen, Hermann, Kilgour, Mediani, Mohr, Niehues, Rottmann, Saam, Stüker, and Waibel}]{cho13_interspeech}
Eunah Cho, Christian Fügen, Teresa Hermann, Kevin Kilgour, Mohammed Mediani, Christian Mohr, Jan Niehues, Kay Rottmann, Christian Saam, Sebastian Stüker, and Alex Waibel. 2013.
\newblock \href {https://doi.org/10.21437/Interspeech.2013-612} {{A real-world system for simultaneous translation of German lectures}}.
\newblock In \emph{Proc. Interspeech 2013}, pages 3473--3477.

\bibitem[{Cho et~al.(2015)Cho, Niehues, Kilgour, and Waibel}]{cho-etal-2015-punctuation}
Eunah Cho, Jan Niehues, Kevin Kilgour, and Alex Waibel. 2015.
\newblock \href {https://aclanthology.org/2015.iwslt-papers.8} {Punctuation insertion for real-time spoken language translation}.
\newblock In \emph{Proceedings of the 12th International Workshop on Spoken Language Translation: Papers}, pages 173--179, Da Nang, Vietnam.

\bibitem[{Cho et~al.(2017)Cho, Niehues, and Waibel}]{cho17_interspeech}
Eunah Cho, Jan Niehues, and Alex Waibel. 2017.
\newblock \href {https://doi.org/10.21437/Interspeech.2017-1320} {{NMT-Based Segmentation and Punctuation Insertion for Real-Time Spoken Language Translation}}.
\newblock In \emph{Proc. Interspeech 2017}, pages 2645--2649.

\bibitem[{Cho and Esipova(2016)}]{cho2016can}
Kyunghyun Cho and Masha Esipova. 2016.
\newblock Can neural machine translation do simultaneous translation?
\newblock \emph{arXiv preprint arXiv:1606.02012}.

\bibitem[{Dai et~al.(2019)Dai, Yang, Yang, Carbonell, Le, and Salakhutdinov}]{dai-etal-2019-transformer}
Zihang Dai, Zhilin Yang, Yiming Yang, Jaime Carbonell, Quoc Le, and Ruslan Salakhutdinov. 2019.
\newblock \href {https://doi.org/10.18653/v1/P19-1285} {Transformer-{XL}: Attentive language models beyond a fixed-length context}.
\newblock In \emph{Proceedings of the 57th Annual Meeting of the Association for Computational Linguistics}, pages 2978--2988, Florence, Italy. Association for Computational Linguistics.

\bibitem[{Dalvi et~al.(2018)Dalvi, Durrani, Sajjad, and Vogel}]{dalvi-etal-2018-incremental}
Fahim Dalvi, Nadir Durrani, Hassan Sajjad, and Stephan Vogel. 2018.
\newblock \href {https://doi.org/10.18653/v1/N18-2079} {Incremental decoding and training methods for simultaneous translation in neural machine translation}.
\newblock In \emph{Proceedings of the 2018 Conference of the North {A}merican Chapter of the Association for Computational Linguistics: Human Language Technologies, Volume 2 (Short Papers)}, pages 493--499, New Orleans, Louisiana. Association for Computational Linguistics.

\bibitem[{Deng et~al.(2022)Deng, Watanabe, Shi, and Arora}]{deng22b_interspeech}
Keqi Deng, Shinji Watanabe, Jiatong Shi, and Siddhant Arora. 2022.
\newblock \href {https://doi.org/10.21437/Interspeech.2022-933} {{Blockwise Streaming Transformer for Spoken Language Understanding and Simultaneous Speech Translation}}.
\newblock In \emph{Proc. Interspeech 2022}, pages 1746--1750.

\bibitem[{Deng and Woodland(2024)}]{deng2024label}
Keqi Deng and Phil Woodland. 2024.
\newblock \href {https://doi.org/10.18653/v1/2024.acl-long.448} {Label-synchronous neural transducer for {E}2{E} simultaneous speech translation}.
\newblock In \emph{Proceedings of the 62nd Annual Meeting of the Association for Computational Linguistics (Volume 1: Long Papers)}, pages 8235--8251, Bangkok, Thailand. Association for Computational Linguistics.

\bibitem[{Dessloch et~al.(2018)Dessloch, Ha, M{\"u}ller, Niehues, Nguyen, Pham, Salesky, Sperber, St{\"u}ker, Zenkel, and Waibel}]{dessloch-etal-2018-kit}
Florian Dessloch, Thanh-Le Ha, Markus M{\"u}ller, Jan Niehues, Thai-Son Nguyen, Ngoc-Quan Pham, Elizabeth Salesky, Matthias Sperber, Sebastian St{\"u}ker, Thomas Zenkel, and Alexander Waibel. 2018.
\newblock \href {https://aclanthology.org/C18-2020} {{KIT} lecture translator: Multilingual speech translation with one-shot learning}.
\newblock In \emph{Proceedings of the 27th International Conference on Computational Linguistics: System Demonstrations}, pages 89--93, Santa Fe, New Mexico. Association for Computational Linguistics.

\bibitem[{Donato et~al.(2021)Donato, Yu, and Dyer}]{donato-etal-2021-diverse}
Domenic Donato, Lei Yu, and Chris Dyer. 2021.
\newblock \href {https://doi.org/10.18653/v1/2021.acl-long.104} {Diverse pretrained context encodings improve document translation}.
\newblock In \emph{Proceedings of the 59th Annual Meeting of the Association for Computational Linguistics and the 11th International Joint Conference on Natural Language Processing (Volume 1: Long Papers)}, pages 1299--1311, Online. Association for Computational Linguistics.

\bibitem[{Dong et~al.(2022)Dong, Zhu, Wang, and Li}]{dong-etal-2022-learning}
Qian Dong, Yaoming Zhu, Mingxuan Wang, and Lei Li. 2022.
\newblock \href {https://doi.org/10.18653/v1/2022.acl-long.50} {Learning when to translate for streaming speech}.
\newblock In \emph{Proceedings of the 60th Annual Meeting of the Association for Computational Linguistics (Volume 1: Long Papers)}, pages 680--694, Dublin, Ireland. Association for Computational Linguistics.

\bibitem[{Elbayad et~al.(2020{\natexlab{a}})Elbayad, Besacier, and Verbeek}]{elbayad20_interspeech}
Maha Elbayad, Laurent Besacier, and Jakob Verbeek. 2020{\natexlab{a}}.
\newblock \href {https://doi.org/10.21437/Interspeech.2020-1241} {{Efficient Wait-k Models for Simultaneous Machine Translation}}.
\newblock In \emph{Proc. Interspeech 2020}, pages 1461--1465.

\bibitem[{Elbayad et~al.(2020{\natexlab{b}})Elbayad, Nguyen, Bougares, Tomashenko, Caubri{\`e}re, Lecouteux, Est{\`e}ve, and Besacier}]{elbayad-etal-2020-trac}
Maha Elbayad, Ha~Nguyen, Fethi Bougares, Natalia Tomashenko, Antoine Caubri{\`e}re, Benjamin Lecouteux, Yannick Est{\`e}ve, and Laurent Besacier. 2020{\natexlab{b}}.
\newblock \href {https://doi.org/10.18653/v1/2020.iwslt-1.2} {{ON}-{TRAC} consortium for end-to-end and simultaneous speech translation challenge tasks at {IWSLT} 2020}.
\newblock In \emph{Proceedings of the 17th International Conference on Spoken Language Translation}, pages 35--43, Online. Association for Computational Linguistics.

\bibitem[{Fantinuoli and Prandi(2021)}]{fantinuoli-prandi-2021-towards}
Claudio Fantinuoli and Bianca Prandi. 2021.
\newblock \href {https://doi.org/10.18653/v1/2021.iwslt-1.29} {Towards the evaluation of automatic simultaneous speech translation from a communicative perspective}.
\newblock In \emph{Proceedings of the 18th International Conference on Spoken Language Translation (IWSLT 2021)}, pages 245--254, Bangkok, Thailand (online). Association for Computational Linguistics.

\bibitem[{Fantinuoli and Wang(2024)}]{wang2024exploring}
Claudio Fantinuoli and Xiaoman Wang. 2024.
\newblock \href {https://aclanthology.org/2024.eamt-1.28} {Exploring the correlation between human and machine evaluation of simultaneous speech translation}.
\newblock In \emph{Proceedings of the 25th Annual Conference of the European Association for Machine Translation (Volume 1)}, pages 327--336, Sheffield, UK. European Association for Machine Translation (EAMT).

\bibitem[{Fernandes et~al.(2021)Fernandes, Yin, Neubig, and Martins}]{fernandes-etal-2021-measuring}
Patrick Fernandes, Kayo Yin, Graham Neubig, and Andr{\'e} F.~T. Martins. 2021.
\newblock \href {https://doi.org/10.18653/v1/2021.acl-long.505} {Measuring and increasing context usage in context-aware machine translation}.
\newblock In \emph{Proceedings of the 59th Annual Meeting of the Association for Computational Linguistics and the 11th International Joint Conference on Natural Language Processing (Volume 1: Long Papers)}, pages 6467--6478, Online. Association for Computational Linguistics.

\bibitem[{Ferrer et~al.(2003)Ferrer, Shriberg, and Stolcke}]{ferrer2003prosody}
Luciana Ferrer, Elizabeth Shriberg, and Andreas Stolcke. 2003.
\newblock \href {https://doi.org/10.1109/ICASSP.2003.1198854} {A prosody-based approach to end-of-utterance detection that does not require speech recognition}.
\newblock In \emph{2003 IEEE International Conference on Acoustics, Speech, and Signal Processing, 2003 (ICASSP'03).} IEEE.

\bibitem[{Fu et~al.(2023)Fu, Liao, Fan, Huang, Chen, Chen, and Shi}]{fu-etal-2023-adapting}
Biao Fu, Minpeng Liao, Kai Fan, Zhongqiang Huang, Boxing Chen, Yidong Chen, and Xiaodong Shi. 2023.
\newblock \href {https://doi.org/10.18653/v1/2023.emnlp-main.1033} {Adapting offline speech translation models for streaming with future-aware distillation and inference}.
\newblock In \emph{Proceedings of the 2023 Conference on Empirical Methods in Natural Language Processing}, pages 16600--16619, Singapore. Association for Computational Linguistics.

\bibitem[{F{\"u}gen et~al.(2006{\natexlab{a}})F{\"u}gen, Kolss, Bernreuther, Paulik, Stuker, Vogel, and Waibel}]{1660084}
Christian F{\"u}gen, Muntsin Kolss, Dietmar Bernreuther, Matthias Paulik, Sebastian Stuker, Stephan Vogel, and Alex Waibel. 2006{\natexlab{a}}.
\newblock \href {https://doi.org/10.1109/ICASSP.2006.1660084} {Open domain speech recognition \& translation:lectures and speeches}.
\newblock In \emph{2006 IEEE International Conference on Acoustics Speech and Signal Processing Proceedings}.

\bibitem[{F{\"u}gen et~al.(2006{\natexlab{b}})F{\"u}gen, Kolss, Paulik, and Waibel}]{fugen2006open}
Christian F{\"u}gen, Muntsin Kolss, Matthias Paulik, and Alex Waibel. 2006{\natexlab{b}}.
\newblock Open domain speech translation: from seminars and speeches to lectures.
\newblock In \emph{TC-STAR workshop on speech to speech translation, Barcelona, Spain}, pages 81--86.

\bibitem[{F{\"u}gen et~al.(2007)F{\"u}gen, Waibel, and Kolss}]{fugen2007simultaneous}
Christian F{\"u}gen, Alex Waibel, and Muntsin Kolss. 2007.
\newblock Simultaneous translation of lectures and speeches.
\newblock \emph{Machine translation}, 21:209--252.

\bibitem[{Fujita et~al.(2013)Fujita, Neubig, Sakti, Toda, and Nakamura}]{fujita13_interspeech}
Tomoki Fujita, Graham Neubig, Sakriani Sakti, Tomoki Toda, and Satoshi Nakamura. 2013.
\newblock \href {https://doi.org/10.21437/Interspeech.2013-615} {{Simple, lexicalized choice of translation timing for simultaneous speech translation}}.
\newblock In \emph{Proc. Interspeech 2013}, pages 3487--3491.

\bibitem[{Fukuda et~al.(2022{\natexlab{a}})Fukuda, Ko, Kano, Doi, Tokuyama, Sakti, Sudoh, and Nakamura}]{fukuda-etal-2022-naist}
Ryo Fukuda, Yuka Ko, Yasumasa Kano, Kosuke Doi, Hirotaka Tokuyama, Sakriani Sakti, Katsuhito Sudoh, and Satoshi Nakamura. 2022{\natexlab{a}}.
\newblock \href {https://doi.org/10.18653/v1/2022.iwslt-1.25} {{NAIST} simultaneous speech-to-text translation system for {IWSLT} 2022}.
\newblock In \emph{Proceedings of the 19th International Conference on Spoken Language Translation (IWSLT 2022)}, pages 286--292, Dublin, Ireland (in-person and online). Association for Computational Linguistics.

\bibitem[{Fukuda et~al.(2023)Fukuda, Nishikawa, Kano, Ko, Yanagita, Doi, Makinae, Sakti, Sudoh, and Nakamura}]{fukuda-etal-2023-naist}
Ryo Fukuda, Yuta Nishikawa, Yasumasa Kano, Yuka Ko, Tomoya Yanagita, Kosuke Doi, Mana Makinae, Sakriani Sakti, Katsuhito Sudoh, and Satoshi Nakamura. 2023.
\newblock \href {https://doi.org/10.18653/v1/2023.iwslt-1.31} {{NAIST} simultaneous speech-to-speech translation system for {IWSLT} 2023}.
\newblock In \emph{Proceedings of the 20th International Conference on Spoken Language Translation (IWSLT 2023)}, pages 330--340, Toronto, Canada (in-person and online). Association for Computational Linguistics.

\bibitem[{Fukuda et~al.(2022{\natexlab{b}})Fukuda, Sudoh, and Nakamura}]{fukuda22b_interspeech}
Ryo Fukuda, Katsuhito Sudoh, and Satoshi Nakamura. 2022{\natexlab{b}}.
\newblock \href {https://doi.org/10.21437/Interspeech.2022-11382} {{Speech Segmentation Optimization using Segmented Bilingual Speech Corpus for End-to-end Speech Translation}}.
\newblock In \emph{Proc. Interspeech 2022}, pages 121--125.

\bibitem[{Fügen(2009)}]{Fuegen2009_1000013594}
Christian Fügen. 2009.
\newblock \href {https://doi.org/10.5445/IR/1000013594} {\emph{A System for Simultaneous Translation of Lectures and Speeches}}.
\newblock Ph.D. thesis, {Universität Karlsruhe (TH)}.

\bibitem[{Gaido et~al.(2020)Gaido, Gangi, Negri, Cettolo, and Turchi}]{gaido20_interspeech}
Marco Gaido, Mattia A.~Di Gangi, Matteo Negri, Mauro Cettolo, and Marco Turchi. 2020.
\newblock \href {https://doi.org/10.21437/Interspeech.2020-2860} {{Contextualized Translation of Automatically Segmented Speech}}.
\newblock In \emph{Proc. Interspeech 2020}, pages 1471--1475.

\bibitem[{Gaido et~al.(2021)Gaido, Negri, Cettolo, and Turchi}]{gaido2021beyond}
Marco Gaido, Matteo Negri, Mauro Cettolo, and Marco Turchi. 2021.
\newblock \href {https://aclanthology.org/2021.icnlsp-1.7} {Beyond voice activity detection: Hybrid audio segmentation for direct speech translation}.
\newblock In \emph{Proceedings of the 4th International Conference on Natural Language and Speech Processing (ICNLSP 2021)}, pages 55--62, Trento, Italy. Association for Computational Linguistics.

\bibitem[{Gaido et~al.(2022)Gaido, Papi, Fucci, Fiameni, Negri, and Turchi}]{gaido-etal-2022-efficient}
Marco Gaido, Sara Papi, Dennis Fucci, Giuseppe Fiameni, Matteo Negri, and Marco Turchi. 2022.
\newblock \href {https://doi.org/10.18653/v1/2022.iwslt-1.13} {Efficient yet competitive speech translation: {FBK}@{IWSLT}2022}.
\newblock In \emph{Proceedings of the 19th International Conference on Spoken Language Translation (IWSLT 2022)}, pages 177--189, Dublin, Ireland (in-person and online). Association for Computational Linguistics.

\bibitem[{Gaido et~al.(2024)Gaido, Papi, Negri, and Bentivogli}]{gaido2024speech}
Marco Gaido, Sara Papi, Matteo Negri, and Luisa Bentivogli. 2024.
\newblock \href {https://doi.org/10.18653/v1/2024.acl-long.789} {Speech translation with speech foundation models and large language models: What is there and what is missing?}
\newblock In \emph{Proceedings of the 62nd Annual Meeting of the Association for Computational Linguistics (Volume 1: Long Papers)}, pages 14760--14778, Bangkok, Thailand. Association for Computational Linguistics.

\bibitem[{Gaido et~al.(2023)Gaido, Papi, Negri, and Turchi}]{gaido23_interspeech}
Marco Gaido, Sara Papi, Matteo Negri, and Marco Turchi. 2023.
\newblock \href {https://doi.org/10.21437/Interspeech.2023-1767} {{Joint Speech Translation and Named Entity Recognition}}.
\newblock In \emph{Proc. INTERSPEECH 2023}, pages 47--51.

\bibitem[{Gegenfurtner(2016)}]{doi:10.1177/0301006616657097}
Karl~R. Gegenfurtner. 2016.
\newblock \href {https://doi.org/10.1177/0301006616657097} {The interaction between vision and eye movements}.
\newblock \emph{Perception}, 45(12):1333--1357.
\newblock PMID: 27383394.

\bibitem[{Graves et~al.(2006)Graves, Fern\'{a}ndez, Gomez, and Schmidhuber}]{10.1145/1143844.1143891}
Alex Graves, Santiago Fern\'{a}ndez, Faustino Gomez, and J\"{u}rgen Schmidhuber. 2006.
\newblock \href {https://doi.org/10.1145/1143844.1143891} {Connectionist temporal classification: labelling unsegmented sequence data with recurrent neural networks}.
\newblock In \emph{Proceedings of the 23rd International Conference on Machine Learning}, ICML '06, page 369–376, New York, NY, USA. Association for Computing Machinery.

\bibitem[{Grissom~II et~al.(2014)Grissom~II, He, Boyd-Graber, Morgan, and Daum{\'e}~III}]{grissom-ii-etal-2014-dont}
Alvin Grissom~II, He~He, Jordan Boyd-Graber, John Morgan, and Hal Daum{\'e}~III. 2014.
\newblock \href {https://doi.org/10.3115/v1/D14-1140} {Don{'}t until the final verb wait: Reinforcement learning for simultaneous machine translation}.
\newblock In \emph{Proceedings of the 2014 Conference on Empirical Methods in Natural Language Processing ({EMNLP})}, pages 1342--1352, Doha, Qatar. Association for Computational Linguistics.

\bibitem[{Guo et~al.(2022)Guo, Liu, Zhang, Chen, Mu, Li, Cui, Wang, and Guo}]{guo-etal-2022-xiaomi}
Bao Guo, Mengge Liu, Wen Zhang, Hexuan Chen, Chang Mu, Xiang Li, Jianwei Cui, Bin Wang, and Yuhang Guo. 2022.
\newblock \href {https://doi.org/10.18653/v1/2022.iwslt-1.17} {The xiaomi text-to-text simultaneous speech translation system for {IWSLT} 2022}.
\newblock In \emph{Proceedings of the 19th International Conference on Spoken Language Translation (IWSLT 2022)}, pages 216--224, Dublin, Ireland (in-person and online). Association for Computational Linguistics.

\bibitem[{Guo et~al.(2023)Guo, Wei, Wu, Li, Rao, Wang, Shang, Chen, Yu, Li, Xie, Lei, and Yang}]{guo-etal-2023-hw}
Jiaxin Guo, Daimeng Wei, Zhanglin Wu, Zongyao Li, Zhiqiang Rao, Minghan Wang, Hengchao Shang, Xiaoyu Chen, Zhengzhe Yu, Shaojun Li, Yuhao Xie, Lizhi Lei, and Hao Yang. 2023.
\newblock \href {https://doi.org/10.18653/v1/2023.iwslt-1.35} {The {HW}-{TSC}{'}s simultaneous speech-to-text translation system for {IWSLT} 2023 evaluation}.
\newblock In \emph{Proceedings of the 20th International Conference on Spoken Language Translation (IWSLT 2023)}, pages 376--382, Toronto, Canada (in-person and online). Association for Computational Linguistics.

\bibitem[{Guo et~al.(2024)Guo, Wu, Li, Shang, Wei, Chen, Rao, Li, and Yang}]{guo2024r}
Jiaxin Guo, Zhanglin Wu, Zongyao Li, Hengchao Shang, Daimeng Wei, Xiaoyu Chen, Zhiqiang Rao, Shaojun Li, and Hao Yang. 2024.
\newblock R-bi: Regularized batched inputs enhance incremental decoding framework for low-latency simultaneous speech translation.
\newblock \emph{arXiv preprint arXiv:2401.05700}.

\bibitem[{Han et~al.(2020)Han, Zaidi, Indurthi, Lakumarapu, Lee, and Kim}]{han-etal-2020-end}
Hou~Jeung Han, Mohd~Abbas Zaidi, Sathish~Reddy Indurthi, Nikhil~Kumar Lakumarapu, Beomseok Lee, and Sangha Kim. 2020.
\newblock \href {https://doi.org/10.18653/v1/2020.iwslt-1.5} {End-to-end simultaneous translation system for {IWSLT}2020 using modality agnostic meta-learning}.
\newblock In \emph{Proceedings of the 17th International Conference on Spoken Language Translation}, pages 62--68, Online. Association for Computational Linguistics.

\bibitem[{Huang et~al.(2022)Huang, Chang, Rybach, Sainath, Prabhavalkar, Peyser, Lu, and Allauzen}]{huang2022e2e}
W.~Ronny Huang, Shuo-Yiin Chang, David Rybach, Tara Sainath, Rohit Prabhavalkar, Cal Peyser, Zhiyun Lu, and Cyril Allauzen. 2022.
\newblock \href {https://doi.org/10.21437/Interspeech.2022-38} {E2e segmenter: Joint segmenting and decoding for long-form asr}.
\newblock In \emph{Interspeech 2022}, pages 4995--4999.

\bibitem[{Huang et~al.(2023)Huang, Liu, Li, Tian, Yang, Zhang, Luan, Wang, Guo, and Su}]{huang-etal-2023-xiaomi}
Wuwei Huang, Mengge Liu, Xiang Li, Yanzhi Tian, Fengyu Yang, Wen Zhang, Jian Luan, Bin Wang, Yuhang Guo, and Jinsong Su. 2023.
\newblock \href {https://doi.org/10.18653/v1/2023.iwslt-1.39} {The xiaomi {AI} lab{'}s speech translation systems for {IWSLT} 2023 offline task, simultaneous task and speech-to-speech task}.
\newblock In \emph{Proceedings of the 20th International Conference on Spoken Language Translation (IWSLT 2023)}, pages 411--419, Toronto, Canada (in-person and online). Association for Computational Linguistics.

\bibitem[{Huber et~al.(2023)Huber, Dinh, Mullov, Pham, Nguyen, Retkowski, Constantin, Ugan, Liu, Li, Koneru, Niehues, and Waibel}]{huber-etal-2023-end}
Christian Huber, Tu~Anh Dinh, Carlos Mullov, Ngoc-Quan Pham, Thai~Binh Nguyen, Fabian Retkowski, Stefan Constantin, Enes Ugan, Danni Liu, Zhaolin Li, Sai Koneru, Jan Niehues, and Alexander Waibel. 2023.
\newblock \href {https://doi.org/10.18653/v1/2023.emnlp-demo.2} {End-to-end evaluation for low-latency simultaneous speech translation}.
\newblock In \emph{Proceedings of the 2023 Conference on Empirical Methods in Natural Language Processing: System Demonstrations}, pages 12--20, Singapore. Association for Computational Linguistics.

\bibitem[{Huber et~al.(2022)Huber, Ugan, and Waibel}]{huber2022code}
Christian Huber, Enes~Yavuz Ugan, and Alexander Waibel. 2022.
\newblock Code-switching without switching: Language agnostic end-to-end speech translation.
\newblock \emph{arXiv preprint arXiv:2210.01512}.

\bibitem[{Hwang et~al.(2024)Hwang, Kulikov, Peloquin, Gong, Chen, and Lee}]{hwang2024textless}
Min-Jae Hwang, Ilia Kulikov, Benjamin Peloquin, Hongyu Gong, Peng-Jen Chen, and Ann Lee. 2024.
\newblock \href {https://doi.org/10.18653/v1/2024.findings-acl.917} {Textless acoustic model with self-supervised distillation for noise-robust expressive speech-to-speech translation}.
\newblock In \emph{Findings of the Association for Computational Linguistics ACL 2024}, pages 15524--15541, Bangkok, Thailand and virtual meeting. Association for Computational Linguistics.

\bibitem[{Inaguma et~al.(2021)Inaguma, Yan, Dalmia, Guo, Shi, Duh, and Watanabe}]{inaguma-etal-2021-espnet}
Hirofumi Inaguma, Brian Yan, Siddharth Dalmia, Pengcheng Guo, Jiatong Shi, Kevin Duh, and Shinji Watanabe. 2021.
\newblock \href {https://doi.org/10.18653/v1/2021.iwslt-1.10} {{ESP}net-{ST} {IWSLT} 2021 offline speech translation system}.
\newblock In \emph{Proceedings of the 18th International Conference on Spoken Language Translation (IWSLT 2021)}, pages 100--109, Bangkok, Thailand (online). Association for Computational Linguistics.

\bibitem[{Indurthi et~al.(2022)Indurthi, Zaidi, Lee, Lakumarapu, and Kim}]{indurthi-etal-2022-language}
Sathish~Reddy Indurthi, Mohd~Abbas Zaidi, Beomseok Lee, Nikhil~Kumar Lakumarapu, and Sangha Kim. 2022.
\newblock \href {https://doi.org/10.18653/v1/2022.naacl-main.3} {Language model augmented monotonic attention for simultaneous translation}.
\newblock In \emph{Proceedings of the 2022 Conference of the North American Chapter of the Association for Computational Linguistics: Human Language Technologies}, pages 38--45, Seattle, United States. Association for Computational Linguistics.

\bibitem[{Iranzo-S{\'a}nchez et~al.(2020)Iranzo-S{\'a}nchez, Gim{\'e}nez~Pastor, Silvestre-Cerd{\`a}, Baquero-Arnal, Civera~Saiz, and Juan}]{iranzo-sanchez-etal-2020-direct}
Javier Iranzo-S{\'a}nchez, Adri{\`a} Gim{\'e}nez~Pastor, Joan~Albert Silvestre-Cerd{\`a}, Pau Baquero-Arnal, Jorge Civera~Saiz, and Alfons Juan. 2020.
\newblock \href {https://doi.org/10.18653/v1/2020.emnlp-main.206} {Direct segmentation models for streaming speech translation}.
\newblock In \emph{Proceedings of the 2020 Conference on Empirical Methods in Natural Language Processing (EMNLP)}, pages 2599--2611, Online. Association for Computational Linguistics.

\bibitem[{Iranzo-S{\'a}nchez et~al.(2022)Iranzo-S{\'a}nchez, Jorge~Cano, P{\'e}rez-Gonz{\'a}lez-de Martos, Gim{\'e}nez~Pastor, Garc{\'e}s D{\'\i}az-Mun{\'\i}o, Baquero-Arnal, Silvestre-Cerd{\`a}, Civera~Saiz, Sanchis, and Juan}]{iranzo-sanchez-etal-2022-mllp}
Javier Iranzo-S{\'a}nchez, Javier Jorge~Cano, Alejandro P{\'e}rez-Gonz{\'a}lez-de Martos, Adri{\'a}n Gim{\'e}nez~Pastor, Gon{\c{c}}al Garc{\'e}s D{\'\i}az-Mun{\'\i}o, Pau Baquero-Arnal, Joan~Albert Silvestre-Cerd{\`a}, Jorge Civera~Saiz, Albert Sanchis, and Alfons Juan. 2022.
\newblock \href {https://doi.org/10.18653/v1/2022.iwslt-1.22} {{MLLP}-{VRAIN} {UPV} systems for the {IWSLT} 2022 simultaneous speech translation and speech-to-speech translation tasks}.
\newblock In \emph{Proceedings of the 19th International Conference on Spoken Language Translation (IWSLT 2022)}, pages 255--264, Dublin, Ireland (in-person and online). Association for Computational Linguistics.

\bibitem[{Iranzo-Sánchez et~al.(2024)Iranzo-Sánchez, Iranzo-Sánchez, Giménez, Civera, and Juan}]{iranzo2023segmentation}
Javier Iranzo-Sánchez, Jorge Iranzo-Sánchez, Adrià Giménez, Jorge Civera, and Alfons Juan. 2024.
\newblock \href {https://doi.org/10.1162/tacl_a_00691} {{Segmentation-Free Streaming Machine Translation}}.
\newblock \emph{Transactions of the Association for Computational Linguistics}, 12:1104--1121.

\bibitem[{Iranzo-Sánchez et~al.(2021)Iranzo-Sánchez, Jorge, Baquero-Arnal, Silvestre-Cerdà, Giménez, Civera, Sanchis, and Juan}]{IRANZOSANCHEZ2021303}
Javier Iranzo-Sánchez, Javier Jorge, Pau Baquero-Arnal, Joan~Albert Silvestre-Cerdà, Adrià Giménez, Jorge Civera, Albert Sanchis, and Alfons Juan. 2021.
\newblock \href {https://doi.org/https://doi.org/10.1016/j.neunet.2021.05.013} {Streaming cascade-based speech translation leveraged by a direct segmentation model}.
\newblock \emph{Neural Networks}, 142:303--315.

\bibitem[{Irvin(2021)}]{Irvin2021StudentIR}
Christopher Irvin. 2021.
\newblock Student insights related to the use of simultaneous speech translation for video lectures in a university english course.
\newblock \emph{STEM Journal}.

\bibitem[{Javorsk{\'y} et~al.(2022)Javorsk{\'y}, Mach{\'a}{\v{c}}ek, and Bojar}]{javorsky-etal-2022-continuous}
D{\'a}vid Javorsk{\'y}, Dominik Mach{\'a}{\v{c}}ek, and Ond{\v{r}}ej Bojar. 2022.
\newblock \href {https://aclanthology.org/2022.wmt-1.9} {Continuous rating as reliable human evaluation of simultaneous speech translation}.
\newblock In \emph{Proceedings of the Seventh Conference on Machine Translation (WMT)}, pages 154--164, Abu Dhabi, United Arab Emirates (Hybrid). Association for Computational Linguistics.

\bibitem[{Kano et~al.(2023)Kano, Sudoh, and Nakamura}]{kano23_interspeech}
Yasumasa Kano, Katsuhito Sudoh, and Satoshi Nakamura. 2023.
\newblock \href {https://doi.org/10.21437/Interspeech.2023-933} {{Average Token Delay: A Latency Metric for Simultaneous Translation}}.
\newblock In \emph{Proc. INTERSPEECH 2023}, pages 4469--4473.

\bibitem[{Karakanta et~al.(2020)Karakanta, Negri, and Turchi}]{karakanta-etal-2020-42}
Alina Karakanta, Matteo Negri, and Marco Turchi. 2020.
\newblock \href {https://doi.org/10.18653/v1/2020.iwslt-1.26} {Is 42 the answer to everything in subtitling-oriented speech translation?}
\newblock In \emph{Proceedings of the 17th International Conference on Spoken Language Translation}, pages 209--219, Online.

\bibitem[{Karakanta et~al.(2021)Karakanta, Papi, Negri, and Turchi}]{karakanta-etal-2021-simultaneous}
Alina Karakanta, Sara Papi, Matteo Negri, and Marco Turchi. 2021.
\newblock \href {https://aclanthology.org/2021.mtsummit-asltrw.4} {Simultaneous speech translation for live subtitling: from delay to display}.
\newblock In \emph{Proceedings of the 1st Workshop on Automatic Spoken Language Translation in Real-World Settings (ASLTRW)}, pages 35--48, Virtual. Association for Machine Translation in the Americas.

\bibitem[{Kim et~al.(2019)Kim, Tran, and Ney}]{Kim_Tran_Ney_2019}
Yunsu Kim, Duc~Thanh Tran, and Hermann Ney. 2019.
\newblock \href {https://doi.org/10.18653/v1/D19-6503} {When and why is document-level context useful in neural machine translation?}
\newblock In \emph{Proceedings of the Fourth Workshop on Discourse in Machine Translation (DiscoMT 2019)}, page 24–34, Hong Kong, China. Association for Computational Linguistics.

\bibitem[{Ko et~al.(2023)Ko, Fukuda, Nishikawa, Kano, Sudoh, and Nakamura}]{ko-etal-2023-tagged}
Yuka Ko, Ryo Fukuda, Yuta Nishikawa, Yasumasa Kano, Katsuhito Sudoh, and Satoshi Nakamura. 2023.
\newblock \href {https://doi.org/10.18653/v1/2023.iwslt-1.34} {Tagged end-to-end simultaneous speech translation training using simultaneous interpretation data}.
\newblock In \emph{Proceedings of the 20th International Conference on Spoken Language Translation (IWSLT 2023)}, pages 363--375, Toronto, Canada (in-person and online). Association for Computational Linguistics.

\bibitem[{Ko et~al.(2024)Ko, Fukuda, Nishikawa, Kano, Yanagita, Doi, Makinae, Tan, Sakai, Sakti, Sudoh, and Nakamura}]{ko2024naist}
Yuka Ko, Ryo Fukuda, Yuta Nishikawa, Yasumasa Kano, Tomoya Yanagita, Kosuke Doi, Mana Makinae, Haotian Tan, Makoto Sakai, Sakriani Sakti, Katsuhito Sudoh, and Satoshi Nakamura. 2024.
\newblock \href {https://doi.org/10.18653/v1/2024.iwslt-1.23} {{NAIST} simultaneous speech translation system for {IWSLT} 2024}.
\newblock In \emph{Proceedings of the 21st International Conference on Spoken Language Translation (IWSLT 2024)}, pages 170--182, Bangkok, Thailand (in-person and online). Association for Computational Linguistics.

\bibitem[{Kocmi et~al.(2024)Kocmi, Zouhar, Federmann, and Post}]{kocmi-etal-2024-navigating}
Tom Kocmi, Vil{\'e}m Zouhar, Christian Federmann, and Matt Post. 2024.
\newblock \href {https://aclanthology.org/2024.acl-long.110} {Navigating the metrics maze: Reconciling score magnitudes and accuracies}.
\newblock In \emph{Proceedings of the 62nd Annual Meeting of the Association for Computational Linguistics (Volume 1: Long Papers)}, pages 1999--2014, Bangkok, Thailand. Association for Computational Linguistics.

\bibitem[{Kolss et~al.(2008)Kolss, W{\"o}lfel, Kraft, Niehues, Paulik, and Waibel}]{kolss-etal-2008-simultaneous}
Muntsin Kolss, Matthias W{\"o}lfel, Florian Kraft, Jan Niehues, Matthias Paulik, and Alex Waibel. 2008.
\newblock \href {https://aclanthology.org/2008.iwslt-papers.5} {Simultaneous {G}erman-{E}nglish lecture translation.}
\newblock In \emph{Proceedings of the 5th International Workshop on Spoken Language Translation: Papers}, pages 174--181, Waikiki, Hawaii.

\bibitem[{Laplante(1992)}]{laplante1992real}
Phillip~A Laplante. 1992.
\newblock \emph{Real-time systems design and analysis: an engineer's handbook}.
\newblock IEEE press.

\bibitem[{Latif et~al.(2023)Latif, Shoukat, Shamshad, Usama, Cuay{\'a}huitl, and Schuller}]{latif2023sparks}
Siddique Latif, Moazzam Shoukat, Fahad Shamshad, Muhammad Usama, Heriberto Cuay{\'a}huitl, and Bj{\"o}rn~W Schuller. 2023.
\newblock Sparks of large audio models: A survey and outlook.
\newblock \emph{arXiv preprint arXiv:2308.12792}.

\bibitem[{Li et~al.(2022)Li, Sun, and Li}]{li-etal-2022-system}
Zecheng Li, Yue Sun, and Haoze Li. 2022.
\newblock \href {https://doi.org/10.18653/v1/2022.autosimtrans-1.3} {System description on automatic simultaneous translation workshop}.
\newblock In \emph{Proceedings of the Third Workshop on Automatic Simultaneous Translation}, pages 18--21, Online. Association for Computational Linguistics.

\bibitem[{Liu et~al.(2021{\natexlab{a}})Liu, Du, Li, Hu, and Dai}]{liu-etal-2021-ustc}
Dan Liu, Mengge Du, Xiaoxi Li, Yuchen Hu, and Lirong Dai. 2021{\natexlab{a}}.
\newblock \href {https://doi.org/10.18653/v1/2021.iwslt-1.2} {The {USTC}-{NELSLIP} systems for simultaneous speech translation task at {IWSLT} 2021}.
\newblock In \emph{Proceedings of the 18th International Conference on Spoken Language Translation (IWSLT 2021)}, pages 30--38, Bangkok, Thailand (online). Association for Computational Linguistics.

\bibitem[{Liu et~al.(2021{\natexlab{b}})Liu, Du, Li, Li, and Chen}]{liu-etal-2021-cross}
Dan Liu, Mengge Du, Xiaoxi Li, Ya~Li, and Enhong Chen. 2021{\natexlab{b}}.
\newblock \href {https://doi.org/10.18653/v1/2021.emnlp-main.4} {Cross attention augmented transducer networks for simultaneous translation}.
\newblock In \emph{Proceedings of the 2021 Conference on Empirical Methods in Natural Language Processing}, pages 39--55, Online and Punta Cana, Dominican Republic. Association for Computational Linguistics.

\bibitem[{Liu et~al.(2024)Liu, Hu, Du, He, Luo, Xu, Xiao, and Zhu}]{liu2024recent}
Xiaoqian Liu, Guoqiang Hu, Yangfan Du, Erfeng He, YingFeng Luo, Chen Xu, Tong Xiao, and Jingbo Zhu. 2024.
\newblock \href {https://doi.org/10.24963/ijcai.2024/900} {Recent advances in end-to-end simultaneous speech translation}.
\newblock In \emph{Proceedings of the Thirty-Third International Joint Conference on Artificial Intelligence, {IJCAI-24}}, pages 8142--8150. International Joint Conferences on Artificial Intelligence Organization.
\newblock Survey Track.

\bibitem[{Lu and Ng(2010)}]{lu2010better}
Wei Lu and Hwee~Tou Ng. 2010.
\newblock \href {https://aclanthology.org/D10-1018} {Better punctuation prediction with dynamic conditional random fields}.
\newblock In \emph{Proceedings of the 2010 Conference on Empirical Methods in Natural Language Processing}, pages 177--186, Cambridge, MA. Association for Computational Linguistics.

\bibitem[{Lu et~al.(2021)Lu, Pan, Doutre, Haghani, Cao, Prabhavalkar, Zhang, and Strohman}]{lu2021input}
Zhiyun Lu, Yanwei Pan, Thibault Doutre, Parisa Haghani, Liangliang Cao, Rohit Prabhavalkar, Chao Zhang, and Trevor Strohman. 2021.
\newblock Input length matters: Improving rnn-t and mwer training for long-form telephony speech recognition.
\newblock \emph{arXiv preprint arXiv:2110.03841}.

\bibitem[{Luong et~al.(2016)Luong, Le, Sutskever, Vinyals, and Kaiser}]{44928}
Thang Luong, Quoc~V. Le, Ilya Sutskever, Oriol Vinyals, and Lukasz Kaiser. 2016.
\newblock Multi-task sequence to sequence learning.
\newblock In \emph{International Conference on Learning Representations}.

\bibitem[{Lv and Liang(2019)}]{doi:10.1080/0907676X.2018.1498531}
Qianxi Lv and Junying Liang. 2019.
\newblock \href {https://doi.org/10.1080/0907676X.2018.1498531} {Is consecutive interpreting easier than simultaneous interpreting? – a corpus-based study of lexical simplification in interpretation}.
\newblock \emph{Perspectives}, 27(1):91--106.

\bibitem[{Ma et~al.(2019)Ma, Huang, Xiong, Zheng, Liu, Zheng, Zhang, He, Liu, Li, Wu, and Wang}]{ma-etal-2019-stacl}
Mingbo Ma, Liang Huang, Hao Xiong, Renjie Zheng, Kaibo Liu, Baigong Zheng, Chuanqiang Zhang, Zhongjun He, Hairong Liu, Xing Li, Hua Wu, and Haifeng Wang. 2019.
\newblock \href {https://doi.org/10.18653/v1/P19-1289} {{STACL}: Simultaneous translation with implicit anticipation and controllable latency using prefix-to-prefix framework}.
\newblock In \emph{Proceedings of the 57th Annual Meeting of the Association for Computational Linguistics}, pages 3025--3036, Florence, Italy. Association for Computational Linguistics.

\bibitem[{Ma et~al.(2020{\natexlab{a}})Ma, Dousti, Wang, Gu, and Pino}]{ma-etal-2020-simuleval}
Xutai Ma, Mohammad~Javad Dousti, Changhan Wang, Jiatao Gu, and Juan Pino. 2020{\natexlab{a}}.
\newblock \href {https://doi.org/10.18653/v1/2020.emnlp-demos.19} {{SIMULEVAL}: An evaluation toolkit for simultaneous translation}.
\newblock In \emph{Proceedings of the 2020 Conference on Empirical Methods in Natural Language Processing: System Demonstrations}, pages 144--150, Online. Association for Computational Linguistics.

\bibitem[{Ma et~al.(2020{\natexlab{b}})Ma, Pino, and Koehn}]{ma-etal-2020-simulmt}
Xutai Ma, Juan Pino, and Philipp Koehn. 2020{\natexlab{b}}.
\newblock \href {https://aclanthology.org/2020.aacl-main.58} {{S}imul{MT} to {S}imul{ST}: Adapting simultaneous text translation to end-to-end simultaneous speech translation}.
\newblock In \emph{Proceedings of the 1st Conference of the Asia-Pacific Chapter of the Association for Computational Linguistics and the 10th International Joint Conference on Natural Language Processing}, pages 582--587, Suzhou, China. Association for Computational Linguistics.

\bibitem[{Ma et~al.(2023)Ma, Sun, Ouyang, Inaguma, and Tomasello}]{ma2023efficient}
Xutai Ma, Anna Sun, Siqi Ouyang, Hirofumi Inaguma, and Paden Tomasello. 2023.
\newblock Efficient monotonic multihead attention.
\newblock \emph{arXiv preprint arXiv:2312.04515}.

\bibitem[{Ma et~al.(2021)Ma, Wang, Dousti, Koehn, and Pino}]{9414897}
Xutai Ma, Yongqiang Wang, Mohammad~Javad Dousti, Philipp Koehn, and Juan Pino. 2021.
\newblock \href {https://doi.org/10.1109/ICASSP39728.2021.9414897} {Streaming simultaneous speech translation with augmented memory transformer}.
\newblock In \emph{ICASSP 2021 - 2021 IEEE International Conference on Acoustics, Speech and Signal Processing (ICASSP)}, pages 7523--7527.

\bibitem[{Ma et~al.(2024)Ma, Fang, Zhang, Guo, Feng, and Zhang}]{ma2024non}
Zhengrui Ma, Qingkai Fang, Shaolei Zhang, Shoutao Guo, Yang Feng, and Min Zhang. 2024.
\newblock \href {https://doi.org/10.18653/v1/2024.acl-long.85} {A non-autoregressive generation framework for end-to-end simultaneous speech-to-any translation}.
\newblock In \emph{Proceedings of the 62nd Annual Meeting of the Association for Computational Linguistics (Volume 1: Long Papers)}, pages 1557--1575, Bangkok, Thailand. Association for Computational Linguistics.

\bibitem[{Mach{\'a}{\v{c}}ek et~al.(2023)Mach{\'a}{\v{c}}ek, Bojar, and Dabre}]{machacek-etal-2023-mt}
Dominik Mach{\'a}{\v{c}}ek, Ond{\v{r}}ej Bojar, and Raj Dabre. 2023.
\newblock \href {https://doi.org/10.18653/v1/2023.iwslt-1.12} {{MT} metrics correlate with human ratings of simultaneous speech translation}.
\newblock In \emph{Proceedings of the 20th International Conference on Spoken Language Translation (IWSLT 2023)}, pages 169--179, Toronto, Canada (in-person and online). Association for Computational Linguistics.

\bibitem[{Mach{\'a}{\v{c}}ek et~al.(2020)Mach{\'a}{\v{c}}ek, Kratochv{\'\i}l, Sagar, {\v{Z}}ilinec, Bojar, Nguyen, Schneider, Williams, and Yao}]{machacek-etal-2020-elitr}
Dominik Mach{\'a}{\v{c}}ek, Jon{\'a}{\v{s}} Kratochv{\'\i}l, Sangeet Sagar, Mat{\'u}{\v{s}} {\v{Z}}ilinec, Ond{\v{r}}ej Bojar, Thai-Son Nguyen, Felix Schneider, Philip Williams, and Yuekun Yao. 2020.
\newblock \href {https://doi.org/10.18653/v1/2020.iwslt-1.25} {{ELITR} non-native speech translation at {IWSLT} 2020}.
\newblock In \emph{Proceedings of the 17th International Conference on Spoken Language Translation}, pages 200--208, Online. Association for Computational Linguistics.

\bibitem[{Macháček and Bojar(2020)}]{machavcek2020presenting}
Dominik Macháček and Ondřej Bojar. 2020.
\newblock Presenting simultaneous translation in limited space.
\newblock In \emph{Proceedings of the 20th Conference Information Technologies - Applications and Theory (ITAT 2020)}, pages 32--37, Košice, Slovakia. Tomáš Horváth.

\bibitem[{Matusov et~al.(2006)Matusov, Kanthak, and Ney}]{1661501}
Evgeny Matusov, Stephan Kanthak, and Hermann Ney. 2006.
\newblock \href {https://doi.org/10.1109/ICASSP.2006.1661501} {Integrating speech recognition and machine translation: Where do we stand?}
\newblock In \emph{2006 IEEE International Conference on Acoustics Speech and Signal Processing Proceedings}, volume~5, pages V--V.

\bibitem[{Matusov et~al.(2005)Matusov, Leusch, Bender, and Ney}]{matusov-etal-2005-evaluating}
Evgeny Matusov, Gregor Leusch, Oliver Bender, and Hermann Ney. 2005.
\newblock \href {https://aclanthology.org/2005.iwslt-1.19} {Evaluating machine translation output with automatic sentence segmentation}.
\newblock In \emph{Proceedings of the Second International Workshop on Spoken Language Translation}, Pittsburgh, Pennsylvania, USA.

\bibitem[{Matusov et~al.(2019)Matusov, Wilken, and Georgakopoulou}]{matusov-etal-2019-customizing}
Evgeny Matusov, Patrick Wilken, and Yota Georgakopoulou. 2019.
\newblock \href {https://doi.org/10.18653/v1/W19-5209} {Customizing neural machine translation for subtitling}.
\newblock In \emph{Proceedings of the Fourth Conference on Machine Translation (Volume 1: Research Papers)}, pages 82--93, Florence, Italy.

\bibitem[{Moritz et~al.(2020)Moritz, Hori, and Le}]{9054476}
Niko Moritz, Takaaki Hori, and Jonathan Le. 2020.
\newblock \href {https://doi.org/10.1109/ICASSP40776.2020.9054476} {Streaming automatic speech recognition with the transformer model}.
\newblock In \emph{ICASSP 2020 - 2020 IEEE International Conference on Acoustics, Speech and Signal Processing (ICASSP)}, pages 6074--6078.

\bibitem[{M{\"u}ller et~al.(2016)M{\"u}ller, Nguyen, Niehues, Cho, Kr{\"u}ger, Ha, Kilgour, Sperber, Mediani, St{\"u}ker, and Waibel}]{muller-etal-2016-lecture}
Markus M{\"u}ller, Thai~Son Nguyen, Jan Niehues, Eunah Cho, Bastian Kr{\"u}ger, Thanh-Le Ha, Kevin Kilgour, Matthias Sperber, Mohammed Mediani, Sebastian St{\"u}ker, and Alex Waibel. 2016.
\newblock \href {https://doi.org/10.18653/v1/N16-3017} {Lecture translator - speech translation framework for simultaneous lecture translation}.
\newblock In \emph{Proceedings of the 2016 Conference of the North {A}merican Chapter of the Association for Computational Linguistics: Demonstrations}, pages 82--86, San Diego, California. Association for Computational Linguistics.

\bibitem[{Narayanan et~al.(2019)Narayanan, Prabhavalkar, Chiu, Rybach, Sainath, and Strohman}]{narayanan2019recognizing}
Arun Narayanan, Rohit Prabhavalkar, Chung-Cheng Chiu, David Rybach, Tara~N. Sainath, and Trevor Strohman. 2019.
\newblock \href {https://doi.org/10.1109/ASRU46091.2019.9003913} {Recognizing long-form speech using streaming end-to-end models}.
\newblock In \emph{2019 IEEE Automatic Speech Recognition and Understanding Workshop (ASRU)}, pages 920--927.

\bibitem[{Nguyen et~al.(2021{\natexlab{a}})Nguyen, Estève, and Besacier}]{9414276}
Ha~Nguyen, Yannick Estève, and Laurent Besacier. 2021{\natexlab{a}}.
\newblock \href {https://doi.org/10.1109/ICASSP39728.2021.9414276} {An empirical study of end-to-end simultaneous speech translation decoding strategies}.
\newblock In \emph{ICASSP 2021 - 2021 IEEE International Conference on Acoustics, Speech and Signal Processing (ICASSP)}, pages 7528--7532.

\bibitem[{Nguyen et~al.(2021{\natexlab{b}})Nguyen, Estève, and Besacier}]{nguyen2021impact}
Ha~Nguyen, Yannick Estève, and Laurent Besacier. 2021{\natexlab{b}}.
\newblock \href {https://doi.org/10.21437/Interspeech.2021-608} {Impact of encoding and segmentation strategies on end-to-end simultaneous speech translation}.
\newblock In \emph{Interspeech 2021}, pages 2371--2375.

\bibitem[{Niehues et~al.(2018{\natexlab{a}})Niehues, Cattoni, St{\"u}ker, Cettolo, Turchi, and Federico}]{niehues-etal-2018-iwslt}
Jan Niehues, Rolando Cattoni, Sebastian St{\"u}ker, Mauro Cettolo, Marco Turchi, and Marcello Federico. 2018{\natexlab{a}}.
\newblock \href {https://aclanthology.org/2018.iwslt-1.1} {The {IWSLT} 2018 evaluation campaign}.
\newblock In \emph{Proceedings of the 15th International Conference on Spoken Language Translation}, pages 2--6, Brussels. International Conference on Spoken Language Translation.

\bibitem[{Niehues et~al.(2019)Niehues, Cattoni, St{\"u}ker, Negri, Turchi, Ha, Salesky, Sanabria, Barrault, Specia, and Federico}]{niehues-etal-2019-iwslt}
Jan Niehues, Rolando Cattoni, Sebastian St{\"u}ker, Matteo Negri, Marco Turchi, Thanh-Le Ha, Elizabeth Salesky, Ramon Sanabria, Loic Barrault, Lucia Specia, and Marcello Federico. 2019.
\newblock \href {https://aclanthology.org/2019.iwslt-1.1} {The {IWSLT} 2019 evaluation campaign}.
\newblock In \emph{Proceedings of the 16th International Conference on Spoken Language Translation}, Hong Kong. Association for Computational Linguistics.

\bibitem[{Niehues et~al.(2016)Niehues, Nguyen, Cho, Ha, Kilgour, Müller, Sperber, Stüker, and Waibel}]{niehues16_interspeech}
Jan Niehues, Thai~Son Nguyen, Eunah Cho, Thanh-Le Ha, Kevin Kilgour, Markus Müller, Matthias Sperber, Sebastian Stüker, and Alex Waibel. 2016.
\newblock \href {https://doi.org/10.21437/Interspeech.2016-154} {{Dynamic Transcription for Low-Latency Speech Translation}}.
\newblock In \emph{Proc. Interspeech 2016}, pages 2513--2517.

\bibitem[{Niehues et~al.(2018{\natexlab{b}})Niehues, Pham, Ha, Sperber, and Waibel}]{NiehuesPhamHa2018_1000087584}
Jan Niehues, Ngoc-Quan Pham, Thanh-Le Ha, Matthias Sperber, and Alex Waibel. 2018{\natexlab{b}}.
\newblock \href {https://doi.org/10.21437/Interspeech.2018-1055} {Low-latency neural speech translation}.
\newblock In \emph{Interspeech 2018}, pages 1293--1297.

\bibitem[{Novitasari et~al.(2022)Novitasari, Fukuda, and Kurata}]{novitasari2022improving}
Sashi Novitasari, Takashi Fukuda, and Gakuto Kurata. 2022.
\newblock \href {https://doi.org/10.21437/Interspeech.2022-260} {Improving asr robustness in noisy condition through vad integration}.
\newblock In \emph{Interspeech 2022}, pages 3784--3788.

\bibitem[{Novitasari et~al.(2021)Novitasari, Sakti, and Nakamura}]{NOVITASARI20212021EDP7014}
Sashi Novitasari, Sakriani Sakti, and Satoshi Nakamura. 2021.
\newblock \href {https://doi.org/10.1587/transinf.2021EDP7014} {Neural incremental speech recognition toward real-time machine speech translation}.
\newblock \emph{IEICE Transactions on Information and Systems}, E104.D(12):2195--2208.

\bibitem[{Oda et~al.(2014)Oda, Neubig, Sakti, Toda, and Nakamura}]{oda-etal-2014-optimizing}
Yusuke Oda, Graham Neubig, Sakriani Sakti, Tomoki Toda, and Satoshi Nakamura. 2014.
\newblock \href {https://doi.org/10.3115/v1/P14-2090} {Optimizing segmentation strategies for simultaneous speech translation}.
\newblock In \emph{Proceedings of the 52nd Annual Meeting of the Association for Computational Linguistics (Volume 2: Short Papers)}, pages 551--556, Baltimore, Maryland. Association for Computational Linguistics.

\bibitem[{Omachi et~al.(2023)Omachi, Yan, Dalmia, Fujita, and Watanabe}]{10095896}
Motoi Omachi, Brian Yan, Siddharth Dalmia, Yuya Fujita, and Shinji Watanabe. 2023.
\newblock \href {https://doi.org/10.1109/ICASSP49357.2023.10095896} {Align, write, re-order: Explainable end-to-end speech translation via operation sequence generation}.
\newblock In \emph{ICASSP 2023 - 2023 IEEE International Conference on Acoustics, Speech and Signal Processing (ICASSP)}, pages 1--5.

\bibitem[{Papi et~al.(2023{\natexlab{a}})Papi, Gaido, and Negri}]{papi-etal-2023-direct}
Sara Papi, Marco Gaido, and Matteo Negri. 2023{\natexlab{a}}.
\newblock \href {https://doi.org/10.18653/v1/2023.iwslt-1.11} {Direct models for simultaneous translation and automatic subtitling: {FBK}@{IWSLT}2023}.
\newblock In \emph{Proceedings of the 20th International Conference on Spoken Language Translation (IWSLT 2023)}, pages 159--168, Toronto, Canada (in-person and online). Association for Computational Linguistics.

\bibitem[{Papi et~al.(2024{\natexlab{a}})Papi, Gaido, Negri, and Bentivogli}]{papi2024simulseamless}
Sara Papi, Marco Gaido, Matteo Negri, and Luisa Bentivogli. 2024{\natexlab{a}}.
\newblock \href {https://doi.org/10.18653/v1/2024.iwslt-1.11} {{S}imul{S}eamless: {FBK} at {IWSLT} 2024 simultaneous speech translation}.
\newblock In \emph{Proceedings of the 21st International Conference on Spoken Language Translation (IWSLT 2024)}, pages 72--79, Bangkok, Thailand (in-person and online). Association for Computational Linguistics.

\bibitem[{Papi et~al.(2024{\natexlab{b}})Papi, Gaido, Negri, and Bentivogli}]{papi2024streamatt}
Sara Papi, Marco Gaido, Matteo Negri, and Luisa Bentivogli. 2024{\natexlab{b}}.
\newblock \href {https://doi.org/10.18653/v1/2024.acl-long.202} {{S}tream{A}tt: Direct streaming speech-to-text translation with attention-based audio history selection}.
\newblock In \emph{Proceedings of the 62nd Annual Meeting of the Association for Computational Linguistics (Volume 1: Long Papers)}, pages 3692--3707, Bangkok, Thailand. Association for Computational Linguistics.

\bibitem[{Papi et~al.(2022{\natexlab{a}})Papi, Gaido, Negri, and Turchi}]{papi-etal-2022-simultaneous}
Sara Papi, Marco Gaido, Matteo Negri, and Marco Turchi. 2022{\natexlab{a}}.
\newblock \href {https://doi.org/10.18653/v1/2022.findings-emnlp.11} {Does simultaneous speech translation need simultaneous models?}
\newblock In \emph{Findings of the Association for Computational Linguistics: EMNLP 2022}, pages 141--153, Abu Dhabi, United Arab Emirates. Association for Computational Linguistics.

\bibitem[{Papi et~al.(2022{\natexlab{b}})Papi, Gaido, Negri, and Turchi}]{papi-etal-2022-generation}
Sara Papi, Marco Gaido, Matteo Negri, and Marco Turchi. 2022{\natexlab{b}}.
\newblock \href {https://doi.org/10.18653/v1/2022.autosimtrans-1.2} {Over-generation cannot be rewarded: Length-adaptive average lagging for simultaneous speech translation}.
\newblock In \emph{Proceedings of the Third Workshop on Automatic Simultaneous Translation}, pages 12--17, Online. Association for Computational Linguistics.

\bibitem[{Papi et~al.(2022{\natexlab{c}})Papi, Karakanta, Negri, and Turchi}]{papi-etal-2022-dodging}
Sara Papi, Alina Karakanta, Matteo Negri, and Marco Turchi. 2022{\natexlab{c}}.
\newblock \href {https://aclanthology.org/2022.aacl-short.59} {Dodging the data bottleneck: Automatic subtitling with automatically segmented {ST} corpora}.
\newblock In \emph{Proceedings of the 2nd Conference of the Asia-Pacific Chapter of the Association for Computational Linguistics and the 12th International Joint Conference on Natural Language Processing (Volume 2: Short Papers)}, pages 480--487, Online only.

\bibitem[{Papi et~al.(2021)Papi, Negri, and Turchi}]{papi2021visualization}
Sara Papi, Matteo Negri, and Marco Turchi. 2021.
\newblock Visualization: The missing factor in simultaneous speech translation.
\newblock In \emph{CEUR WORKSHOP PROCEEDINGS}, volume 3033.

\bibitem[{Papi et~al.(2023{\natexlab{b}})Papi, Negri, and Turchi}]{papi-etal-2023-attention}
Sara Papi, Matteo Negri, and Marco Turchi. 2023{\natexlab{b}}.
\newblock \href {https://doi.org/10.18653/v1/2023.acl-long.745} {Attention as a guide for simultaneous speech translation}.
\newblock In \emph{Proceedings of the 61st Annual Meeting of the Association for Computational Linguistics (Volume 1: Long Papers)}, pages 13340--13356, Toronto, Canada. Association for Computational Linguistics.

\bibitem[{Papi et~al.(2023{\natexlab{c}})Papi, Turchi, and Negri}]{papi23_interspeech}
Sara Papi, Marco Turchi, and Matteo Negri. 2023{\natexlab{c}}.
\newblock \href {https://doi.org/10.21437/Interspeech.2023-170} {{AlignAtt: Using Attention-based Audio-Translation Alignments as a Guide for Simultaneous Speech Translation}}.
\newblock In \emph{Proc. INTERSPEECH 2023}, pages 3974--3978.

\bibitem[{Papi et~al.(2024{\natexlab{c}})Papi, Wang, Chen, Xue, Kanda, Li, and Gaur}]{papi2024leveraging}
Sara Papi, Peidong Wang, Junkun Chen, Jian Xue, Naoyuki Kanda, Jinyu Li, and Yashesh Gaur. 2024{\natexlab{c}}.
\newblock \href {https://doi.org/10.1109/ICASSP48485.2024.10447565} {Leveraging timestamp information for serialized joint streaming recognition and translation}.
\newblock In \emph{ICASSP 2024 - 2024 IEEE International Conference on Acoustics, Speech and Signal Processing (ICASSP)}, pages 10381--10385.

\bibitem[{Papi et~al.(2023{\natexlab{d}})Papi, Wang, Chen, Xue, Li, and Gaur}]{papi2023token}
Sara Papi, Peidong Wang, Junkun Chen, Jian Xue, Jinyu Li, and Yashesh Gaur. 2023{\natexlab{d}}.
\newblock \href {https://doi.org/10.1109/ASRU57964.2023.10389715} {Token-level serialized output training for joint streaming asr and st leveraging textual alignments}.
\newblock In \emph{2023 IEEE Automatic Speech Recognition and Understanding Workshop (ASRU)}, pages 1--8.

\bibitem[{Papineni et~al.(2002)Papineni, Roukos, Ward, and Zhu}]{papineni-etal-2002-bleu}
Kishore Papineni, Salim Roukos, Todd Ward, and Wei-Jing Zhu. 2002.
\newblock \href {https://doi.org/10.3115/1073083.1073135} {{B}leu: a method for automatic evaluation of machine translation}.
\newblock In \emph{Proceedings of the 40th Annual Meeting of the Association for Computational Linguistics}, pages 311--318, Philadelphia, Pennsylvania, USA. Association for Computational Linguistics.

\bibitem[{Park et~al.(2022)Park, Kanda, Dimitriadis, Han, Watanabe, and Narayanan}]{PARK2022101317}
Tae~Jin Park, Naoyuki Kanda, Dimitrios Dimitriadis, Kyu~J. Han, Shinji Watanabe, and Shrikanth Narayanan. 2022.
\newblock \href {https://doi.org/https://doi.org/10.1016/j.csl.2021.101317} {A review of speaker diarization: Recent advances with deep learning}.
\newblock \emph{Computer Speech \& Language}, 72:101317.

\bibitem[{Paulik and Waibel(2010)}]{paulik10_interspeech}
Matthias Paulik and Alex Waibel. 2010.
\newblock \href {https://doi.org/10.21437/Interspeech.2010-680} {{Rapid development of speech translation using consecutive interpretation}}.
\newblock In \emph{Proc. Interspeech 2010}, pages 2534--2537.

\bibitem[{Perego et~al.(2010)Perego, Missier, Porta, and Mosconi}]{doi:10.1080/15213269.2010.502873}
Elisa Perego, Fabio~Del Missier, Marco Porta, and Mauro Mosconi. 2010.
\newblock \href {https://doi.org/10.1080/15213269.2010.502873} {The cognitive effectiveness of subtitle processing}.
\newblock \emph{Media Psychology}, 13(3):243--272.

\bibitem[{Pol{\'a}k(2023)}]{polak-2023-long}
Peter Pol{\'a}k. 2023.
\newblock \href {https://doi.org/10.18653/v1/2023.ijcnlp-srw.9} {Long-form simultaneous speech translation: Thesis proposal}.
\newblock In \emph{Proceedings of the 13th International Joint Conference on Natural Language Processing and the 3rd Conference of the Asia-Pacific Chapter of the Association for Computational Linguistics: Student Research Workshop}, pages 64--74, Nusa Dua, Bali. Association for Computational Linguistics.

\bibitem[{Pol{\'a}k and Bojar(2023)}]{polak2023long}
Peter Pol{\'a}k and Ond{\v{r}}ej Bojar. 2023.
\newblock Long-form end-to-end speech translation via latent alignment segmentation.
\newblock \emph{arXiv preprint arXiv:2309.11384}.

\bibitem[{Pol{\'a}k et~al.(2023)Pol{\'a}k, Liu, Pham, Niehues, Waibel, and Bojar}]{polak-etal-2023-towards}
Peter Pol{\'a}k, Danni Liu, Ngoc-Quan Pham, Jan Niehues, Alexander Waibel, and Ond{\v{r}}ej Bojar. 2023.
\newblock \href {https://doi.org/10.18653/v1/2023.iwslt-1.37} {Towards efficient simultaneous speech translation: {CUNI}-{KIT} system for simultaneous track at {IWSLT} 2023}.
\newblock In \emph{Proceedings of the 20th International Conference on Spoken Language Translation (IWSLT 2023)}, pages 389--396, Toronto, Canada (in-person and online). Association for Computational Linguistics.

\bibitem[{Pol{\'a}k et~al.(2022)Pol{\'a}k, Pham, Nguyen, Liu, Mullov, Niehues, Bojar, and Waibel}]{polak-etal-2022-cuni}
Peter Pol{\'a}k, Ngoc-Quan Pham, Tuan~Nam Nguyen, Danni Liu, Carlos Mullov, Jan Niehues, Ond{\v{r}}ej Bojar, and Alexander Waibel. 2022.
\newblock \href {https://doi.org/10.18653/v1/2022.iwslt-1.24} {{CUNI}-{KIT} system for simultaneous speech translation task at {IWSLT} 2022}.
\newblock In \emph{Proceedings of the 19th International Conference on Spoken Language Translation (IWSLT 2022)}, pages 277--285, Dublin, Ireland (in-person and online). Association for Computational Linguistics.

\bibitem[{Polák et~al.(2023)Polák, Yan, Watanabe, Waibel, and Bojar}]{polak23_interspeech}
Peter Polák, Brian Yan, Shinji Watanabe, Alex Waibel, and Ondřej Bojar. 2023.
\newblock \href {https://doi.org/10.21437/Interspeech.2023-2225} {{Incremental Blockwise Beam Search for Simultaneous Speech Translation with Controllable Quality-Latency Tradeoff}}.
\newblock In \emph{Proc. INTERSPEECH 2023}, pages 3979--3983.

\bibitem[{Potapczyk and Przybysz(2020)}]{potapczyk-przybysz-2020-srpols}
Tomasz Potapczyk and Pawel Przybysz. 2020.
\newblock \href {https://doi.org/10.18653/v1/2020.iwslt-1.9} {{SRPOL}{'}s system for the {IWSLT} 2020 end-to-end speech translation task}.
\newblock In \emph{Proceedings of the 17th International Conference on Spoken Language Translation}, pages 89--94, Online. Association for Computational Linguistics.

\bibitem[{Radford et~al.(2023)Radford, Kim, Xu, Brockman, McLeavey, and Sutskever}]{radford2023robust}
Alec Radford, Jong~Wook Kim, Tao Xu, Greg Brockman, Christine McLeavey, and Ilya Sutskever. 2023.
\newblock Robust speech recognition via large-scale weak supervision.
\newblock In \emph{International Conference on Machine Learning}, pages 28492--28518. PMLR.

\bibitem[{Raffel and Chen(2023)}]{raffel-chen-2023-implicit}
Matthew Raffel and Lizhong Chen. 2023.
\newblock \href {https://doi.org/10.18653/v1/2023.findings-acl.816} {Implicit memory transformer for computationally efficient simultaneous speech translation}.
\newblock In \emph{Findings of the Association for Computational Linguistics: ACL 2023}, pages 12900--12907, Toronto, Canada. Association for Computational Linguistics.

\bibitem[{Raffel et~al.(2023)Raffel, Penney, and Chen}]{10.5555/3618408.3619591}
Matthew Raffel, Drew Penney, and Lizhong Chen. 2023.
\newblock Shiftable context: addressing training-inference context mismatch in simultaneous speech translation.
\newblock In \emph{Proceedings of the 40th International Conference on Machine Learning}, ICML'23. JMLR.org.

\bibitem[{Rajendran et~al.(2013)Rajendran, Duchowski, Orero, Martínez, and Romero-Fresco}]{doi:10.1080/0907676X.2012.722651}
Dhevi~J. Rajendran, Andrew~T. Duchowski, Pilar Orero, Juan Martínez, and Pablo Romero-Fresco. 2013.
\newblock \href {https://doi.org/10.1080/0907676X.2012.722651} {Effects of text chunking on subtitling: A quantitative and qualitative examination}.
\newblock \emph{Perspectives}, 21(1):5--21.

\bibitem[{Rangarajan~Sridhar et~al.(2013)Rangarajan~Sridhar, Chen, Bangalore, Ljolje, and Chengalvarayan}]{rangarajan-sridhar-etal-2013-segmentation}
Vivek~Kumar Rangarajan~Sridhar, John Chen, Srinivas Bangalore, Andrej Ljolje, and Rathinavelu Chengalvarayan. 2013.
\newblock \href {https://aclanthology.org/N13-1023} {Segmentation strategies for streaming speech translation}.
\newblock In \emph{Proceedings of the 2013 Conference of the North {A}merican Chapter of the Association for Computational Linguistics: Human Language Technologies}, pages 230--238, Atlanta, Georgia. Association for Computational Linguistics.

\bibitem[{Rei et~al.(2022)Rei, C.~de Souza, Alves, Zerva, Farinha, Glushkova, Lavie, Coheur, and Martins}]{rei-etal-2022-comet}
Ricardo Rei, Jos{\'e}~G. C.~de Souza, Duarte Alves, Chrysoula Zerva, Ana~C Farinha, Taisiya Glushkova, Alon Lavie, Luisa Coheur, and Andr{\'e} F.~T. Martins. 2022.
\newblock \href {https://aclanthology.org/2022.wmt-1.52} {{COMET}-22: Unbabel-{IST} 2022 submission for the metrics shared task}.
\newblock In \emph{Proceedings of the Seventh Conference on Machine Translation (WMT)}, pages 578--585, Abu Dhabi, United Arab Emirates (Hybrid). Association for Computational Linguistics.

\bibitem[{Rei et~al.(2020)Rei, Stewart, Farinha, and Lavie}]{rei-etal-2020-comet}
Ricardo Rei, Craig Stewart, Ana~C Farinha, and Alon Lavie. 2020.
\newblock \href {https://doi.org/10.18653/v1/2020.emnlp-main.213} {{COMET}: A neural framework for {MT} evaluation}.
\newblock In \emph{Proceedings of the 2020 Conference on Empirical Methods in Natural Language Processing (EMNLP)}, pages 2685--2702, Online. Association for Computational Linguistics.

\bibitem[{Ren et~al.(2020)Ren, Liu, Tan, Zhang, Qin, Zhao, and Liu}]{ren-etal-2020-simulspeech}
Yi~Ren, Jinglin Liu, Xu~Tan, Chen Zhang, Tao Qin, Zhou Zhao, and Tie-Yan Liu. 2020.
\newblock \href {https://doi.org/10.18653/v1/2020.acl-main.350} {{S}imul{S}peech: End-to-end simultaneous speech to text translation}.
\newblock In \emph{Proceedings of the 58th Annual Meeting of the Association for Computational Linguistics}, pages 3787--3796, Online. Association for Computational Linguistics.

\bibitem[{Romero-Fresco(2010)}]{eyefixation}
Pablo Romero-Fresco. 2010.
\newblock Standing on quicksand: Hearing viewers' comprehension and reading patterns of respoken subtitles for the news.
\newblock In \emph{New insights into audiovisual translation and media accessibility}, pages 175--194. Brill.

\bibitem[{Romero-Fresco(2011)}]{respeaking}
Pablo Romero-Fresco. 2011.
\newblock \href {https://doi.org/10.4324/9781003073147} {\emph{Subtitling through Speech Recognition: Respeaking}}.
\newblock Routledge.

\bibitem[{Ryu et~al.(2006)Ryu, Matsubara, and Inagaki}]{ryu-etal-2006-simultaneous}
Koichiro Ryu, Shigeki Matsubara, and Yasuyoshi Inagaki. 2006.
\newblock \href {https://aclanthology.org/P06-2088} {Simultaneous {E}nglish-{J}apanese spoken language translation based on incremental dependency parsing and transfer}.
\newblock In \emph{Proceedings of the {COLING}/{ACL} 2006 Main Conference Poster Sessions}, pages 683--690, Sydney, Australia. Association for Computational Linguistics.

\bibitem[{Sakamoto et~al.(2013)Sakamoto, Abe, Sumita, and Kamatani}]{sakamoto-etal-2013-evaluation}
Akiko Sakamoto, Kazuhiko Abe, Kazuo Sumita, and Satoshi Kamatani. 2013.
\newblock \href {https://aclanthology.org/2013.iwslt-papers.18} {Evaluation of a simultaneous interpretation system and analysis of speech log for user experience assessment}.
\newblock In \emph{Proceedings of the 10th International Workshop on Spoken Language Translation: Papers}, Heidelberg, Germany.

\bibitem[{Schneider and Waibel(2020)}]{schneider-waibel-2020-towards}
Felix Schneider and Alexander Waibel. 2020.
\newblock \href {https://doi.org/10.18653/v1/2020.iwslt-1.28} {Towards stream translation: Adaptive computation time for simultaneous machine translation}.
\newblock In \emph{Proceedings of the 17th International Conference on Spoken Language Translation}, pages 228--236, Online. Association for Computational Linguistics.

\bibitem[{Sen et~al.(2022)Sen, Bojar, and Haddow}]{sen2022simultaneous}
Sukanta Sen, Ond{\v{r}}ej Bojar, and Barry Haddow. 2022.
\newblock Simultaneous translation for unsegmented input: A sliding window approach.
\newblock \emph{arXiv preprint arXiv:2210.09754}.

\bibitem[{Shavarani et~al.(2015)Shavarani, Siahbani, Seraj, and Sarkar}]{shavarani-etal-2015-learning}
Hassan Shavarani, Maryam Siahbani, Ramtin~Mehdizadeh Seraj, and Anoop Sarkar. 2015.
\newblock \href {https://aclanthology.org/2015.iwslt-papers.14} {Learning segmentations that balance latency versus quality in spoken language translation}.
\newblock In \emph{Proceedings of the 12th International Workshop on Spoken Language Translation: Papers}, pages 217--224, Da Nang, Vietnam.

\bibitem[{Shimizu et~al.(2013)Shimizu, Neubig, Sakti, Toda, and Nakamura}]{shimizu-etal-2013-constructing}
Hiroaki Shimizu, Graham Neubig, Sakriani Sakti, Tomoki Toda, and Satoshi Nakamura. 2013.
\newblock \href {https://aclanthology.org/2013.iwslt-papers.3} {Constructing a speech translation system using simultaneous interpretation data}.
\newblock In \emph{Proceedings of the 10th International Workshop on Spoken Language Translation: Papers}, Heidelberg, Germany.

\bibitem[{Siahbani et~al.(2018)Siahbani, Shavarani, Alinejad, and Sarkar}]{siahbani-etal-2018-simultaneous}
Maryam Siahbani, Hassan Shavarani, Ashkan Alinejad, and Anoop Sarkar. 2018.
\newblock \href {https://aclanthology.org/W18-1815} {Simultaneous translation using optimized segmentation}.
\newblock In \emph{Proceedings of the 13th Conference of the Association for Machine Translation in the {A}mericas (Volume 1: Research Track)}, pages 154--167, Boston, MA. Association for Machine Translation in the Americas.

\bibitem[{Sinclair et~al.(2014)Sinclair, Bell, Birch, and McInnes}]{sinclair14_interspeech}
Mark Sinclair, Peter Bell, Alexandra Birch, and Fergus McInnes. 2014.
\newblock \href {https://doi.org/10.21437/Interspeech.2014-511} {{A semi-Markov model for speech segmentation with an utterance-break prior}}.
\newblock In \emph{Proc. Interspeech 2014}, pages 2351--2355.

\bibitem[{Sohn et~al.(1999)Sohn, Kim, and Sung}]{sohn1999statistical}
Jongseo Sohn, Nam~Soo Kim, and Wonyong Sung. 1999.
\newblock \href {https://doi.org/10.1109/97.736233} {A statistical model-based voice activity detection}.
\newblock \emph{IEEE Signal Processing Letters}, 6(1):1--3.

\bibitem[{Sperber and Paulik(2020)}]{sperber-paulik-2020-speech}
Matthias Sperber and Matthias Paulik. 2020.
\newblock \href {https://doi.org/10.18653/v1/2020.acl-main.661} {Speech translation and the end-to-end promise: Taking stock of where we are}.
\newblock In \emph{Proceedings of the 58th Annual Meeting of the Association for Computational Linguistics}, pages 7409--7421, Online. Association for Computational Linguistics.

\bibitem[{Stentiford and Steer(1988)}]{cascade}
Fred~WM Stentiford and Martin~G Steer. 1988.
\newblock Machine translation of speech.
\newblock \emph{British Telecom technology journal}, 6(2):116--122.

\bibitem[{Subramanya and Niehues(2022)}]{subramanya2022multilingual}
Shashank Subramanya and Jan Niehues. 2022.
\newblock Multilingual simultaneous speech translation.
\newblock \emph{arXiv preprint arXiv:2203.14835}.

\bibitem[{Tan et~al.(2024)Tan, Chen, Chen, Qin, Xu, Zhang, Van~Durme, and Koehn}]{tan2024streaming}
Weiting Tan, Yunmo Chen, Tongfei Chen, Guanghui Qin, Haoran Xu, Heidi~C Zhang, Benjamin Van~Durme, and Philipp Koehn. 2024.
\newblock Streaming sequence transduction through dynamic compression.
\newblock \emph{arXiv preprint arXiv:2402.01172}.

\bibitem[{Tang et~al.(2023)Tang, Sun, Inaguma, Chen, Dong, Ma, Tomasello, and Pino}]{tang-etal-2023-hybrid}
Yun Tang, Anna Sun, Hirofumi Inaguma, Xinyue Chen, Ning Dong, Xutai Ma, Paden Tomasello, and Juan Pino. 2023.
\newblock \href {https://doi.org/10.18653/v1/2023.acl-long.695} {Hybrid transducer and attention based encoder-decoder modeling for speech-to-text tasks}.
\newblock In \emph{Proceedings of the 61st Annual Meeting of the Association for Computational Linguistics (Volume 1: Long Papers)}, pages 12441--12455, Toronto, Canada. Association for Computational Linguistics.

\bibitem[{Tay et~al.(2022)Tay, Dehghani, Bahri, and Metzler}]{10.1145/3530811}
Yi~Tay, Mostafa Dehghani, Dara Bahri, and Donald Metzler. 2022.
\newblock \href {https://doi.org/10.1145/3530811} {Efficient transformers: A survey}.
\newblock \emph{ACM Comput. Surv.}, 55(6).

\bibitem[{Tiedemann and Scherrer(2017)}]{tiedemann-scherrer-2017-neural}
J{\"o}rg Tiedemann and Yves Scherrer. 2017.
\newblock \href {https://doi.org/10.18653/v1/W17-4811} {Neural machine translation with extended context}.
\newblock In \emph{Proceedings of the Third Workshop on Discourse in Machine Translation}, pages 82--92, Copenhagen, Denmark. Association for Computational Linguistics.

\bibitem[{Tsiamas et~al.(2022)Tsiamas, Gállego, Fonollosa, and Costa-jussà}]{tsiamas22_interspeech}
Ioannis Tsiamas, Gerard~I. Gállego, José A.~R. Fonollosa, and Marta~R. Costa-jussà. 2022.
\newblock \href {https://doi.org/10.21437/Interspeech.2022-59} {{SHAS: Approaching optimal Segmentation for End-to-End Speech Translation}}.
\newblock In \emph{Proc. Interspeech 2022}, pages 106--110.

\bibitem[{Waibel(2004)}]{Waibel2004SpeechTP}
Alexander~H. Waibel. 2004.
\newblock \href {https://api.semanticscholar.org/CorpusID:18867313} {Speech translation: past, present and future}.
\newblock In \emph{Interspeech}.

\bibitem[{Waibel et~al.(1991)Waibel, Jain, McNair, Saito, Hauptmann, and Tebelskis}]{Waibel1991JANUSAS}
Alexander~H. Waibel, Ajay~N. Jain, Arthur~E. McNair, Hiroaki Saito, Alexander Hauptmann, and Joe Tebelskis. 1991.
\newblock \href {https://api.semanticscholar.org/CorpusID:17834225} {Janus: a speech-to-speech translation system using connectionist and symbolic processing strategies}.
\newblock \emph{[Proceedings] ICASSP 91: 1991 International Conference on Acoustics, Speech, and Signal Processing}, pages 793--796 vol.2.

\bibitem[{Wang et~al.(2022{\natexlab{a}})Wang, Tong, Guo, He, and Maas}]{9746873}
Jinhan Wang, Xiaosu Tong, Jinxi Guo, Di~He, and Roland Maas. 2022{\natexlab{a}}.
\newblock \href {https://doi.org/10.1109/ICASSP43922.2022.9746873} {Vadoi: Voice-activity-detection overlapping inference for end-to-end long-form speech recognition}.
\newblock In \emph{ICASSP 2022 - 2022 IEEE International Conference on Acoustics, Speech and Signal Processing (ICASSP)}, pages 6977--6981.

\bibitem[{Wang et~al.(2022{\natexlab{b}})Wang, Guo, Li, Qiao, Wang, Li, Su, Chen, Zhang, Tao, Yang, and Qin}]{wang-etal-2022-hw-tscs}
Minghan Wang, Jiaxin Guo, Yinglu Li, Xiaosong Qiao, Yuxia Wang, Zongyao Li, Chang Su, Yimeng Chen, Min Zhang, Shimin Tao, Hao Yang, and Ying Qin. 2022{\natexlab{b}}.
\newblock \href {https://doi.org/10.18653/v1/2022.iwslt-1.21} {The {HW}-{TSC}{'}s simultaneous speech translation system for {IWSLT} 2022 evaluation}.
\newblock In \emph{Proceedings of the 19th International Conference on Spoken Language Translation (IWSLT 2022)}, pages 247--254, Dublin, Ireland (in-person and online). Association for Computational Linguistics.

\bibitem[{Wang et~al.(2023)Wang, Sun, Xue, Wu, Zhou, Gaur, Liu, and Li}]{wang23oa_interspeech}
Peidong Wang, Eric Sun, Jian Xue, Yu~Wu, Long Zhou, Yashesh Gaur, Shujie Liu, and Jinyu Li. 2023.
\newblock \href {https://doi.org/10.21437/Interspeech.2023-2004} {{LAMASSU: A Streaming Language-Agnostic Multilingual Speech Recognition and Translation Model Using Neural Transducers}}.
\newblock In \emph{Proc. INTERSPEECH 2023}, pages 57--61.

\bibitem[{Wang et~al.(2016)Wang, Finch, Utiyama, and Sumita}]{wang-etal-2016-efficient}
Xiaolin Wang, Andrew Finch, Masao Utiyama, and Eiichiro Sumita. 2016.
\newblock \href {https://aclanthology.org/W16-4613} {An efficient and effective online sentence segmenter for simultaneous interpretation}.
\newblock In \emph{Proceedings of the 3rd Workshop on {A}sian Translation ({WAT}2016)}, pages 139--148, Osaka, Japan. The COLING 2016 Organizing Committee.

\bibitem[{Wang et~al.(2019)Wang, Utiyama, and Sumita}]{wang-etal-2019-online}
Xiaolin Wang, Masao Utiyama, and Eiichiro Sumita. 2019.
\newblock \href {https://aclanthology.org/W19-6601} {Online sentence segmentation for simultaneous interpretation using multi-shifted recurrent neural network}.
\newblock In \emph{Proceedings of Machine Translation Summit XVII: Research Track}, pages 1--11, Dublin, Ireland. European Association for Machine Translation.

\bibitem[{Weiss et~al.(2017)Weiss, Chorowski, Jaitly, Wu, and Chen}]{Weiss2017SequencetoSequenceMC}
Ron~J. Weiss, Jan Chorowski, Navdeep Jaitly, Yonghui Wu, and Zhifeng Chen. 2017.
\newblock \href {https://doi.org/10.21437/Interspeech.2017-503} {Sequence-to-sequence models can directly translate foreign speech}.
\newblock In \emph{Interspeech 2017}, pages 2625--2629.

\bibitem[{Weller et~al.(2021)Weller, Sperber, Gollan, and Kluivers}]{weller-etal-2021-streaming}
Orion Weller, Matthias Sperber, Christian Gollan, and Joris Kluivers. 2021.
\newblock \href {https://doi.org/10.18653/v1/2021.eacl-main.216} {Streaming models for joint speech recognition and translation}.
\newblock In \emph{Proceedings of the 16th Conference of the European Chapter of the Association for Computational Linguistics: Main Volume}, pages 2533--2539, Online. Association for Computational Linguistics.

\bibitem[{Weller et~al.(2022)Weller, Sperber, Pires, Setiawan, Gollan, Telaar, and Paulik}]{weller-etal-2022-end}
Orion Weller, Matthias Sperber, Telmo Pires, Hendra Setiawan, Christian Gollan, Dominic Telaar, and Matthias Paulik. 2022.
\newblock \href {https://doi.org/10.18653/v1/2022.findings-acl.113} {End-to-end speech translation for code switched speech}.
\newblock In \emph{Findings of the Association for Computational Linguistics: ACL 2022}, pages 1435--1448, Dublin, Ireland. Association for Computational Linguistics.

\bibitem[{Wilken et~al.(2020)Wilken, Alkhouli, Matusov, and Golik}]{wilken-etal-2020-neural}
Patrick Wilken, Tamer Alkhouli, Evgeny Matusov, and Pavel Golik. 2020.
\newblock \href {https://doi.org/10.18653/v1/2020.iwslt-1.29} {Neural simultaneous speech translation using alignment-based chunking}.
\newblock In \emph{Proceedings of the 17th International Conference on Spoken Language Translation}, pages 237--246, Online. Association for Computational Linguistics.

\bibitem[{Wolfel et~al.(2008)Wolfel, Kolss, Kraft, Niehues, Paulik, and Waibel}]{4777883}
Matthias Wolfel, Muntsin Kolss, Florian Kraft, Jan Niehues, Matthias Paulik, and Alex Waibel. 2008.
\newblock \href {https://doi.org/10.1109/SLT.2008.4777883} {Simultaneous machine translation of german lectures into english: Investigating research challenges for the future}.
\newblock In \emph{2008 IEEE Spoken Language Technology Workshop}, pages 233--236.

\bibitem[{Wo{\l}k and Marasek(2014)}]{wolk2014real}
Krzysztof Wo{\l}k and Krzysztof Marasek. 2014.
\newblock Real-time statistical speech translation.
\newblock In \emph{New Perspectives in Information Systems and Technologies, Volume 1}, pages 107--113. Springer.

\bibitem[{Woszczyna et~al.(1998)Woszczyna, Broadhead, Gates, Gavalda, Lavie, Levin, and Waibel}]{woszczyna1998modular}
Monika Woszczyna, Matthew Broadhead, Donna Gates, Marsal Gavalda, Alon Lavie, Lori Levin, and Alex Waibel. 1998.
\newblock A modular approach to spoken language translation for large domains.
\newblock In \emph{Machine Translation and the Information Soup: Third Conference of the Association for Machine Translation in the Americas AMTA’98 Langhorne, PA, USA, October 28--31, 1998 Proceedings 3}, pages 31--40. Springer.

\bibitem[{Wu et~al.(2020)Wu, Wang, Shi, Yeh, and Zhang}]{wu20i_interspeech}
Chunyang Wu, Yongqiang Wang, Yangyang Shi, Ching-Feng Yeh, and Frank Zhang. 2020.
\newblock \href {https://doi.org/10.21437/Interspeech.2020-2079} {{Streaming Transformer-Based Acoustic Models Using Self-Attention with Augmented Memory}}.
\newblock In \emph{Proc. Interspeech 2020}, pages 2132--2136.

\bibitem[{Xiong et~al.(2019)Xiong, Zhang, Zhang, He, Wu, and Wang}]{xiong2019dutongchuan}
Hao Xiong, Ruiqing Zhang, Chuanqiang Zhang, Zhongjun He, Hua Wu, and Haifeng Wang. 2019.
\newblock Dutongchuan: Context-aware translation model for simultaneous interpreting.
\newblock \emph{arXiv preprint arXiv:1907.12984}.

\bibitem[{Xue et~al.(2022)Xue, Wang, Li, Post, and Gaur}]{xue22d_interspeech}
Jian Xue, Peidong Wang, Jinyu Li, Matt Post, and Yashesh Gaur. 2022.
\newblock \href {https://doi.org/10.21437/Interspeech.2022-10953} {{Large-Scale Streaming End-to-End Speech Translation with Neural Transducers}}.
\newblock In \emph{Proc. Interspeech 2022}, pages 3263--3267.

\bibitem[{Xue et~al.(2023)Xue, Wang, Li, and Sun}]{10389799}
Jian Xue, Peidong Wang, Jinyu Li, and Eric Sun. 2023.
\newblock \href {https://doi.org/10.1109/ASRU57964.2023.10389799} {A weakly-supervised streaming multilingual speech model with truly zero-shot capability}.
\newblock In \emph{2023 IEEE Automatic Speech Recognition and Understanding Workshop (ASRU)}, pages 1--7.

\bibitem[{Yan et~al.(2023)Yan, Shi, Maiti, Chen, Li, Peng, Arora, and Watanabe}]{yan-etal-2023-cmus}
Brian Yan, Jiatong Shi, Soumi Maiti, William Chen, Xinjian Li, Yifan Peng, Siddhant Arora, and Shinji Watanabe. 2023.
\newblock \href {https://doi.org/10.18653/v1/2023.iwslt-1.20} {{CMU}{'}s {IWSLT} 2023 simultaneous speech translation system}.
\newblock In \emph{Proceedings of the 20th International Conference on Spoken Language Translation (IWSLT 2023)}, pages 235--240, Toronto, Canada (in-person and online). Association for Computational Linguistics.

\bibitem[{Yang et~al.(2024)Yang, Kanda, Wang, Chen, Wang, Xue, Li, and Yoshioka}]{10446050}
Mu~Yang, Naoyuki Kanda, Xiaofei Wang, Junkun Chen, Peidong Wang, Jian Xue, Jinyu Li, and Takuya Yoshioka. 2024.
\newblock \href {https://doi.org/10.1109/ICASSP48485.2024.10446050} {Diarist: Streaming speech translation with speaker diarization}.
\newblock In \emph{ICASSP 2024 - 2024 IEEE International Conference on Acoustics, Speech and Signal Processing (ICASSP)}, pages 10866--10870.

\bibitem[{Yao and Haddow(2020)}]{yao-haddow-2020-dynamic}
Yuekun Yao and Barry Haddow. 2020.
\newblock \href {https://aclanthology.org/2020.amta-research.12} {Dynamic masking for improved stability in online spoken language translation}.
\newblock In \emph{Proceedings of the 14th Conference of the Association for Machine Translation in the Americas (Volume 1: Research Track)}, pages 123--136, Virtual. Association for Machine Translation in the Americas.

\bibitem[{Yarmohammadi et~al.(2013)Yarmohammadi, Rangarajan~Sridhar, Bangalore, and Sankaran}]{yarmohammadi-etal-2013-incremental}
Mahsa Yarmohammadi, Vivek~Kumar Rangarajan~Sridhar, Srinivas Bangalore, and Baskaran Sankaran. 2013.
\newblock \href {https://aclanthology.org/I13-1141} {Incremental segmentation and decoding strategies for simultaneous translation}.
\newblock In \emph{Proceedings of the Sixth International Joint Conference on Natural Language Processing}, pages 1032--1036, Nagoya, Japan. Asian Federation of Natural Language Processing.

\bibitem[{Yoshimura et~al.(2020)Yoshimura, Hayashi, Takeda, and Watanabe}]{yoshimura2020end}
Takenori Yoshimura, Tomoki Hayashi, Kazuya Takeda, and Shinji Watanabe. 2020.
\newblock \href {https://doi.org/10.1109/ICASSP40776.2020.9054358} {End-to-end automatic speech recognition integrated with ctc-based voice activity detection}.
\newblock In \emph{ICASSP 2020 - 2020 IEEE International Conference on Acoustics, Speech and Signal Processing (ICASSP)}, pages 6999--7003.

\bibitem[{Zaidi et~al.(2021)Zaidi, Lee, Kim, and Kim}]{zaidi2021decision}
Mohd~Abbas Zaidi, Beomseok Lee, Sangha Kim, and Chanwoo Kim. 2021.
\newblock Decision attentive regularization to improve simultaneous speech translation systems.
\newblock \emph{arXiv preprint arXiv:2110.15729}.

\bibitem[{Zaidi et~al.(2022)Zaidi, Lee, Kim, and Kim}]{zaidi22_interspeech}
Mohd~Abbas Zaidi, Beomseok Lee, Sangha Kim, and Chanwoo Kim. 2022.
\newblock \href {https://doi.org/10.21437/Interspeech.2022-10617} {{Cross-Modal Decision Regularization for Simultaneous Speech Translation}}.
\newblock In \emph{Proc. Interspeech 2022}, pages 116--120.

\bibitem[{Zeng et~al.(2021)Zeng, Li, and Liu}]{zeng-etal-2021-realtrans}
Xingshan Zeng, Liangyou Li, and Qun Liu. 2021.
\newblock \href {https://doi.org/10.18653/v1/2021.findings-acl.218} {{R}eal{T}ran{S}: End-to-end simultaneous speech translation with convolutional weighted-shrinking transformer}.
\newblock In \emph{Findings of the Association for Computational Linguistics: ACL-IJCNLP 2021}, pages 2461--2474, Online. Association for Computational Linguistics.

\bibitem[{Zeng et~al.(2022)Zeng, Li, Li, and Liu}]{zeng-etal-2022-end}
Xingshan Zeng, Pengfei Li, Liangyou Li, and Qun Liu. 2022.
\newblock \href {https://doi.org/10.18653/v1/2022.autosimtrans-1.5} {End-to-end simultaneous speech translation with pretraining and distillation: Huawei {N}oah{'}s system for {A}uto{S}im{T}ran{S} 2022}.
\newblock In \emph{Proceedings of the Third Workshop on Automatic Simultaneous Translation}, pages 25--33, Online. Association for Computational Linguistics.

\bibitem[{Zhang et~al.(2021)Zhang, Titov, Haddow, and Sennrich}]{zhang-etal-2021-beyond}
Biao Zhang, Ivan Titov, Barry Haddow, and Rico Sennrich. 2021.
\newblock \href {https://doi.org/10.18653/v1/2021.acl-long.200} {Beyond sentence-level end-to-end speech translation: Context helps}.
\newblock In \emph{Proceedings of the 59th Annual Meeting of the Association for Computational Linguistics and the 11th International Joint Conference on Natural Language Processing (Volume 1: Long Papers)}, pages 2566--2578, Online.

\bibitem[{Zhang et~al.(2023{\natexlab{a}})Zhang, Fan, Bu, and Huang}]{zhang-etal-2023-training}
Linlin Zhang, Kai Fan, Jiajun Bu, and Zhongqiang Huang. 2023{\natexlab{a}}.
\newblock \href {https://doi.org/10.18653/v1/2023.emnlp-main.484} {Training simultaneous speech translation with robust and random wait-k-tokens strategy}.
\newblock In \emph{Proceedings of the 2023 Conference on Empirical Methods in Natural Language Processing}, pages 7814--7831, Singapore. Association for Computational Linguistics.

\bibitem[{Zhang et~al.(2020)Zhang, Lu, Sak, Tripathi, McDermott, Koo, and Kumar}]{9053896}
Qian Zhang, Han Lu, Hasim Sak, Anshuman Tripathi, Erik McDermott, Stephen Koo, and Shankar Kumar. 2020.
\newblock \href {https://doi.org/10.1109/ICASSP40776.2020.9053896} {Transformer transducer: A streamable speech recognition model with transformer encoders and rnn-t loss}.
\newblock In \emph{ICASSP 2020 - 2020 IEEE International Conference on Acoustics, Speech and Signal Processing (ICASSP)}, pages 7829--7833.

\bibitem[{Zhang et~al.(2022)Zhang, He, Wu, and Wang}]{zhang-etal-2022-learning}
Ruiqing Zhang, Zhongjun He, Hua Wu, and Haifeng Wang. 2022.
\newblock \href {https://doi.org/10.18653/v1/2022.acl-long.542} {Learning adaptive segmentation policy for end-to-end simultaneous translation}.
\newblock In \emph{Proceedings of the 60th Annual Meeting of the Association for Computational Linguistics (Volume 1: Long Papers)}, pages 7862--7874, Dublin, Ireland. Association for Computational Linguistics.

\bibitem[{Zhang et~al.(2024)Zhang, Fang, Guo, Ma, Zhang, and Feng}]{zhang2024streamspeech}
Shaolei Zhang, Qingkai Fang, Shoutao Guo, Zhengrui Ma, Min Zhang, and Yang Feng. 2024.
\newblock \href {https://doi.org/10.18653/v1/2024.acl-long.485} {{S}tream{S}peech: Simultaneous speech-to-speech translation with multi-task learning}.
\newblock In \emph{Proceedings of the 62nd Annual Meeting of the Association for Computational Linguistics (Volume 1: Long Papers)}, pages 8964--8986, Bangkok, Thailand. Association for Computational Linguistics.

\bibitem[{Zhang and Feng(2022)}]{zhang-feng-2022-information}
Shaolei Zhang and Yang Feng. 2022.
\newblock \href {https://doi.org/10.18653/v1/2022.emnlp-main.65} {Information-transport-based policy for simultaneous translation}.
\newblock In \emph{Proceedings of the 2022 Conference on Empirical Methods in Natural Language Processing}, pages 992--1013, Abu Dhabi, United Arab Emirates. Association for Computational Linguistics.

\bibitem[{Zhang and Feng(2023)}]{zhang-feng-2023-end}
Shaolei Zhang and Yang Feng. 2023.
\newblock \href {https://doi.org/10.18653/v1/2023.findings-acl.485} {End-to-end simultaneous speech translation with differentiable segmentation}.
\newblock In \emph{Findings of the Association for Computational Linguistics: ACL 2023}, pages 7659--7680, Toronto, Canada. Association for Computational Linguistics.

\bibitem[{Zhang and Feng(2024)}]{zhang2024unified}
Shaolei Zhang and Yang Feng. 2024.
\newblock Unified segment-to-segment framework for simultaneous sequence generation.
\newblock \emph{Advances in Neural Information Processing Systems}, 36.

\bibitem[{Zhang et~al.(2023{\natexlab{b}})Zhang, Han, Qin, Wang, Bapna, Chen, Chen, Li, Axelrod, Wang et~al.}]{zhang2023google}
Yu~Zhang, Wei Han, James Qin, Yongqiang Wang, Ankur Bapna, Zhehuai Chen, Nanxin Chen, Bo~Li, Vera Axelrod, Gary Wang, et~al. 2023{\natexlab{b}}.
\newblock Google usm: Scaling automatic speech recognition beyond 100 languages.
\newblock \emph{arXiv preprint arXiv:2303.01037}.

\bibitem[{Zheng et~al.(2020)Zheng, Ma, Zheng, Liu, and Huang}]{zheng-etal-2020-opportunistic}
Renjie Zheng, Mingbo Ma, Baigong Zheng, Kaibo Liu, and Liang Huang. 2020.
\newblock \href {https://doi.org/10.18653/v1/2020.acl-main.42} {Opportunistic decoding with timely correction for simultaneous translation}.
\newblock In \emph{Proceedings of the 58th Annual Meeting of the Association for Computational Linguistics}, pages 437--442, Online. Association for Computational Linguistics.

\bibitem[{Zhu et~al.(2022)Zhu, Wu, Liu, Zhu, Chen, Zhou, Miao, Wang, and Yu}]{zhu-etal-2022-aisp}
Qinpei Zhu, Renshou Wu, Guangfeng Liu, Xinyu Zhu, Xingyu Chen, Yang Zhou, Qingliang Miao, Rui Wang, and Kai Yu. 2022.
\newblock \href {https://doi.org/10.18653/v1/2022.iwslt-1.16} {The {AISP}-{SJTU} simultaneous translation system for {IWSLT} 2022}.
\newblock In \emph{Proceedings of the 19th International Conference on Spoken Language Translation (IWSLT 2022)}, pages 208--215, Dublin, Ireland (in-person and online). Association for Computational Linguistics.

\end{thebibliography}
\bibliographystyle{acl_natbib}

\newpage

\appendix

\section{Categorized Papers}
\label{app:papers-list}

The papers retrieved for the statistics provided in \S\ref{sec:analysis-and-discussion} are obtained by searching on Semantic Scholar using the following queries:\footnote{Accessed July 6th, 2024.}
\begin{table}[h!]
\small
    \centering
    \begin{tblr}{
      colspec={|X[5]|X|}, row{1} = {c}, hlines,
      }
       \textbf{Query} & \textbf{\#papers} \\
       \hline
        \texttt{simultaneous+speech+translation} & 265 \\
        \texttt{streaming+speech+translation} & 218 \\
        \texttt{real-time+speech+translation} & 265 \\
        \texttt{online+speech+translation} & 250 \\
        \texttt{simultaneous+spoken+ language+translation} & 181 \\
        \texttt{streaming+spoken+language+} \texttt{translation} & 85 \\
        \texttt{real-time+spoken+language+} \texttt{translation} & 218 \\
        \texttt{online+spoken+language+} \texttt{translation} & 69
    \end{tblr}
    \caption{Queries used for research on the Semantic Scholar database with their corresponding number of resulting papers.}
    \label{tab:queries}
\end{table}

Notice that querying for \enquote{\texttt{speech}} already includes the results for \enquote{\texttt{speech-to-text}} and similar combinations. Moreover, since we are interested in trends in SimulST systems, we include only papers proposing models (i.e., excluding corpora, surveys, and metrics) and providing results for the speech-to-text task (i.e., speech-to-speech and/or text-to-text are not considered). Only papers written in English and with an open-access version have been considered.

The analysis resulted in \NPAPERS{} papers, categorized following our taxonomy (Figure \ref{fig:taxonomy}) and reported in the following in chronological order. Notice that, in some cases, the number of papers on the various dichotomies does not sum to \NPAPERS{} since some work proposes, for instance, both cascade and direct models and appear in both categories.

\def\sortas#1{}  
\subsection{By Input Type}
\subsubsection{Bounded Speech (90 papers)}
\paragraph{Automatic Pre-Segmentation (2 papers).}
\sortas{2008}	\citet{kolss-etal-2008-simultaneous},
\sortas{2013}	\citet{shimizu-etal-2013-constructing}

\paragraph{Gold Pre-Segmentation (88 papers).}
\sortas{2006}	\citet{ryu-etal-2006-simultaneous},
\sortas{2008}	\citet{kolss-etal-2008-simultaneous},
\sortas{2013}	\citet{fujita13_interspeech},
\sortas{2013}	\citet{rangarajan-sridhar-etal-2013-segmentation},
\sortas{2013}	\citet{yarmohammadi-etal-2013-incremental},
\sortas{2014}	\citet{oda-etal-2014-optimizing},
\sortas{2014}	\citet{wolk2014real},
\sortas{2015}	\citet{cho-etal-2015-punctuation},
\sortas{2015}	\citet{shavarani-etal-2015-learning},
\sortas{2017}	\citet{cho17_interspeech},
\sortas{2018}	\citet{siahbani-etal-2018-simultaneous},
\sortas{2019}	\citet{xiong2019dutongchuan},
\sortas{2020}	\citet{arivazhagan-etal-2020-translation},
\sortas{2020}	\citet{bahar-etal-2020-start},
\sortas{2020}	\citet{elbayad20_interspeech},
\sortas{2020}	\citet{elbayad-etal-2020-trac},
\sortas{2020}	\citet{han-etal-2020-end},
\sortas{2020}	\citet{ma-etal-2020-simulmt},
\sortas{2020}	\citet{ren-etal-2020-simulspeech},
\sortas{2020}	\citet{wilken-etal-2020-neural},
\sortas{2020}	\citet{yao-haddow-2020-dynamic},
\sortas{2021}	\citet{9414276},
\sortas{2021}	\citet{9414897},\footnote{Unbounded speech theoretically possible but not tested.}
\sortas{2021}	\citet{bahar-etal-2021-without},
\sortas{2021}	\citet{chen-etal-2021-direct},
\sortas{2021}	\citet{karakanta-etal-2021-simultaneous},
\sortas{2021}	\citet{liu-etal-2021-cross},
\sortas{2021}	\citet{liu-etal-2021-ustc},
\sortas{2021}	\citet{nguyen2021impact},
\sortas{2021}	\citet{NOVITASARI20212021EDP7014},
\sortas{2021}	\citet{weller-etal-2021-streaming},
\sortas{2021}	\citet{zaidi2021decision},
\sortas{2021}	\citet{zeng-etal-2021-realtrans},
\sortas{2022}	\citet{chang22f_interspeech},
\sortas{2022}	\citet{deng22b_interspeech},
\sortas{2022}	\citet{dong-etal-2022-learning},
\sortas{2022}	\citet{fukuda-etal-2022-naist},
\sortas{2022}	\citet{gaido-etal-2022-efficient},
\sortas{2022}	\citet{guo-etal-2022-xiaomi},
\sortas{2022}	\citet{indurthi-etal-2022-language},
\sortas{2022}	\citet{iranzo-sanchez-etal-2022-mllp},
\sortas{2022}	\citet{li-etal-2022-system},
\sortas{2022}	\citet{papi-etal-2022-simultaneous},
\sortas{2022}	\citet{polak-etal-2022-cuni},
\sortas{2022}	\citet{subramanya2022multilingual},
\sortas{2022}	\citet{wang-etal-2022-hw-tscs},
\sortas{2022}	\citet{xue22d_interspeech},
\sortas{2022}	\citet{zaidi22_interspeech},
\sortas{2022}	\citet{zeng-etal-2022-end},
\sortas{2022}	\citet{zhang-etal-2022-learning},
\sortas{2022}	\citet{zhang-feng-2022-information},
\sortas{2022}	\citet{zhu-etal-2022-aisp},
\sortas{2023}	\citet{10095896},
\sortas{2023}	\citet{10389709},
\sortas{2023}	\citet{10389799},
\sortas{2023}	\citet{10.5555/3618408.3619591},
\sortas{2023}	\citet{alastruey-etal-2023-towards},
\sortas{2023}	\citet{barrault2023seamless},
\sortas{2023}	\citet{fu-etal-2023-adapting},
\sortas{2023}	\citet{fukuda-etal-2023-naist},
\sortas{2023}	\citet{gaido23_interspeech},
\sortas{2023}	\citet{guo-etal-2023-hw},
\sortas{2023}	\citet{huang-etal-2023-xiaomi},
\sortas{2023}	\citet{ko-etal-2023-tagged},
\sortas{2023}	\citet{ma2023efficient},
\sortas{2023}	\citet{papi2023token},
\sortas{2023}	\citet{papi23_interspeech},
\sortas{2023}	\citet{papi-etal-2023-attention},
\sortas{2023}	\citet{papi-etal-2023-direct},
\sortas{2023}	\citet{polak23_interspeech},
\sortas{2023}	\citet{polak-etal-2023-towards},
\sortas{2023}	\citet{raffel-chen-2023-implicit},
\sortas{2023}	\citet{tang-etal-2023-hybrid},
\sortas{2023}	\citet{wang23oa_interspeech},
\sortas{2023}	\citet{yan-etal-2023-cmus},
\sortas{2023}	\citet{zhang-etal-2023-training},
\sortas{2023}	\citet{zhang-feng-2023-end},
\sortas{2024}	\citet{10446050},
\sortas{2024}	\citet{Chen_Fan_Luo_Zhang_Zhao_Liu_Huang_2024},
\sortas{2024}	\citet{deng2024label},
\sortas{2024}	\citet{guo2024r},
\sortas{2024}	\citet{ko2024naist},
\sortas{2024}	\citet{ma2024non},
\sortas{2024}	\citet{papi2024leveraging},
\sortas{2024}	\citet{papi2024simulseamless},
\sortas{2024}	\citet{tan2024streaming},
\sortas{2024}	\citet{zhang2024streamspeech},
\sortas{2024}	\citet{zhang2024unified}

\subsubsection{Unbounded Speech (20 papers)}

\paragraph{Simultaneous (Automatic) Segmentation (14 papers).}
\sortas{2006}	\citet{1660084},
\sortas{2007}	\citet{fugen2007simultaneous},
\sortas{2008}	\citet{4777883},
\sortas{2009}	\citet{Fuegen2009_1000013594},
\sortas{2013}	\citet{cho13_interspeech},
\sortas{2016}	\citet{muller-etal-2016-lecture},
\sortas{2016}	\citet{niehues16_interspeech},
\sortas{2016}	\citet{wang-etal-2016-efficient},
\sortas{2019}	\citet{wang-etal-2019-online},
\sortas{2020}	\citet{9054585},
\sortas{2020}	\citet{iranzo-sanchez-etal-2020-direct},
\sortas{2020}	\citet{machacek-etal-2020-elitr},
\sortas{2021}	\citet{bojar-etal-2021-operating},
\sortas{2021}	\citet{IRANZOSANCHEZ2021303},

\paragraph{Segmentation-free (6 papers).}
\sortas{2020}	\citet{schneider-waibel-2020-towards},
\sortas{2022}	\citet{amrhein-haddow-2022-dont},
\sortas{2022}	\citet{sen2022simultaneous},
\sortas{2023}	\citet{iranzo2023segmentation},
\sortas{2023}	\citet{polak-2023-long},
\sortas{2024}	\citet{papi2024streamatt}

\subsubsection{Undefined (1 paper)}
\sortas{2018}	\citet{dessloch-etal-2018-kit}

\subsection{By Architecture}
\subsubsection{Direct (64 papers)}
\sortas{2020}	\citet{han-etal-2020-end},
\sortas{2020}	\citet{ma-etal-2020-simulmt},
\sortas{2020}	\citet{ren-etal-2020-simulspeech},
\sortas{2021}	\citet{9414276},
\sortas{2021}	\citet{9414897},
\sortas{2021}	\citet{chen-etal-2021-direct},
\sortas{2021}	\citet{karakanta-etal-2021-simultaneous},
\sortas{2021}	\citet{liu-etal-2021-cross},
\sortas{2021}	\citet{liu-etal-2021-ustc},
\sortas{2021}	\citet{nguyen2021impact},
\sortas{2021}	\citet{zaidi2021decision},
\sortas{2021}	\citet{zeng-etal-2021-realtrans},
\sortas{2022}	\citet{amrhein-haddow-2022-dont},
\sortas{2022}	\citet{chang22f_interspeech},
\sortas{2022}	\citet{deng22b_interspeech},
\sortas{2022}	\citet{dong-etal-2022-learning},
\sortas{2022}	\citet{fukuda-etal-2022-naist},
\sortas{2022}	\citet{gaido-etal-2022-efficient},
\sortas{2022}	\citet{papi-etal-2022-simultaneous},
\sortas{2022}	\citet{polak-etal-2022-cuni},
\sortas{2022}	\citet{subramanya2022multilingual},
\sortas{2022}	\citet{wang-etal-2022-hw-tscs},
\sortas{2022}	\citet{xue22d_interspeech},
\sortas{2022}	\citet{zaidi22_interspeech},
\sortas{2022}	\citet{zhang-etal-2022-learning},
\sortas{2022}	\citet{zhang-feng-2022-information},
\sortas{2022}	\citet{zhu-etal-2022-aisp},
\sortas{2023}	\citet{10095896},
\sortas{2023}	\citet{10389709},
\sortas{2023}	\citet{10389799},
\sortas{2023}	\citet{10.5555/3618408.3619591},
\sortas{2023}	\citet{alastruey-etal-2023-towards},
\sortas{2023}	\citet{barrault2023seamless},
\sortas{2023}	\citet{fu-etal-2023-adapting},
\sortas{2023}	\citet{fukuda-etal-2023-naist},
\sortas{2023}	\citet{gaido23_interspeech},
\sortas{2023}	\citet{huang-etal-2023-xiaomi},
\sortas{2023}	\citet{ko-etal-2023-tagged},
\sortas{2023}	\citet{ma2023efficient},
\sortas{2023}	\citet{papi2023token},
\sortas{2023}	\citet{papi23_interspeech},
\sortas{2023}	\citet{papi-etal-2023-attention},
\sortas{2023}	\citet{papi-etal-2023-direct},
\sortas{2023}	\citet{polak-2023-long},
\sortas{2023}	\citet{polak23_interspeech},
\sortas{2023}	\citet{polak-etal-2023-towards},
\sortas{2023}	\citet{raffel-chen-2023-implicit},
\sortas{2023}	\citet{tang-etal-2023-hybrid},
\sortas{2023}	\citet{wang23oa_interspeech},
\sortas{2023}	\citet{yan-etal-2023-cmus},
\sortas{2023}	\citet{zhang-etal-2023-training},
\sortas{2023}	\citet{zhang-feng-2023-end},
\sortas{2024}	\citet{10446050},
\sortas{2024}	\citet{Chen_Fan_Luo_Zhang_Zhao_Liu_Huang_2024},
\sortas{2024}	\citet{deng2024label},
\sortas{2024}	\citet{guo2024r},
\sortas{2024}	\citet{ko2024naist},
\sortas{2024}	\citet{ma2024non},
\sortas{2024}	\citet{papi2024leveraging},
\sortas{2024}	\citet{papi2024simulseamless},
\sortas{2024}	\citet{papi2024streamatt},
\sortas{2024}	\citet{tan2024streaming},
\sortas{2024}	\citet{zhang2024streamspeech},
\sortas{2024}	\citet{zhang2024unified}

\subsubsection{Cascade (49 papers)}
\sortas{2006}	\citet{1660084},
\sortas{2006}	\citet{ryu-etal-2006-simultaneous},
\sortas{2007}	\citet{fugen2007simultaneous},
\sortas{2008}	\citet{4777883},
\sortas{2008}	\citet{kolss-etal-2008-simultaneous},
\sortas{2009}	\citet{Fuegen2009_1000013594},
\sortas{2013}	\citet{cho13_interspeech},
\sortas{2013}	\citet{fujita13_interspeech},
\sortas{2013}	\citet{rangarajan-sridhar-etal-2013-segmentation},
\sortas{2013}	\citet{shimizu-etal-2013-constructing},
\sortas{2013}	\citet{yarmohammadi-etal-2013-incremental},
\sortas{2014}	\citet{oda-etal-2014-optimizing},
\sortas{2014}	\citet{wolk2014real},
\sortas{2015}	\citet{cho-etal-2015-punctuation},
\sortas{2015}	\citet{shavarani-etal-2015-learning},
\sortas{2016}	\citet{muller-etal-2016-lecture},
\sortas{2016}	\citet{niehues16_interspeech},
\sortas{2016}	\citet{wang-etal-2016-efficient},
\sortas{2017}	\citet{cho17_interspeech},
\sortas{2018}	\citet{dessloch-etal-2018-kit},
\sortas{2018}	\citet{siahbani-etal-2018-simultaneous},
\sortas{2019}	\citet{wang-etal-2019-online},
\sortas{2019}	\citet{xiong2019dutongchuan},
\sortas{2020}	\citet{9054585},
\sortas{2020}	\citet{arivazhagan-etal-2020-translation},
\sortas{2020}	\citet{bahar-etal-2020-start},
\sortas{2020}	\citet{elbayad20_interspeech},
\sortas{2020}	\citet{elbayad-etal-2020-trac},
\sortas{2020}	\citet{iranzo-sanchez-etal-2020-direct},
\sortas{2020}	\citet{machacek-etal-2020-elitr},
\sortas{2020}	\citet{schneider-waibel-2020-towards},
\sortas{2020}	\citet{wilken-etal-2020-neural},
\sortas{2020}	\citet{yao-haddow-2020-dynamic},
\sortas{2021}	\citet{bahar-etal-2021-without},
\sortas{2021}	\citet{bojar-etal-2021-operating},
\sortas{2021}	\citet{IRANZOSANCHEZ2021303},
\sortas{2021}	\citet{NOVITASARI20212021EDP7014},
\sortas{2021}	\citet{weller-etal-2021-streaming},
\sortas{2022}	\citet{guo-etal-2022-xiaomi},
\sortas{2022}	\citet{indurthi-etal-2022-language},
\sortas{2022}	\citet{iranzo-sanchez-etal-2022-mllp},
\sortas{2022}	\citet{li-etal-2022-system},
\sortas{2022}	\citet{sen2022simultaneous},
\sortas{2022}	\citet{subramanya2022multilingual},
\sortas{2022}	\citet{wang-etal-2022-hw-tscs},
\sortas{2022}	\citet{zeng-etal-2022-end},
\sortas{2023}	\citet{guo-etal-2023-hw},
\sortas{2023}	\citet{iranzo2023segmentation},
\sortas{2024}	\citet{guo2024r}

\subsection{By Presentation Strategy}
\subsubsection{Incremental (93 papers)}
\sortas{2006}	\citet{ryu-etal-2006-simultaneous},
\sortas{2007}	\citet{fugen2007simultaneous},
\sortas{2008}	\citet{4777883},
\sortas{2008}	\citet{kolss-etal-2008-simultaneous},
\sortas{2009}	\citet{Fuegen2009_1000013594},
\sortas{2013}	\citet{cho13_interspeech},
\sortas{2013}	\citet{fujita13_interspeech},
\sortas{2013}	\citet{rangarajan-sridhar-etal-2013-segmentation},
\sortas{2013}	\citet{shimizu-etal-2013-constructing},
\sortas{2013}	\citet{yarmohammadi-etal-2013-incremental},
\sortas{2014}	\citet{oda-etal-2014-optimizing},
\sortas{2015}	\citet{shavarani-etal-2015-learning},
\sortas{2016}	\citet{wang-etal-2016-efficient},
\sortas{2018}	\citet{siahbani-etal-2018-simultaneous},
\sortas{2019}	\citet{wang-etal-2019-online},
\sortas{2019}	\citet{xiong2019dutongchuan},
\sortas{2020}	\citet{arivazhagan-etal-2020-translation},
\sortas{2020}	\citet{bahar-etal-2020-start},
\sortas{2020}	\citet{elbayad20_interspeech},
\sortas{2020}	\citet{elbayad-etal-2020-trac},
\sortas{2020}	\citet{han-etal-2020-end},
\sortas{2020}	\citet{iranzo-sanchez-etal-2020-direct},
\sortas{2020}	\citet{ma-etal-2020-simulmt},
\sortas{2020}	\citet{ren-etal-2020-simulspeech},
\sortas{2020}	\citet{schneider-waibel-2020-towards},
\sortas{2020}	\citet{wilken-etal-2020-neural},
\sortas{2021}	\citet{9414276},
\sortas{2021}	\citet{9414897},
\sortas{2021}	\citet{bahar-etal-2021-without},
\sortas{2021}	\citet{chen-etal-2021-direct},
\sortas{2021}	\citet{IRANZOSANCHEZ2021303},
\sortas{2021}	\citet{karakanta-etal-2021-simultaneous},
\sortas{2021}	\citet{liu-etal-2021-cross},
\sortas{2021}	\citet{liu-etal-2021-ustc},
\sortas{2021}	\citet{nguyen2021impact},
\sortas{2021}	\citet{NOVITASARI20212021EDP7014},
\sortas{2021}	\citet{zaidi2021decision},
\sortas{2021}	\citet{zeng-etal-2021-realtrans},
\sortas{2022}	\citet{chang22f_interspeech},
\sortas{2022}	\citet{deng22b_interspeech},
\sortas{2022}	\citet{dong-etal-2022-learning},
\sortas{2022}	\citet{fukuda-etal-2022-naist},
\sortas{2022}	\citet{gaido-etal-2022-efficient},
\sortas{2022}	\citet{guo-etal-2022-xiaomi},
\sortas{2022}	\citet{indurthi-etal-2022-language},
\sortas{2022}	\citet{iranzo-sanchez-etal-2022-mllp},
\sortas{2022}	\citet{li-etal-2022-system},
\sortas{2022}	\citet{papi-etal-2022-simultaneous},
\sortas{2022}	\citet{polak-etal-2022-cuni},
\sortas{2022}	\citet{subramanya2022multilingual},
\sortas{2022}	\citet{wang-etal-2022-hw-tscs},
\sortas{2022}	\citet{xue22d_interspeech},
\sortas{2022}	\citet{zaidi22_interspeech},
\sortas{2022}	\citet{zeng-etal-2022-end},
\sortas{2022}	\citet{zhang-etal-2022-learning},
\sortas{2022}	\citet{zhang-feng-2022-information},
\sortas{2022}	\citet{zhu-etal-2022-aisp},
\sortas{2023}	\citet{10389799},
\sortas{2023}	\citet{10.5555/3618408.3619591},
\sortas{2023}	\citet{barrault2023seamless},
\sortas{2023}	\citet{fu-etal-2023-adapting},
\sortas{2023}	\citet{fukuda-etal-2023-naist},
\sortas{2023}	\citet{gaido23_interspeech},
\sortas{2023}	\citet{guo-etal-2023-hw},
\sortas{2023}	\citet{huang-etal-2023-xiaomi},
\sortas{2023}	\citet{iranzo2023segmentation},
\sortas{2023}	\citet{ko-etal-2023-tagged},
\sortas{2023}	\citet{ma2023efficient},
\sortas{2023}	\citet{papi2023token},
\sortas{2023}	\citet{papi23_interspeech},
\sortas{2023}	\citet{papi-etal-2023-attention},
\sortas{2023}	\citet{papi-etal-2023-direct},
\sortas{2023}	\citet{polak-2023-long},
\sortas{2023}	\citet{polak23_interspeech},
\sortas{2023}	\citet{polak-etal-2023-towards},
\sortas{2023}	\citet{raffel-chen-2023-implicit},
\sortas{2023}	\citet{tang-etal-2023-hybrid},
\sortas{2023}	\citet{wang23oa_interspeech},
\sortas{2023}	\citet{yan-etal-2023-cmus},
\sortas{2023}	\citet{zhang-etal-2023-training},
\sortas{2023}	\citet{zhang-feng-2023-end},
\sortas{2024}	\citet{10446050},
\sortas{2024}	\citet{Chen_Fan_Luo_Zhang_Zhao_Liu_Huang_2024},
\sortas{2024}	\citet{deng2024label},
\sortas{2024}	\citet{guo2024r},
\sortas{2024}	\citet{ko2024naist},
\sortas{2024}	\citet{ma2024non},
\sortas{2024}	\citet{papi2024leveraging},
\sortas{2024}	\citet{papi2024simulseamless},
\sortas{2024}	\citet{papi2024streamatt},
\sortas{2024}	\citet{tan2024streaming},
\sortas{2024}	\citet{zhang2024streamspeech},
\sortas{2024}	\citet{zhang2024unified}

\subsubsection{Re-translation (13)}
\sortas{2016}	\citet{muller-etal-2016-lecture},
\sortas{2016}	\citet{niehues16_interspeech},
\sortas{2020}	\citet{9054585},
\sortas{2020}	\citet{arivazhagan-etal-2020-translation},
\sortas{2020}	\citet{machacek-etal-2020-elitr},
\sortas{2020}	\citet{yao-haddow-2020-dynamic},
\sortas{2021}	\citet{bojar-etal-2021-operating},
\sortas{2021}	\citet{weller-etal-2021-streaming},
\sortas{2022}	\citet{amrhein-haddow-2022-dont},
\sortas{2022}	\citet{sen2022simultaneous},
\sortas{2023}	\citet{10095896},
\sortas{2023}	\citet{10389709},
\sortas{2023}	\citet{alastruey-etal-2023-towards}

\subsubsection{Undefined (5)}
\sortas{2006}	\citet{1660084},
\sortas{2014}	\citet{wolk2014real},
\sortas{2015}	\citet{cho-etal-2015-punctuation},
\sortas{2017}	\citet{cho17_interspeech},
\sortas{2018}	\citet{dessloch-etal-2018-kit}

\subsection{By Papers Mentioning Automatic Segmentation}

\subsubsection{Not Mentioned}
\sortas{2006}	\citet{ryu-etal-2006-simultaneous},
\sortas{2013}	\citet{fujita13_interspeech},
\sortas{2014}	\citet{wolk2014real},
\sortas{2015}	\citet{cho-etal-2015-punctuation},
\sortas{2017}	\citet{cho17_interspeech},
\sortas{2018}	\citet{dessloch-etal-2018-kit},
\sortas{2018}	\citet{siahbani-etal-2018-simultaneous},
\sortas{2019}	\citet{xiong2019dutongchuan},
\sortas{2020}	\citet{arivazhagan-etal-2020-translation},
\sortas{2020}	\citet{bahar-etal-2020-start},
\sortas{2020}	\citet{elbayad20_interspeech},
\sortas{2020}	\citet{elbayad-etal-2020-trac},
\sortas{2020}	\citet{han-etal-2020-end},
\sortas{2020}	\citet{ma-etal-2020-simulmt},
\sortas{2020}	\citet{ren-etal-2020-simulspeech},
\sortas{2020}	\citet{wilken-etal-2020-neural},
\sortas{2020}	\citet{yao-haddow-2020-dynamic},
\sortas{2021}	\citet{9414276},
\sortas{2021}	\citet{chen-etal-2021-direct},
\sortas{2021}	\citet{karakanta-etal-2021-simultaneous},
\sortas{2021}	\citet{liu-etal-2021-cross},
\sortas{2021}	\citet{nguyen2021impact},
\sortas{2021}	\citet{NOVITASARI20212021EDP7014},
\sortas{2021}	\citet{weller-etal-2021-streaming},
\sortas{2021}	\citet{zaidi2021decision},
\sortas{2021}	\citet{zeng-etal-2021-realtrans},
\sortas{2022}	\citet{chang22f_interspeech},
\sortas{2022}	\citet{deng22b_interspeech},
\sortas{2022}	\citet{dong-etal-2022-learning},
\sortas{2022}	\citet{fukuda-etal-2022-naist},
\sortas{2022}	\citet{guo-etal-2022-xiaomi},
\sortas{2022}	\citet{indurthi-etal-2022-language},
\sortas{2022}	\citet{iranzo-sanchez-etal-2022-mllp},
\sortas{2022}	\citet{papi-etal-2022-simultaneous},
\sortas{2022}	\citet{polak-etal-2022-cuni},
\sortas{2022}	\citet{subramanya2022multilingual},
\sortas{2022}	\citet{wang-etal-2022-hw-tscs},
\sortas{2022}	\citet{xue22d_interspeech},
\sortas{2022}	\citet{zaidi22_interspeech},
\sortas{2022}	\citet{zeng-etal-2022-end},
\sortas{2022}	\citet{zhang-etal-2022-learning},
\sortas{2022}	\citet{zhang-feng-2022-information},
\sortas{2022}	\citet{zhu-etal-2022-aisp},
\sortas{2023}	\citet{10095896},
\sortas{2023}	\citet{10389709},
\sortas{2023}	\citet{10389799},
\sortas{2023}	\citet{10.5555/3618408.3619591},
\sortas{2023}	\citet{alastruey-etal-2023-towards},
\sortas{2023}	\citet{barrault2023seamless},
\sortas{2023}	\citet{fu-etal-2023-adapting},
\sortas{2023}	\citet{fukuda-etal-2023-naist},
\sortas{2023}	\citet{gaido23_interspeech},
\sortas{2023}	\citet{guo-etal-2023-hw},
\sortas{2023}	\citet{huang-etal-2023-xiaomi},
\sortas{2023}	\citet{ko-etal-2023-tagged},
\sortas{2023}	\citet{ma2023efficient},
\sortas{2023}	\citet{papi2023token},
\sortas{2023}	\citet{papi23_interspeech},
\sortas{2023}	\citet{papi-etal-2023-attention},
\sortas{2023}	\citet{papi-etal-2023-direct},
\sortas{2023}	\citet{polak23_interspeech},
\sortas{2023}	\citet{polak-etal-2023-towards},
\sortas{2023}	\citet{raffel-chen-2023-implicit},
\sortas{2023}	\citet{tang-etal-2023-hybrid},
\sortas{2023}	\citet{wang23oa_interspeech},
\sortas{2023}	\citet{yan-etal-2023-cmus},
\sortas{2023}	\citet{zhang-etal-2023-training},
\sortas{2023}	\citet{zhang-feng-2023-end},
\sortas{2024}	\citet{10446050},
\sortas{2024}	\citet{Chen_Fan_Luo_Zhang_Zhao_Liu_Huang_2024},
\sortas{2024}	\citet{deng2024label},
\sortas{2024}	\citet{guo2024r},
\sortas{2024}	\citet{ko2024naist},
\sortas{2024}	\citet{ma2024non},
\sortas{2024}	\citet{papi2024leveraging},
\sortas{2024}	\citet{papi2024simulseamless},
\sortas{2024}	\citet{tan2024streaming},
\sortas{2024}	\citet{zhang2024streamspeech},
\sortas{2024}	\citet{zhang2024unified}

\subsubsection{Mentioned}
\sortas{2006}	\citet{1660084},
\sortas{2007}	\citet{fugen2007simultaneous},
\sortas{2008}	\citet{4777883},
\sortas{2008}	\citet{kolss-etal-2008-simultaneous},
\sortas{2009}	\citet{Fuegen2009_1000013594},
\sortas{2013}	\citet{cho13_interspeech},
\sortas{2013}	\citet{rangarajan-sridhar-etal-2013-segmentation},
\sortas{2013}	\citet{shimizu-etal-2013-constructing},
\sortas{2013}	\citet{yarmohammadi-etal-2013-incremental},
\sortas{2014}	\citet{oda-etal-2014-optimizing},
\sortas{2015}	\citet{shavarani-etal-2015-learning},
\sortas{2016}	\citet{muller-etal-2016-lecture},
\sortas{2016}	\citet{niehues16_interspeech},
\sortas{2016}	\citet{wang-etal-2016-efficient},
\sortas{2019}	\citet{wang-etal-2019-online},
\sortas{2020}	\citet{9054585},
\sortas{2020}	\citet{iranzo-sanchez-etal-2020-direct},
\sortas{2020}	\citet{machacek-etal-2020-elitr},
\sortas{2020}	\citet{schneider-waibel-2020-towards},
\sortas{2021}	\citet{9414897},
\sortas{2021}	\citet{bahar-etal-2021-without},
\sortas{2021}	\citet{bojar-etal-2021-operating},
\sortas{2021}	\citet{IRANZOSANCHEZ2021303},
\sortas{2021}	\citet{liu-etal-2021-ustc},
\sortas{2022}	\citet{amrhein-haddow-2022-dont},
\sortas{2022}	\citet{gaido-etal-2022-efficient},
\sortas{2022}	\citet{li-etal-2022-system},
\sortas{2022}	\citet{sen2022simultaneous},
\sortas{2023}	\citet{iranzo2023segmentation},
\sortas{2023}	\citet{polak-2023-long},
\sortas{2024}	\citet{papi2024streamatt}

\end{document}